\documentclass[journal]{IEEEtran}

\usepackage{comment}
\usepackage{multirow}
\usepackage{subfigure}

\usepackage{cite}

\ifCLASSINFOpdf
   \usepackage[pdftex]{graphicx}
\else
\fi
\usepackage{amsfonts}
\usepackage[cmex10]{amsmath}
\usepackage{hyperref}
\hypersetup{
  colorlinks   = true, 
  urlcolor     = blue, 
  linkcolor    = blue, 
  citecolor   = red 
}
\usepackage{url}

\begin{document}
%
\title{ORB-SLAM: a Versatile and Accurate\\ Monocular SLAM System}
%
%
%

\author{Ra\'ul~Mur-Artal*,
	J.~M.~M.~Montiel,~\IEEEmembership{Member,~IEEE,}
        and~Juan~D.~Tard\'os,~\IEEEmembership{Member,~IEEE,}
\thanks{This work was supported by the Direcci\'on General de Investigaci\'on of Spain
under Project DPI2012-32168, the Ministerio de Educaci\'on Scholarship FPU13/04175 and Gobierno de Arag\'on Scholarship B121/13.}
\thanks{The authors are with the Instituto de Investigaci\'on en Ingenier\'ia de Arag\'on (I3A), Universidad de Zaragoza, 
Mar\'ia de Luna 1, 50018 Zaragoza, Spain (e-mail: raulmur@unizar.es; josemari@unizar.es; tardos@unizar.es).} 
\thanks{* Corresponding author.}}

%
%

\thispagestyle{empty}
\onecolumn

\begin{center}
\noindent

This paper has been accepted for publication in \emph{IEEE Transactions on Robotics}.

\vspace{2em}

DOI: \href{http://dx.doi.org/10.1109/TRO.2015.2463671}{10.1109/TRO.2015.2463671}

IEEE Xplore: \url{http://ieeexplore.ieee.org/xpl/articleDetails.jsp?arnumber=7219438}
\end{center}
\vspace{3em}

\copyright2015 IEEE.
Personal use of this material is permitted. Permission 
from IEEE must be obtained for all other uses, in any current or 
future media, including reprinting
/republishing this material for 
advertising or promotional purposes,
creating new collective works, for resale or 
redistribution to servers or lists, or reuse of any copyrighted 
component of this work in other works.

\twocolumn

\markboth{IEEE TRANSACTIONS ON ROBOTICS}%
{Mur-Artal \MakeLowercase{\textit{et al.}}: ORB-SLAM: a Versatile and Accurate Monocular SLAM System}
%



\setcounter{page}{1}

\maketitle


\begin{abstract}
This paper presents ORB-SLAM, a feature-based monocular SLAM system that operates in real time, in small and large, indoor and outdoor environments.  The system is robust to severe motion clutter,
allows wide baseline loop closing and relocalization, and includes full automatic initialization. Building on excellent algorithms of recent years, we designed from scratch a novel system that uses the same features for all SLAM tasks: tracking, mapping, relocalization, and loop closing. A survival of the fittest strategy 
that selects the points and keyframes of the reconstruction leads to excellent robustness and generates a compact and trackable map that only grows if the scene content changes, allowing 
lifelong operation. We present an exhaustive evaluation in 27 sequences from the most popular datasets.  ORB-SLAM achieves unprecedented performance with respect to other state-of-the-art monocular SLAM approaches. 
For the benefit of the community, we make the source code public.
\end{abstract}


%
\IEEEpeerreviewmaketitle

\section{Introduction}
%
%
%
%

\IEEEPARstart{B}{undle Adjustment} (BA) is known to provide accurate estimates of camera localizations as well as a sparse geometrical reconstruction \cite{triggs,hartley}, 
given that a strong network of matches and good initial guesses are provided. For long time this approach was considered  unaffordable for real time applications such 
as Visual Simultaneous Localisation and Mapping (Visual SLAM).  Visual SLAM has the goal of estimating the camera trajectory while reconstructing the environment.
Nowadays we know that to achieve accurate results at non-prohibitive computational cost, a real time SLAM algorithm has to provide BA with:

\begin{itemize}
 \item Corresponding observations of scene features (map points) among a subset of selected frames (keyframes).
 \item As complexity grows with the number of keyframes, their selection should avoid unnecessary redundancy.
 \item A strong network configuration of keyframes and points to produce accurate results, that is,
 a well spread set of keyframes observing points with significant parallax and with plenty of loop closure matches.
 \item An initial estimation of the keyframe poses and point locations for the non-linear optimization.
 \item A local map in exploration where optimization is focused to achieve scalability.  
 \item The ability to perform fast global optimizations (e.g. pose graph) to close loops in real-time.
\end{itemize}

The first real time application of BA was the visual odometry work of Mouragon et. al. \cite{mouragnon}, followed by the ground breaking SLAM work of Klein and Murray \cite{ptam}, known
as Parallel Tracking and Mapping (PTAM). This algorithm, while limited to small scale operation, provides simple but effective methods for keyframe selection,
feature matching, point triangulation, camera localization for every frame, and relocalization after tracking failure. Unfortunately several factors severely limit its application:  lack of loop closing and adequate handling of 
occlusions, low invariance to viewpoint of the relocalization and the need of human intervention for map bootstrapping.

In this work we build on the main ideas of PTAM, the place recognition work of G\'alvez-L\'opez and Tard\'os \cite{dorian}, the scale-aware loop closing
of Strasdat et. al \cite{haukeScale} and the use of covisibility information for large scale operation \cite{DWO,Mei}, to
design from scratch ORB-SLAM, a novel monocular SLAM system whose main contributions are:

 \begin{itemize}
 \item Use of the same features for all tasks: tracking, mapping, relocalization and loop closing. This makes our system more efficient, simple and reliable.
 We use ORB features \cite{orb} which allow real-time performance without GPUs, providing good invariance to changes in viewpoint and
 illumination.
 \item Real time operation in large environments.  Thanks to the use of a covisibility graph,  tracking and mapping is focused in a local covisible area, independent of global map size.
 \item Real time loop closing based on the optimization of a pose graph that we call the \emph{Essential Graph}.  It is
 built from a spanning tree maintained by the system, loop closure links and strong edges from the covisibility graph.
 \item Real time camera relocalization with significant invariance to viewpoint and illumination. This allows recovery from tracking failure and also enhances
 map reuse.
 \item A new automatic and robust initialization procedure based on model selection that permits to create an initial map of
 planar and non-planar scenes. 
 \item A \emph{survival of the fittest} approach to map point and keyframe selection that is generous in the spawning but
 very restrictive in the culling. This policy improves tracking robustness, and enhances lifelong operation because redundant keyframes are discarded. \end{itemize}

We present an extensive evaluation in popular public datasets from indoor and outdoor environments, including hand-held, car and robot sequences.
Notably, we achieve better camera localization accuracy than the state of the art in direct methods \cite{LSDSLAM}, which optimize
directly over pixel intensities instead of feature reprojection errors. We include a discussion in Section \ref{sec:concl:vs} on the possible causes that 
can make feature-based methods more accurate than direct methods.

The loop closing and relocalization methods here presented are based on our previous work \cite{icra14}. A preliminary
version of the system was presented in \cite{orbSLAM}. In the current paper we add the initialization method, the \emph{Essential Graph},
and perfect all methods involved. We also describe in detail all building blocks and perform an exhaustive experimental validation.

To the best of our knowledge, this is the most complete and reliable solution to monocular SLAM, and for the benefit of the community 
we make the source code public. Demonstration videos and the code can be found in our project webpage\footnote{\url{http://webdiis.unizar.es/~raulmur/orbslam}}.

\section{Related Work}

\subsection{Place Recognition}

The survey by Williams et al. \cite{placeSurvey} compared several approaches for place recognition and 
concluded that techniques based on appearance, that is image to image matching,
scale better in large environments than map to map or image to map methods.
Within appearance based methods, bags of words techniques \cite{nister}, such as the probabilistic approach FAB-MAP \cite{fabmap2}, are to the fore
because of their high efficiency. 
DBoW2 \cite{dorian} used for the first time bags of binary words obtained from
BRIEF descriptors \cite{brief} along with the very efficient FAST feature detector \cite{fast}. This reduced in more than one order of magnitude
the time needed for feature extraction, compared to SURF \cite{surf} and SIFT \cite{sift} features that were used in bags of words approaches so far. 
Although the system demonstrated to be very efficient and robust, the use of BRIEF, neither rotation nor scale invariant, limited the system 
to in-plane trajectories and loop detection from similar viewpoints. In our previous work \cite{icra14}, we proposed a bag of words place recognizer
built on DBoW2 with ORB \cite{orb}. ORB are binary features invariant to rotation and scale (in a certain range), resulting in a very fast recognizer with
good invariance to viewpoint. We demonstrated the high recall and robustness of the recognizer
in four different datasets, requiring less than 39ms (including feature extraction) to retrieve a loop candidate from a 10K image database. 
In this work we use an improved version of that place recognizer, using covisibility information and 
returning several hypotheses when querying the database instead of just the best match.

\subsection{Map Initialization}
Monocular SLAM requires a procedure to create an initial map because depth cannot be recovered from a single image.
One way to solve the problem is to initially track a known structure \cite{monoSLAM}. In the context of filtering approaches, points can be  initialized 
with high uncertainty in depth using an inverse depth parametrization \cite{inverseDepth}, which hopefully will later converge to their real positions. 
The recent semi-dense work of Engel et al. \cite{LSDSLAM}, follows a similar approach initializing the depth of the pixels to a random value with high variance.

Initialization methods from two views either assumes locally scene planarity \cite{ptam,SVO} and 
recover the relative camera pose from a homography using the method of Faugeras et. al \cite{faugeras}, or compute an essential matrix \cite{robustSLAM, icra14BRIEF} that
models planar and general scenes, using the five-point algorithm of Nister \cite{nister5}, which requires to deal with multiple solutions.
Both reconstruction methods are not well constrained under low parallax and suffer from a twofold ambiguity solution if all points of a planar scene are 
closer to one of the camera centers \cite{longuet}. On the other hand if a non-planar scene is seen with parallax a unique fundamental matrix can be computed with the eight-point algorithm \cite{hartley}
 and the relative camera pose can be recovered without ambiguity.

We present in Section \ref{sec:ini} a new automatic approach based on model selection between a homography for planar scenes and a fundamental matrix for non-planar scenes.
A statistical approach to model selection was proposed by Torr et al. \cite{torr}.
Under a similar rationale we have developed a heuristic initialization algorithm that takes into account the risk of selecting a fundamental matrix 
in close to degenerate cases (i.e. planar, nearly planar, and low parallax), favoring the selection of the homography.
In the planar case, for the sake of safe operation, 
we refrain from initializing if the solution has a twofold ambiguity, as a corrupted solution could be selected.
We delay the initialization until the method produces a unique solution with significant parallax.

\subsection{Monocular SLAM}
Monocular SLAM was initially solved by filtering \cite{monoSLAM, chiuso, eadeEKF, inverseDepth}. In that approach every frame is processed by the filter to jointly estimate the 
map feature locations and the camera pose. It has the drawbacks of wasting computation in processing consecutive frames 
with little new information and the accumulation of linearization errors. 
On the other hand keyframe-based approaches \cite{mouragnon,ptam} estimate the map using only selected frames (keyframes) allowing to perform more costly but accurate
bundle adjustment optimizations, as mapping is not tied to frame-rate. 
Strasdat et. al \cite{whyFilter} demonstrated that keyframe-based techniques are more accurate than filtering for the same computational cost. 

The most representative keyframe-based SLAM system is probably PTAM by Klein and Murray \cite{ptam}.
It was the first work to introduce the idea of splitting camera tracking and mapping in parallel threads,
and demonstrated to be successful for real time augmented reality applications in small environments. The original version
was later improved with edge features, a rotation estimation step during tracking and a better relocalization method \cite{PTAM2}. 
The map points of PTAM correspond to FAST corners matched by patch correlation. This makes the points only useful for tracking but not for 
place recognition. In fact PTAM does not detect large loops, and the relocalization is based on the correlation
of low resolution thumbnails of the keyframes, yielding a low invariance to viewpoint. 

Strasdat et. al \cite{haukeScale} presented a large scale monocular SLAM system with a front-end based on optical flow implemented on a GPU,
followed by FAST feature matching and \emph{motion-only BA}, and a back-end based on sliding-window BA. Loop closures were solved with a 
pose graph optimization with similarity constraints (7DoF), that was able to correct the scale drift appearing in monocular SLAM. 
From this work we take the idea of loop closing with 7DoF pose graph optimization and apply it to the 
\emph{Essential Graph} defined in Section \ref{sec:sys:graph}

Strasdat et. al \cite{DWO} used the front-end of PTAM, but performed the tracking only in a local map retrieved from a covisibility graph.  
They proposed a double window optimization back-end that continuously performs BA in the inner window, and pose graph in a limited-size outer window. 
However, loop closing is only effective if the size of the outer window is large enough to include the whole loop.  In our system we 
take advantage of the excellent ideas of using a local map based on covisibility, and building the pose graph from the covisibility graph, but apply 
them in a totally redesigned front-end and back-end. Another difference is that, instead of using specific features for loop detection (SURF), 
we perform the place recognition on the same tracked and mapped features, obtaining robust frame-rate relocalization and loop detection.

Pirker et. al \cite{CDSLAM} proposed CD-SLAM, a very complete system including loop closing, relocalization, large scale operation and efforts
to work on dynamic environments. However map initialization is not mentioned. 
The lack of a public implementation does not allow us to perform a comparison of accuracy, robustness or large-scale capabilities.

The visual odometry of Song et al. \cite{icraSong} uses ORB features for tracking and a temporal sliding window BA back-end.
In comparison our system is more general as they do not have global relocalization, loop closing and do not reuse the map. 
They are also using the known distance from the camera to the ground to limit monocular scale drift. 

Lim et. al \cite{icra14BRIEF}, work published after we submitted our preliminary version of this work \cite{orbSLAM}, use also the same 
features for tracking, mapping and loop detection. However the choice of BRIEF limits the system to in-plane trajectories. Their system
only tracks points from the last keyframe so the map is not reused if revisited (similar to visual odometry) and has the problem of growing unbounded.
We compare qualitatively our results with this approach in section \ref{sec:exp:kitti}.

The recent work of Engel et. al \cite{LSDSLAM}, known as LSD-SLAM, is able to build large scale semi-dense maps, using direct methods (i.e. optimization
directly over image pixel intensities) instead
of bundle adjustment over features. Their results are very impressive as the system is able to operate in real time, without GPU acceleration,
building a semi-dense map, with more potential applications for robotics than the sparse output generated by feature-based SLAM.
Nevertheless they still need features for loop detection and their camera localization accuracy is significantly lower than in our system and PTAM, as
we show experimentally in Section \ref{sec:exp:tum}. This surprising result is discussed in Section \ref{sec:concl:vs}.

In a halfway between direct and feature-based methods is the semi-direct visual odometry SVO of Forster et al. \cite{SVO}. 
Without requiring to extract features in every frame they are able to operate at high frame-rates obtaining impressive results 
in quadracopters. However no loop detection is performed and the current implementation is mainly thought for downward looking cameras.

Finally we want to discuss about keyframe selection. All visual SLAM works in the literature agree that running BA with all the points and all the frames is not feasible. 
The work of Strasdat et al. \cite{whyFilter} showed that the most cost-effective approach is to keep as much points as possible, 
while keeping only non-redundant keyframes. The PTAM approach was to insert keyframes very cautiously to avoid an excessive growth of the computational complexity. 
This restrictive keyframe insertion policy  makes the tracking fail in hard exploration conditions. 
Our \emph{survival of the fittest} strategy achieves unprecedented robustness in difficult scenarios by inserting keyframes as 
quickly as possible, and removing later the redundant ones, to avoid the extra cost.

\section{System Overview}

\begin{figure}[t]
        \centering
        \includegraphics[width=0.47\textwidth]{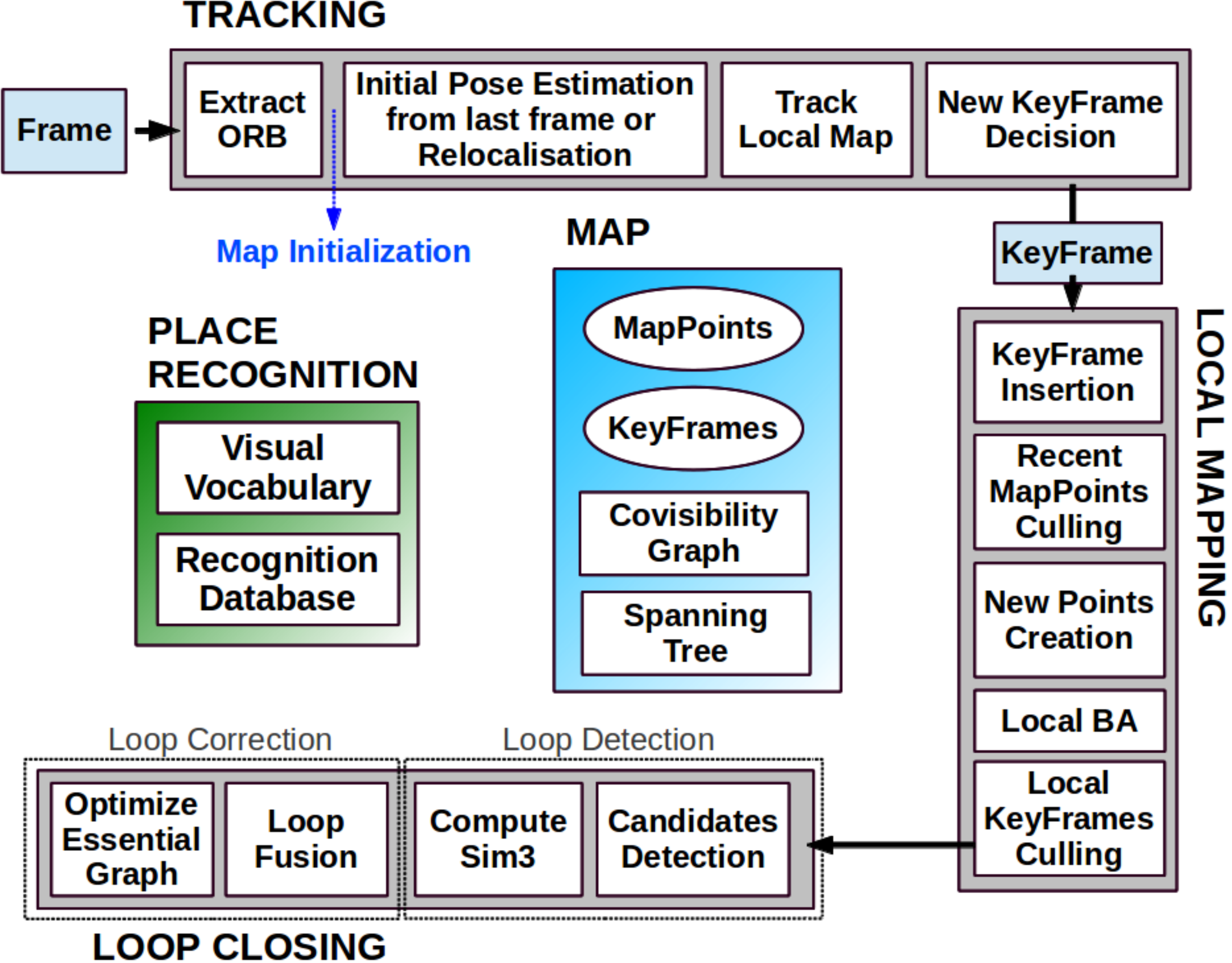}

        \caption{ORB-SLAM system overview, showing all the steps performed by the tracking, local mapping and loop closing threads. The main components of the place
        recognition module and the map are also shown.}
        \label{fig:system} 
\end{figure}

\subsection{Feature Choice}
One of the main design ideas in our system is that the same features used by the mapping
and tracking are used for place recognition to perform frame-rate relocalization and loop detection. This makes our system efficient and avoids the need to interpolate 
the depth of the recognition features from near SLAM features as in previous works \cite{haukeScale,DWO}.
We requiere features that need for extraction much less than 33ms per image, 
which excludes the popular SIFT ($\sim300\textrm{ms}$) \cite{sift}, SURF ($\sim300\textrm{ms}$) \cite{surf} or the recent A-KAZE ($\sim100\textrm{ms}$) \cite{akaze}. 
To obtain general place recognition capabilities, we require rotation invariance, which excludes BRIEF \cite{brief} and LDB \cite{ldb}. 

We chose ORB \cite{orb}, which are oriented multi-scale FAST corners with a 
256 bits descriptor associated. They are extremely fast to compute and match, 
while they have good invariance to viewpoint. This allows to match them from wide baselines, boosting the accuracy of BA.
We already shown the good performance of ORB for place recognition in \cite{icra14}.
While our current implementation make use of ORB, the techniques proposed are not restricted to these features. 

\subsection{Three Threads: Tracking, Local Mapping and Loop Closing}
\begin{figure}[t]
\centering
        \subfigure[KeyFrames (blue), Current Camera (green), MapPoints (black, red), Current Local MapPoints (red)]
        {
    \includegraphics[width=0.22\textwidth]{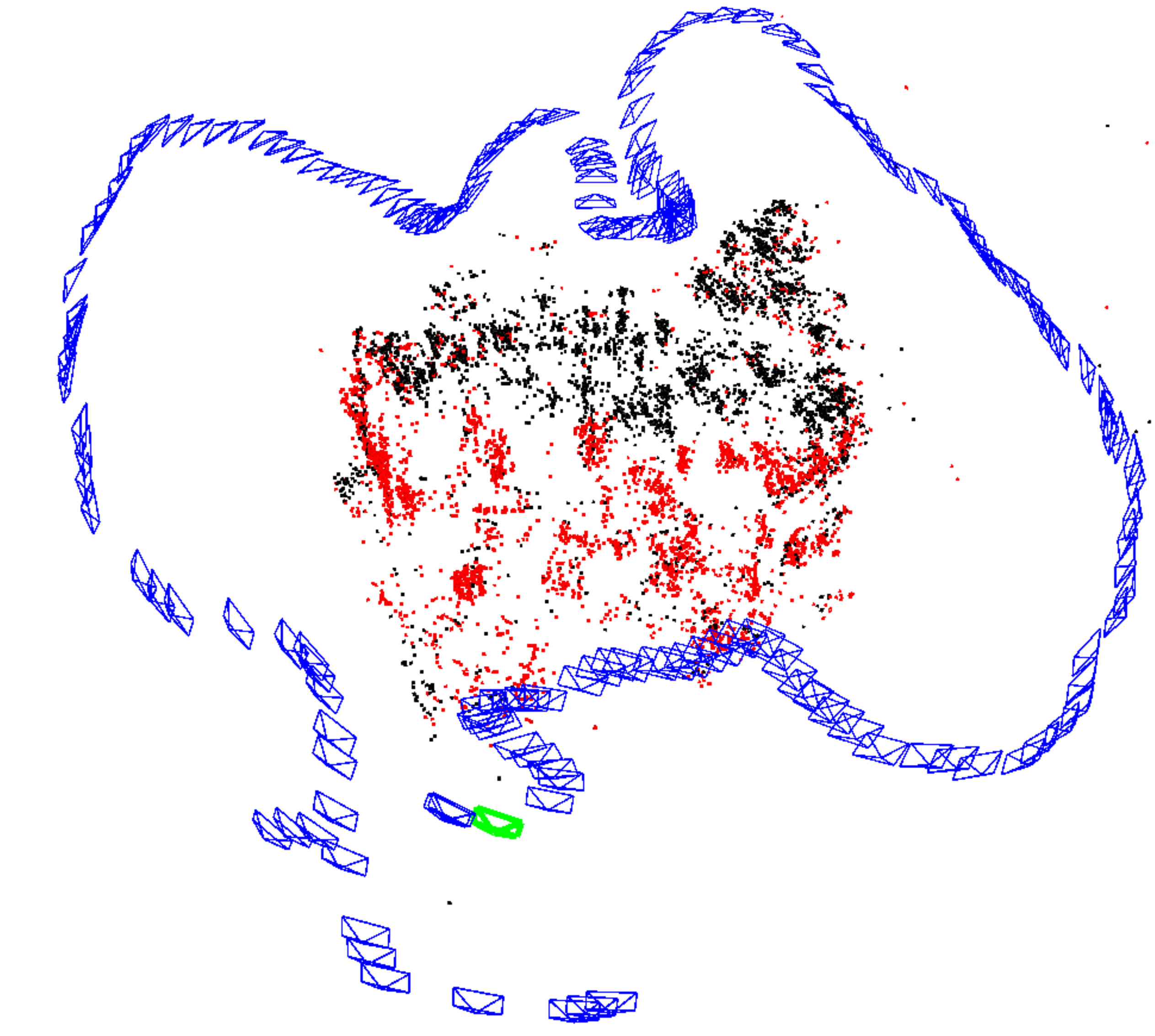}
    \label{fig:graphs:KFMP}
    }
    \subfigure[Covisibility Graph]
        {
    \includegraphics[width=0.22\textwidth]{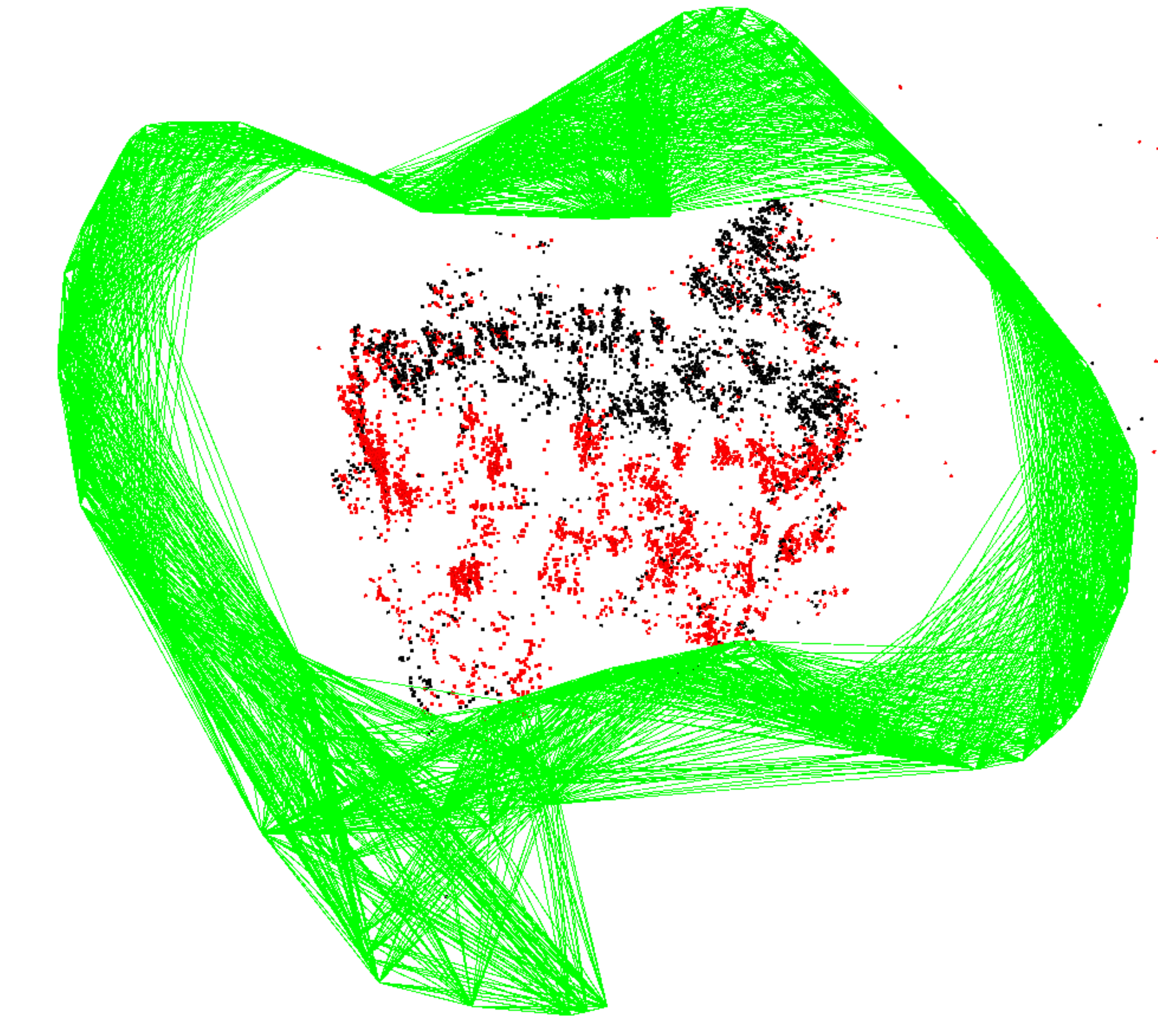}
        \label{fig:graphs:cov}
    }
    
    \subfigure[Spanning Tree (green) and Loop Closure (red)]
        {
    \includegraphics[width=0.22\textwidth]{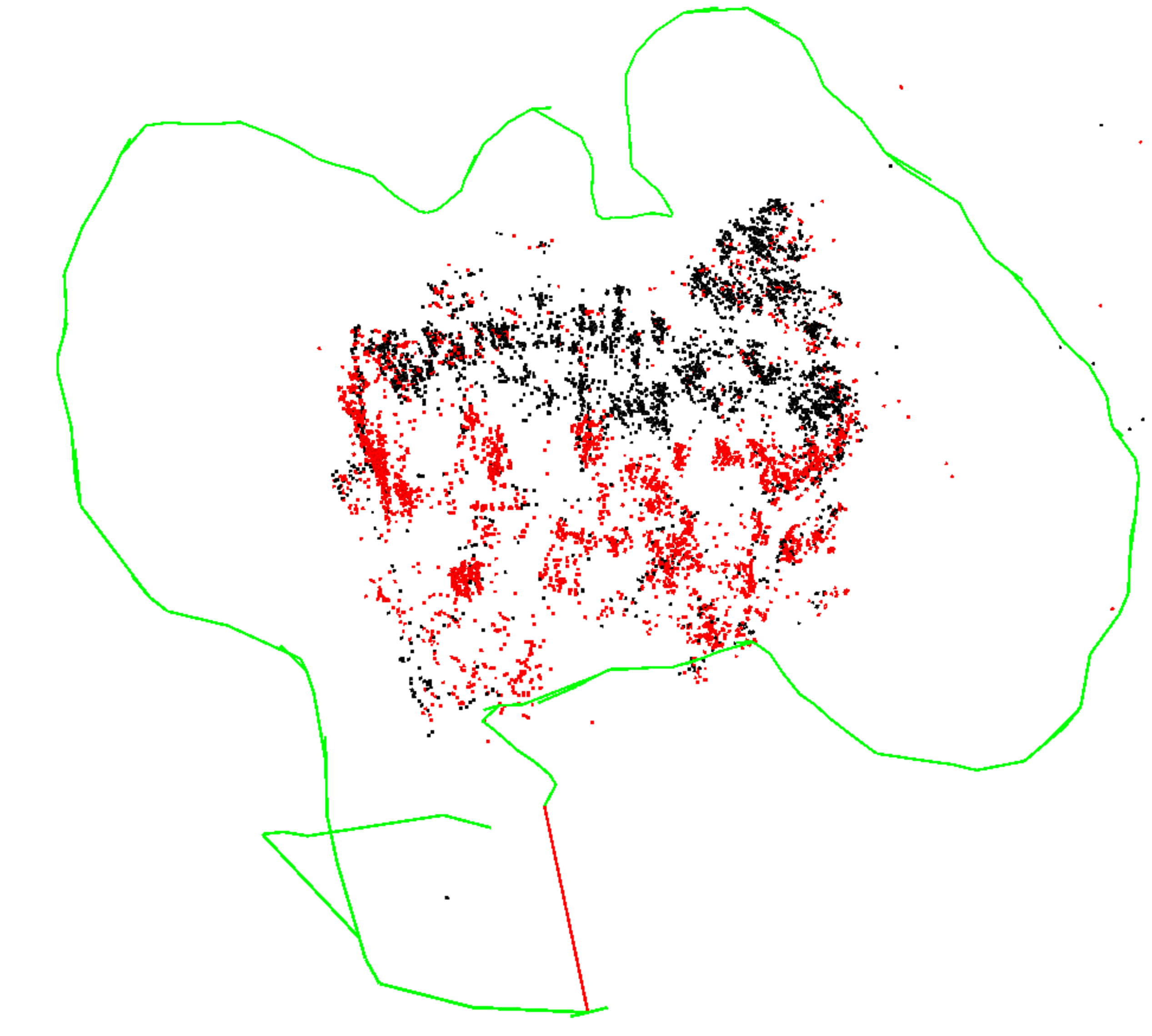}
        \label{fig:graphs:tree}
    }
    \subfigure[Essential Graph]
        {
    \includegraphics[width=0.22\textwidth]{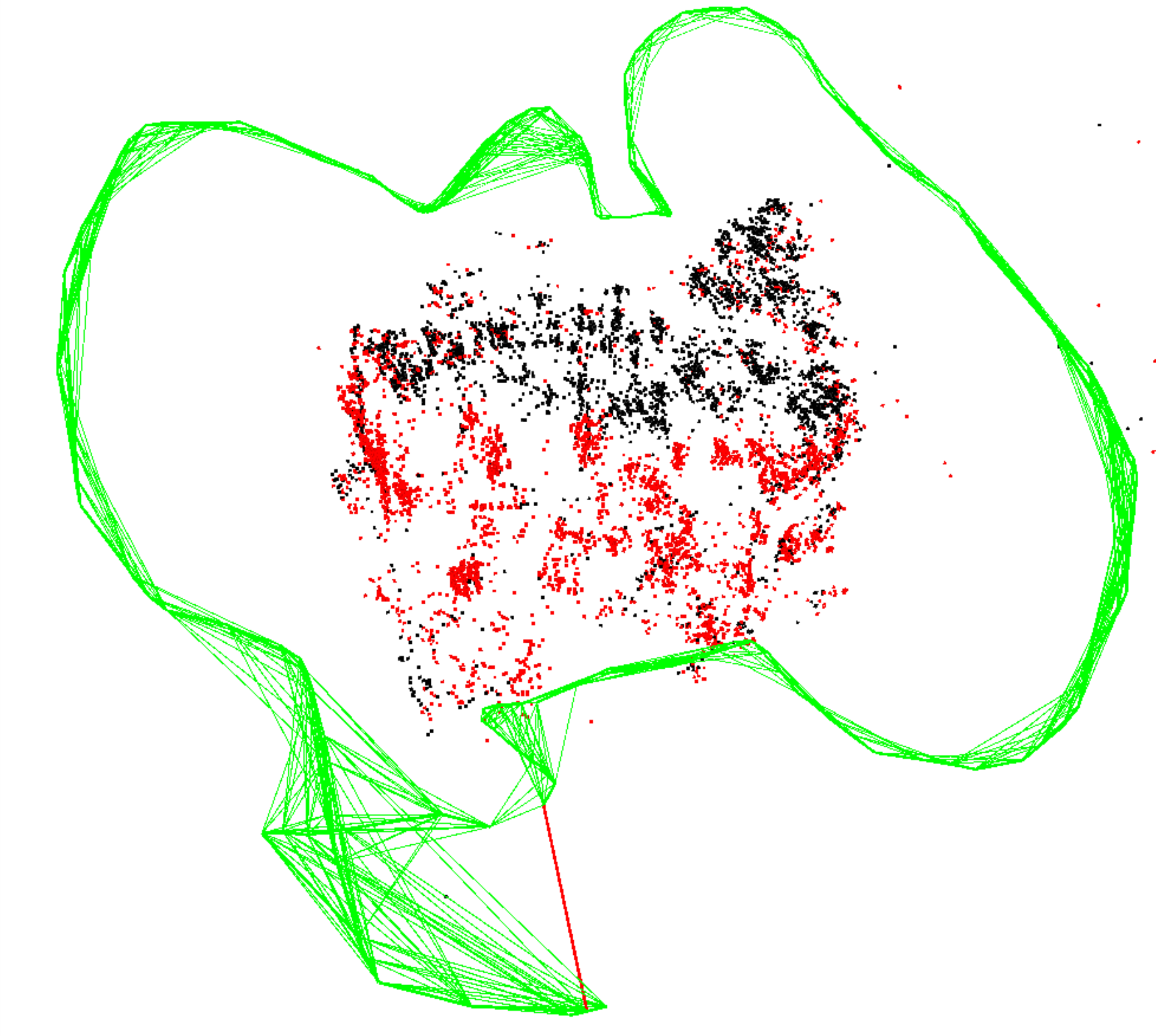}
        \label{fig:graphs:essential}
    }
    
    \caption{Reconstruction and graphs in the sequence \emph{fr3\_long\_office\_household} from the TUM RGB-D Benchmark \cite{tumrgbd}.}
    \label{fig:graphs}
 
\end{figure}

Our system, see an overview in Fig. \ref{fig:system}, incorporates three threads that run in parallel: tracking, local mapping and loop closing.
The tracking is in charge of localizing the camera with every frame and deciding when to insert a new keyframe. 
We perform first an initial feature matching with the previous frame and optimize the pose using \emph{motion-only BA}. 
If the tracking is lost (e.g. due to occlusions or abrupt movements), the place recognition module is used to perform a global 
relocalization. Once there is
an initial estimation of the camera pose and feature matchings, a local visible map is retrieved using the covisibility graph of keyframes that is maintained by the system, see 
Fig. \ref{fig:graphs:KFMP} and Fig. \ref{fig:graphs:cov}.
Then matches with the local map points are searched by reprojection, and camera pose is optimized again with all matches. Finally the tracking thread 
decides if a new keyframe is inserted. All the tracking steps are explained in detail in Section \ref{sec:track}.
 The novel procedure to create an initial map is presented in Section \ref{sec:ini}.

The local mapping processes new keyframes and performs \emph{local BA} to achieve an optimal reconstruction
in the surroundings of the camera pose. New correspondences for unmatched ORB in the new keyframe are searched in connected keyframes
in the covisibility graph to triangulate new points. Some time after creation, based on the information gathered during the tracking, an exigent point culling 
policy is applied in order to retain only high quality points. The local mapping is also in charge of culling redundant keyframes. 
We explain in detail all local mapping steps in Section \ref{sec:mapping}.

The loop closing searches for loops with every new keyframe. If a loop is detected, we compute a similarity transformation that informs about the drift
accumulated in the loop. Then both sides of the loop are aligned and duplicated points are fused. Finally a pose graph optimization over similarity constraints \cite{haukeScale}
is performed to achieve global consistency. The main novelty is that we perform the optimization over the \emph{Essential Graph}, a sparser subgraph of the covisibility graph  which is explained
in Section \ref{sec:sys:graph}. The loop detection and correction steps are explained in detail in Section \ref{sec:loop}.

We use the Levenberg-Marquardt algorithm implemented in g2o \cite{g2o} to carry out all optimizations. In the Appendix we describe the error terms, cost functions, 
and variables involved in each optimization.

\subsection{Map Points, KeyFrames and their Selection}

Each map point $p_i$ stores:
\begin{itemize}
 \item Its 3D position $\textbf{X}_{w,i}$ in the world coordinate system.
 \item The viewing direction $\textbf{n}_i$, which is the mean unit vector of all its viewing directions (the rays that join the point with the optical center of  the keyframes 
 that observe it).
\item A representative ORB descriptor $\textbf{D}_i$, which is the associated ORB descriptor whose hamming distance is minimum with 
respect to all other associated descriptors in the keyframes in which the point is observed.
\item The maximum $d_{\mathrm{max}}$ and minimum $d_{\mathrm{min}}$ distances at which the point can be observed, 
according to the scale invariance limits of the ORB features.
\end{itemize}

Each keyframe $K_i$ stores:
\begin{itemize}
 \item The camera pose $\textbf{T}_{iw}$, which is a rigid body transformation that transforms points from the world to the camera coordinate system.
 \item The camera intrinsics, including focal length and principal point.
 \item All the ORB features extracted in the frame, associated or not to a map point, whose coordinates are undistorted if a distortion model is provided.
\end{itemize}

Map points and keyframes are created with a generous policy, while a later very exigent culling mechanism
is in charge of detecting redundant keyframes and wrongly matched or not trackable map points. This permits a flexible map expansion during exploration, which boost tracking 
robustness under hard conditions (e.g. rotations, fast movements),
while its size is bounded in continual revisits to the same environment, i.e. lifelong operation. Additionally our maps contain very few outliers 
compared with PTAM, at the expense of containing less points. Culling procedures of map points and keyframes are 
explained in Sections \ref{sec:cull:points} and \ref{sec:cull:kf} respectively.

\subsection{Covisibility Graph and Essential Graph} \label{sec:sys:graph}
Covisibility information between keyframes is very useful in several tasks of our system, and is represented as an undirected weighted graph as in \cite{DWO}.
Each node is a keyframe and an edge between two keyframes
exists if they share observations of the same map points (at least 15), being the weight $\theta$ of the edge the number of common map points. 

In order to correct a loop we perform a pose graph optimization \cite{haukeScale} that distributes the loop closing error along the graph. In order not to include 
all the edges provided by the covisibility graph, which can be very dense, we propose to build an \emph{Essential Graph} that retains 
all the nodes (keyframes), but less edges, still preserving a strong network that yields accurate results. The system builds incrementally a 
spanning tree from the initial keyframe, which provides a connected subgraph of the covisibility graph with minimal number
 of edges. When a new keyframe is inserted, it is included in the tree linked to the 
keyframe which shares most point observations, and when a keyframe is erased by the culling policy, the system updates the links 
affected by that keyframe. The \emph{Essential Graph} contains the spanning tree, the subset of edges from the covisibility graph 
with high covisibility ($\theta_{\mathrm{min}}=100$), and the loop closure edges, resulting in a strong network of cameras.
Fig. \ref{fig:graphs} shows an example of a covisibility graph, spanning tree and associated essential graph.
As shown in the experiments of Section \ref{sec:exp:kitti}, when performing the pose graph optimization, the 
solution is so accurate that an additional full bundle adjustment optimization barely improves the solution.
The efficiency of the essential graph and the influence of the $\theta_{\mathrm{min}}$ is shown at the end of Section \ref{sec:exp:kitti}.

\subsection{Bags of Words Place Recognition} \label{sec:sys:bow}
The system has embedded a bags of words place recognition module, based on DBoW2\footnote{\url{https://github.com/dorian3d/DBoW2}} \cite{dorian}, to perform loop detection and relocalization.
Visual words are just a discretization of the descriptor space, which is known as the visual vocabulary. 
The vocabulary is created offline with the ORB descriptors extracted from a large set of images. 
If the images are general enough, the same vocabulary can be used for different environments getting a good performance, as shown
in our previous work \cite{icra14}. The system builds incrementally a database that contains an invert index, which stores for each visual word in the vocabulary, 
in which keyframes it has been seen, so that querying the database can be done very efficiently. 
The database is also updated when a keyframe is deleted by the culling procedure.

Because there exists visual overlap between keyframes, when querying the database there will not exist a unique keyframe with a high score. 
The original DBoW2 took this overlapping into account, adding up the score of images that are close
in time. This has the limitation of not including keyframes viewing the same place but inserted at a different time.
Instead we group those keyframes that are connected in the covisibility graph.
In addition our database returns all keyframe matches whose scores are higher than the $75\%$ of the best score.

An additional benefit of the bags of words representation for feature matching was reported in \cite{dorian}. When we want
to compute the correspondences between two sets of ORB features, we can constraint the brute force matching only to those features
that belong to the same node in the vocabulary tree at a certain level (we select the second out of six), speeding up the search. 
We use this \emph{trick} when searching matches
for triangulating new points, and at loop detection and relocalization. We also
refine the correspondences with an orientation consistency test, see \cite{icra14} for details, that discards outliers ensuring a coherent rotation for all correspondences.

\section{Automatic Map Initialization} \label{sec:ini}
The goal of the map initialization is to compute the relative pose between two frames to triangulate an initial set of map points.
This method should be independent of the scene (planar or general) and 
should not require human intervention to select a good two-view configuration, i.e. a configuration with significant parallax.
We propose to compute in parallel two geometrical models, a homography assuming a planar scene and a fundamental matrix assuming
a non-planar scene. We then use a heuristic to select a model and try to recover the relative pose with a specific method for the selected model. 
Our method only initializes when it is certain that the two-view configuration is safe, detecting low-parallax cases and the well-known twofold
 planar ambiguity \cite{longuet}, avoiding to initialize a corrupted map. The steps of our algorithm are:

\begin{enumerate}\itemsep0.5em
  
  \item Find initial correspondences:
  
  Extract ORB features (only at the finest scale) in the current frame $F_c$ and search for matches $\textbf{x}_c \leftrightarrow \textbf{x}_r$ in the reference frame $F_r$.
  If not enough matches are found, reset the reference frame. 
  
 \item Parallel computation of the two models:
 
 Compute in parallel threads a homography $\textbf{H}_{cr}$ and a fundamental matrix $\textbf{F}_{cr}$:
 \begin{equation}
  \textbf{x}_c = \textbf{H}_{cr} \, \textbf{x}_r
\qquad \qquad
  \textbf{x}_c^T \, \textbf{F}_{cr} \, \textbf{x}_r = 0
 \end{equation}
 with the normalized DLT and 8-point algorithms respectively as explained in \cite{hartley} 
 inside a RANSAC scheme. To make homogeneous the procedure for both models, the number of iterations is prefixed and the same for both models, along with the points to be used at each iteration, 
 8 for the fundamental matrix, and 4 of them for the homography.
 At each iteration we compute a score $S_M$ for each model $M$ ($H$ for the homography, $F$ for the fundamental matrix):
 \begin{equation}
  \begin{gathered}
   S_M = \sum_i \big( \rho_M \big(d_{cr}^2(\textbf{x}_c^i,\textbf{x}_r^i,M)\big) + \rho_M (d_{rc}^2\big(\textbf{x}_c^i,\textbf{x}_r^i,M)\big) \big) 
  \\
  \rho_{M} (d^2) = \begin{cases}
         \Gamma - d^2 & \textrm{if} \quad d^2 < T_M \\
         0  & \textrm{if}  \quad d^2 \geq T_M 
        \end{cases}
  \end{gathered}
 \end{equation}

 where $d_{cr}^2$ and $d_{rc}^2$ are the symmetric transfer errors \cite{hartley} from one frame to the other.  $T_M$ is the outlier rejection threshold based on the ${\chi}^2$ test at 95\% 
 ($T_H = 5.99$, $T_F = 3.84$, assuming a standard deviation of 1 pixel in the measurement error). $\Gamma$ is defined equal to $T_H$ so that both models score equally for 
 the same $d$ in their inlier region, again to make the process homogeneous. 
 
 We keep the homography and fundamental matrix with highest score. If no model could be found (not enough inliers), we restart the process again from step 1.

 \item Model selection:
 
 If the scene is planar, nearly planar or there is low parallax, it can be explained by a homography. However a fundamental matrix can also be found, but the problem
 is not well constrained \cite{hartley} and any attempt to recover the motion from the fundamental matrix would yield wrong results. We should select the homography 
 as the reconstruction method will correctly initialize from a plane or it will detect the low parallax case and refuse the initialization.
 On the other hand a non-planar scene with enough parallax can only be explained by the fundamental matrix, but a homography can also be found explaining a subset
 of the matches if they lie on a plane or they have low parallax (they are far away). In this case we should select the fundamental matrix.
 We have found that a robust heuristic is to compute:
 \begin{equation}
  R_H = \frac{S_H}{S_H+S_F} 
 \end{equation}
 and select the homography if $R_H>0.45$, which adequately captures the planar and low parallax cases. Otherwise, we select the fundamental matrix.

 \item Motion and Structure from Motion recovery:

 Once a model is selected we retrieve the motion hypotheses associated. In the case of the homography we retrieve 8 motion hypotheses using
 the method of Faugeras et. al \cite{faugeras}. The method proposes cheriality tests to select the valid solution. However
 these tests fail if there is low parallax as points easily go in front or back of the cameras, which could yield the selection of a wrong solution.
 We propose to directly triangulate the eight solutions, and check if there is one solution 
 with most points seen with parallax, in front of both cameras and with low reprojection error. If there is not a clear winner solution, we do not initialize and continue
 from step 1. This technique to disambiguate the solutions makes our initialization robust under low parallax and the twofold ambiguity configuration,
 and could be considered the key of the robustness of our method.
 
 In the case of the fundamental matrix, we convert it in an essential matrix using the calibration matrix $\mathbf{K}$:
 \begin{equation}
  \textbf{E}_{rc} = \textbf{K}^{T} \, \textbf{F}_{rc} \, \textbf{K}
 \end{equation}

 and then retrieve 4 motion hypotheses with the singular value decomposition method explained in \cite{hartley}. 
 We triangulate the four solutions and select the reconstruction as done for the homography.

 \item Bundle adjustment:
 
 Finally we perform a \emph{full BA}, see the Appendix for details,
 to refine the initial reconstruction.
 
 \end{enumerate}
 
\begin{figure}[t]
        \centering
        \includegraphics[width=0.485\textwidth]{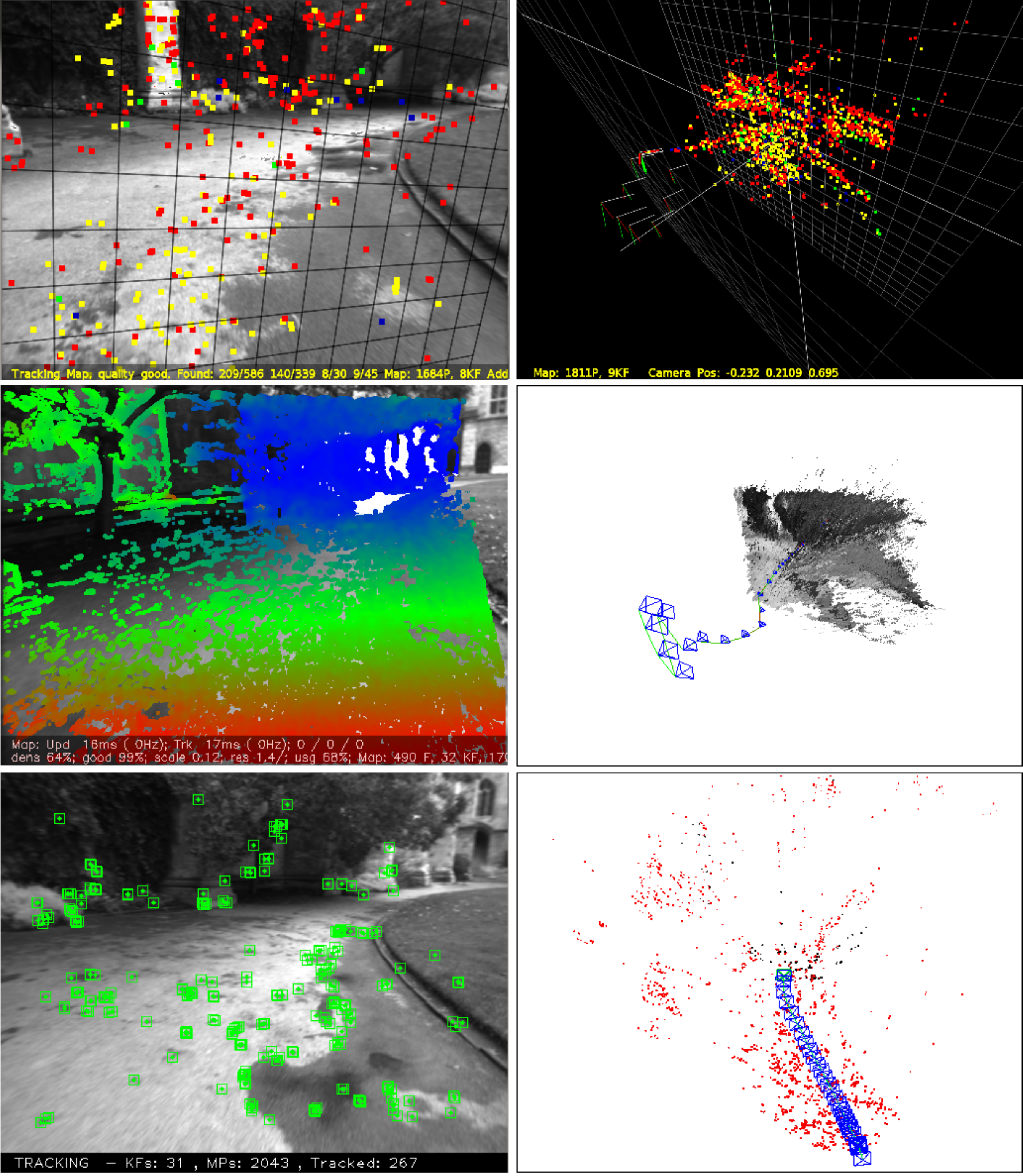}

        \caption{Top: PTAM, middle LSD-SLAM, bottom: ORB-SLAM, some time after initialization in the NewCollege sequence \cite{NewCollege}.
        PTAM and LSD-SLAM initialize a corrupted planar solution while our method has automatically initialized from the fundamental matrix 
        when it has detected enough parallax. Depending on which keyframes are manually selected, PTAM is also able to initialize well. }
        \label{fig:ini} 
\end{figure}

 An example of a challenging initialization in the outdoor NewCollege robot sequence \cite{NewCollege} is shown in Fig. \ref{fig:ini}.
 It can be seen how PTAM and LSD-SLAM have initialized all points in a plane, while our method has waited until there is enough parallax, 
 initializing correctly from the fundamental matrix.

\section{Tracking}\label{sec:track}

In this section we describe the steps of the tracking thread that are performed with every frame from the camera.
The camera pose optimizations, mentioned in several steps, consist in \emph{motion-only BA}, which is described in the Appendix.

 \subsection{ORB Extraction}
 
 We extract FAST corners at 8 scale levels with a scale factor of 1.2. 
 For image resolutions from $512\times384$ to $752\times480$ pixels we found suitable to extract 1000 corners, for higher
 resolutions, as the $1241\times376$ in the KITTI dataset \cite{kitti} we extract 2000 corners.
 In order to ensure an homogeneous distribution we divide each scale level in a grid, trying to extract at least 5 corners per cell. 
 Then we detect corners in each cell, adapting
 the detector threshold if not enough corners are found. The amount of corners retained per cell is also adapted if some cells contains no corners (textureless  
 or low contrast). The orientation and ORB descriptor are then computed on the retained FAST corners. 
 The ORB descriptor is used in all feature matching, in contrast to the search by patch correlation in PTAM.
 
 \subsection{Initial Pose Estimation from Previous Frame} \label{sec:track:1}
 
  
  If tracking was successful for last frame, we use a constant velocity motion model to predict the camera pose and perform a guided search of the 
  map points observed in the last frame. If not enough matches were found (i.e. motion model is clearly violated), we use a wider search
  of the map points around their position in the last frame.
  The pose is then optimized with the found correspondences.
  
   \subsection{Initial Pose Estimation via Global Relocalization} \label{sec:track:2}
  If the tracking is lost, we convert the frame into bag of words and query the recognition database for keyframe candidates for global relocalization.
  We compute correspondences with ORB associated to map points in each keyframe, as explained in section \ref{sec:sys:bow}. 
  We then perform alternatively RANSAC iterations for each keyframe and try
  to find a camera pose using the PnP algorithm \cite{pnp}. 
  If we find a camera pose with enough inliers, we optimize the pose and perform a guided search of more matches with the map points of the candidate keyframe.
  Finally the camera pose is again optimized, and if supported with enough inliers, tracking procedure continues.

 \subsection{Track Local Map} \label{sec:sub:trackmap} \label{sec:track:3}
 Once we have an estimation of the camera pose and an initial set of feature matches, we can project the map into the frame and search more map point correspondences. 
 To bound the complexity in large maps, we only project a local map. 
 This local map contains the set of keyframes $\mathcal{K}_1$, that share map points with the current frame, 
 and a set $\mathcal{K}_2$ with neighbors to the keyframes $\mathcal{K}_1$ in the covisibility graph.
The local map also has a reference keyframe $K_{\mathrm{ref}}\in\mathcal{K}_1$ which shares most map points with current frame.
Now each map point seen in $\mathcal{K}_1$ and $\mathcal{K}_2$ is searched in the current frame as follows:

\begin{enumerate}
 \item Compute the map point projection $\mathbf{x}$ in the current frame. Discard if it lays out of the image bounds.
 \item Compute the angle between the current viewing ray $\mathbf{v}$ and the map point mean viewing direction $\mathbf{n}$. Discard
 if $\mathbf{v} \cdot \mathbf{n}<\textrm{cos}(60^{\circ})$.
 \item Compute the distance $d$ from map point to camera center. 
 Discard if it is out of the scale invariance region of the map point $d \notin [d_{\textrm{min}},d_{\textrm{max}}]$.
 \item Compute the scale in the frame by the ratio $d/d_{\textrm{min}}$.
 \item Compare the representative descriptor $\mathbf{D}$ of the map point with the still unmatched ORB features in the frame, at the predicted scale, and 
 near $\mathbf{x}$, and
 associate the map point with the best match.
\end{enumerate}

The camera pose is finally optimized with all the map points found in the frame. 
 
  \subsection{New Keyframe Decision}\label{sec:track:kf}
  
  The last step is to decide if the current frame is spawned as a new keyframe. As there is a mechanism in the local mapping to cull redundant keyframes, we
  will try to insert keyframes as fast as possible, because that makes the tracking more robust to challenging camera movements, typically rotations.
  To insert a new keyframe all the following conditions must be met:
\begin{enumerate}
 \item More than 20 frames must have passed from the last global relocalization.
 \item Local mapping is idle, or more than 20 frames have passed from last keyframe insertion.
 \item Current frame tracks at least 50 points.
 \item Current frame tracks less than 90\% points than $K_{\mathrm{ref}}$.
 \end{enumerate}
 
 Instead of using a distance criterion to other keyframes as PTAM, we impose a minimum visual change (condition 4). Condition 1 ensures a good relocalization
 and condition 3 a good tracking. If a keyframe is inserted when the local mapping is busy (second part of condition 2), a signal is sent to stop local bundle adjustment, so that
 it can process as soon as possible the new keyframe.

\section{Local Mapping}\label{sec:mapping}

In this section we describe the steps performed by the local mapping with every new keyframe $K_i$.

\subsection{KeyFrame Insertion } \label{sec:sub:bowconv}

At first we update the covisibility graph, adding a new node for $K_i$ and updating the edges resulting from the shared map points with other keyframes. We then 
update the spanning tree linking $K_i$ with the keyframe with most points in common. We then
compute the bags of words representation of the keyframe, that will help in the data association for triangulating new points.

\subsection{Recent Map Points Culling} \label{sec:cull:points}
Map points, in order to be retained in the map, must pass a restrictive test during the first three keyframes after creation, 
that ensures that they are trackable and not wrongly triangulated, i.e due to spurious data association. A point must fulfill these
two conditions:

\begin{enumerate}
 \item The tracking must find the point in more than the 25\% of the frames in which it is predicted to be visible.
 \item If more than one keyframe has passed from map point creation, it must be observed from at least three keyframes.
 \end{enumerate}
 
 Once a map point have passed this test, it can only be removed if at any time it is observed from less than three keyframes. This
 can happen when keyframes are culled and when local bundle adjustment discards outlier observations. This policy makes our map 
 contain very few outliers.
 
\subsection{New Map Point Creation}

New map points are created by triangulating ORB from connected keyframes $\mathcal{K}_c$ in the covisibility graph. For each unmatched ORB in $K_i$
we search a match with other unmatched point in other keyframe. This matching is done as explained in Section \ref{sec:sys:bow} and 
discard those matches that do not fulfill the epipolar constraint.
ORB pairs are triangulated, and to accept the new points, positive depth in both cameras, parallax, reprojection error and scale consistency are checked.
Initially a map point is observed from two keyframes but it could be matched in others, so it is 
projected in the rest of connected keyframes, and correspondences are searched as detailed in section \ref{sec:sub:trackmap}. 

\subsection{Local Bundle Adjustment} \label{sec:sub:localba}

The \emph{local BA} optimizes the currently processed keyframe $K_i$, all the keyframes connected to it in the covisibility graph $\mathcal{K}_c$, 
and all the map points seen by those keyframes. All other keyframes that see those points but are not connected to the currently processed keyframe 
are included in the optimization but remain fixed. Observations that are marked as outliers are discarded at the middle and at the end of the optimization.
See the Appendix for more details about this optimization.

\subsection{Local Keyframe Culling} \label{sec:cull:kf}

In order to maintain a compact reconstruction, the local mapping tries to detect redundant keyframes and delete them.
This is beneficial as bundle adjustment complexity grows with the number of keyframes, but also because it enables lifelong operation
in the same environment as the number of keyframes will not grow unbounded, unless the visual content in the scene changes.
We discard all the keyframes in $\mathcal{K}_c$  whose 90\% of the map points have been seen in at least other three keyframes in the same or finer scale.
The scale condition ensures that map points maintain keyframes from which they are measured with most accuracy.
This policy was inspired by the one proposed in the work of Tan et. al \cite{robustSLAM}, where keyframes were discarded after a process of change detection. 

\section{Loop Closing}\label{sec:loop}
The loop closing thread takes $K_i$, the last keyframe processed by the local mapping, and tries to detect and close loops. The steps are next described.

\subsection{Loop Candidates Detection} \label{sec:sub:detect}
At first we compute the similarity between the bag of words vector of $K_i$ and all its neighbors in the covisibility graph ($ \theta_{min}=30$) 
and retain the lowest score $s_{\textrm{min}}$.
Then we query the recognition database and discard all those keyframes whose score is lower than $s_{\textrm{min}}$. This is a similar operation to gain robustness
as the normalizing score in DBoW2, which is computed from the previous image, but here we use covisibility information. In addition all those keyframes 
directly connected to $K_i$ are discarded from the results. 
To accept a loop candidate we must detect consecutively three loop candidates that are consistent (keyframes connected in the covisibility graph).
There can be several loop candidates if there are several places with similar appearance to $K_i$.

\subsection{Compute the Similarity Transformation} \label{sec:sub:sim3}

In monocular SLAM there are seven degrees of freedom in which the map can drift, three translations, three rotations and a scale factor \cite{haukeScale}. 
Therefore to close a loop we need to compute a similarity transformation from the current keyframe $K_i$ to the loop keyframe $K_l$ that informs us about the error
accumulated in the loop. The computation of this similarity will serve also as geometrical validation of the loop.

We first compute correspondences between ORB associated to map points in the current keyframe and the loop candidate keyframes, 
following the procedure explained in section \ref{sec:sys:bow}. At this point we have 3D to 3D correspondences for each loop candidate. We alternatively perform RANSAC
iterations with each candidate, trying
to find a similarity transformation using the method of Horn \cite{horn}. 
If we find a similarity $\mathbf{S}_{il}$ with enough inliers, we optimize it (see the Appendix), and perform a guided search of more correspondences. We optimize it again 
and, if $\mathbf{S}_{il}$ is supported by enough inliers, the loop with $K_l$ is accepted.

\subsection{Loop Fusion}

The first step in the loop correction is to fuse duplicated map points and insert new edges in the covisibility graph that will attach the loop closure.
At first the current keyframe pose $\mathbf{T}_{iw}$ is corrected with the similarity transformation $\mathbf{S}_{il}$ and this correction is propagated to all the neighbors
of $K_i$, concatenating transformations, so that both sides of the loop get aligned. 
All map points seen by the loop keyframe and its neighbors are projected into $K_i$ and its neighbors and matches are searched in a narrow area around the projection, as done in 
section \ref{sec:sub:trackmap}. All those map points matched and those that were inliers in the computation of $\mathbf{S}_{il}$ are fused.
All keyframes involved in the fusion will update their edges in the covisibility graph effectively creating edges that attach the loop closure.

\subsection{Essential Graph Optimization} \label{sec:sub:pg}

To effectively close the loop, we perform a pose graph optimization over the \emph{Essential Graph}, described in Section \ref{sec:sys:graph}, 
that distributes the loop closing error along the graph. The optimization is performed over 
similarity transformations to correct the scale drift \cite{haukeScale}. The error terms and cost function are detailed in 
the Appendix.
After the optimization each map point is transformed according to the correction
of one of the keyframes that observes it.

\section{Experiments}

We have performed an extensive experimental validation of our system in the large robot sequence of NewCollege \cite{NewCollege}, evaluating
the general performance of the system, in 16 hand-held indoor sequences
of the TUM RGB-D benchmark \cite{tumrgbd}, evaluating the localization
accuracy, relocalization and lifelong capabilities, and in 10 car outdoor sequences from
the KITTI dataset \cite{kitti}, evaluating real-time large scale operation, localization accuracy and efficiency of the pose graph optimization.

Our system runs in real time and processes the images exactly at the frame rate they were acquired. We have carried out 
all experiments with an Intel Core i7-4700MQ (4 cores @ 2.40GHz) and 8Gb RAM. ORB-SLAM has three main threads, that run in parallel with other tasks 
from ROS and the operating system, which introduces some randomness in the results. For this reason, in some experiments, we report the median from several runs.

\subsection{System Performance in the NewCollege Dataset}

The NewCollege dataset \cite{NewCollege} contains a 2.2km sequence from a robot traversing a campus and adjacent parks. The sequence is recorded by a stereo camera
at 20 fps and a resolution $512\times382$. It contains
several loops and fast rotations that makes the sequence quite challenging for monocular vision. To the best of our knowledge there is no other 
monocular system in the literature able to process this whole sequence. For example Strasdat et al. \cite{DWO}, despite being able to close loops and work in
large scale environments, only showed monocular results for a small part of this sequence.

\begin{table} [t]
\caption{Tracking and Mapping times in NewCollege}
\label{tb:nc:t1}
\begin{center}
\begin{tabular}{|c|l|c|c|c|}
  \hline
  &&&& \\[-0.9em]
  Thread & Operation & \parbox[c]{0.85cm}{\centering Median  \\(ms)} & \parbox[c][2.5em]{0.75cm}{\centering Mean \\ (ms)} & \parbox[c]{0.75cm}{\centering Std \\(ms)} \\[0.5em]
  \hline
  \hline
  & & & & \\[-0.8em]
  \multirow{4}{*}{TRACKING} & ORB extraction & 11.10 & 11.42 & 1.61 \\[0.2em]
  \cline{2-5}
   &  & & & \\[-0.8em]
 & Initial Pose Est. & 3.38 & 3.45 & 0.99 \\[0.2em]
 \cline{2-5}
   & & & & \\[-0.8em]
 & Track Local Map & 14.84 & 16.01 & 9.98 \\[0.2em]
 \cline{2-5}
   & & & & \\[-0.8em]
 & Total & 30.57 & 31.60 & 10.39 \\[0.2em]
  \hline
  \hline
  & & & & \\[-0.8em]
  \multirow{6}{*}{\parbox[c]{1.5cm}{\centering LOCAL \\ MAPPING}} &  KeyFrame Insertion & 10.29 & 11.88 & 5.03 \\[0.2em]
   \cline{2-5}
   & & & & \\[-0.8em]
 & Map Point Culling & 0.10 & 3.18 & 6.70 \\[0.2em]
  \cline{2-5}
  & & & & \\[-0.8em]
 & Map Point Creation & 66.79 & 72.96 & 31.48 \\[0.2em]
  \cline{2-5}
  & & & & \\[-0.8em]
 & Local BA & 296.08 & 360.41 & 171.11 \\[0.2em]
  \cline{2-5}
  & & & & \\[-0.8em]
 & KeyFrame Culling & 8.07 & 15.79 & 18.98 \\[0.2em]
  \cline{2-5}
  & & & & \\[-0.8em]
 & Total & 383.59 & 464.27 & 217.89 \\[0.2em]
 \hline
\end{tabular}
\end{center}
\end{table}

\begin{table*} [t]
\caption{Loop closing times in NewCollege}
\label{tb:nc:t2}
\begin{center}
 \begin{tabular}{|c|c|c|c|c|c|c|c|}
 \cline{4-7}
  \multicolumn{3}{c|}{} & \multicolumn{2}{c|}{} & \multicolumn{2}{c|}{} & \multicolumn{1}{}{}\\[-0.9em]
 \multicolumn{3}{c|}{} & \multicolumn{2}{c|}{Loop Detection (ms)} & \multicolumn{2}{c|}{Loop Correction (s)} & \multicolumn{1}{}{}\\[0.1em]
 \hline
 &&&&&&& \\[-0.9em]
  Loop & KeyFrames & \parbox[c][2.5em]{2.1cm}{\centering Essential Graph \\ Edges} & \parbox[c][2.5em]{2cm}{\centering Candidates \\ Detection} 
  & \parbox[c][2.5em]{2cm}{\centering Similarity \\ Transformation} & Fusion & \parbox[c][2.5em]{2.1cm}{\centering Essential Graph \\ Optimization} & Total (s) \\[0.1em]
  \hline
   &&&&&&& \\[-0.9em]
 1 & 287 & 1347 & 4.71 & 20.77 & 0.20 & 0.26 & 0.51 \\[0.1em]
 \hline
  &&&&&&& \\[-0.9em]
 2 & 1082 & 5950 & 4.14 & 17.98 & 0.39 & 1.06 & 1.52\\[0.1em]
 \hline
  &&&&&&& \\[-0.9em]
 3 & 1279 & 7128 & 9.82 & 31.29 & 0.95 & 1.26 & 2.27\\[0.1em]
 \hline
  &&&&&&& \\[-0.9em]
 4 & 2648 & 12547 & 12.37 & 30.36 & 0.97 & 2.30 & 3.33\\[0.1em]
 \hline
  &&&&&&& \\[-0.9em]
 5 & 3150 & 16033 & 14.71 & 41.28 & 1.73 & 2.80 & 4.60\\[0.1em]
 \hline
  &&&&&&& \\[-0.9em]
 6 & 4496 & 21797 & 13.52 & 48.68 & 0.97 & 3.62 & 4.69 \\[0.1em]
 \hline
 \end{tabular} 
\end{center}
\end{table*}

\begin{figure}[t]
    \includegraphics[width=0.47\textwidth]{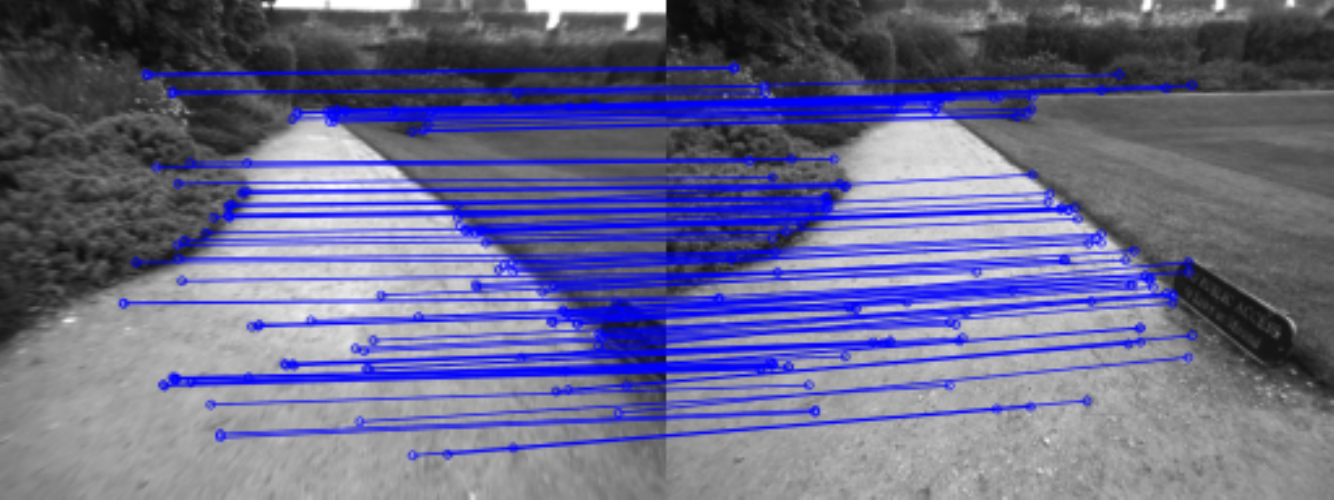}
    \caption{Example of loop detected in the NewCollege sequence. We draw the inlier correspondences supporting the similarity transformation found.}
    \label{fig:nc:loop}
\end{figure}

\begin{figure}[t]
  \centering
        \subfigure
        {
        \includegraphics[width=0.47\textwidth]{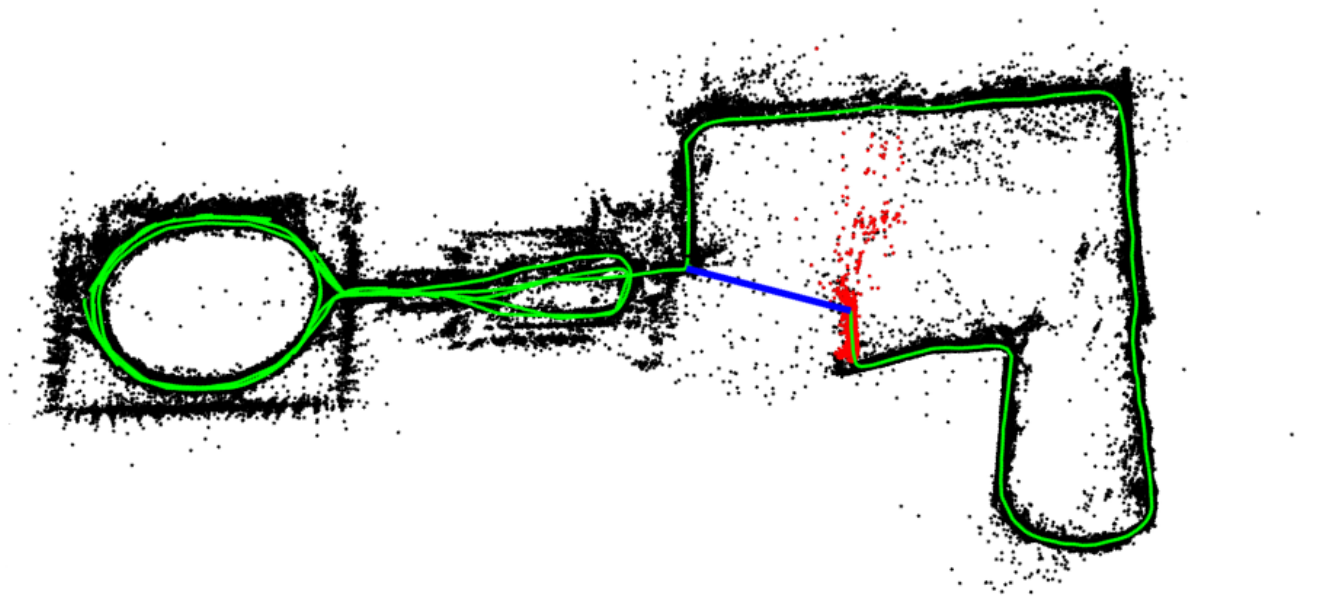}
        }
        \subfigure
        {
        \includegraphics[width=0.47\textwidth]{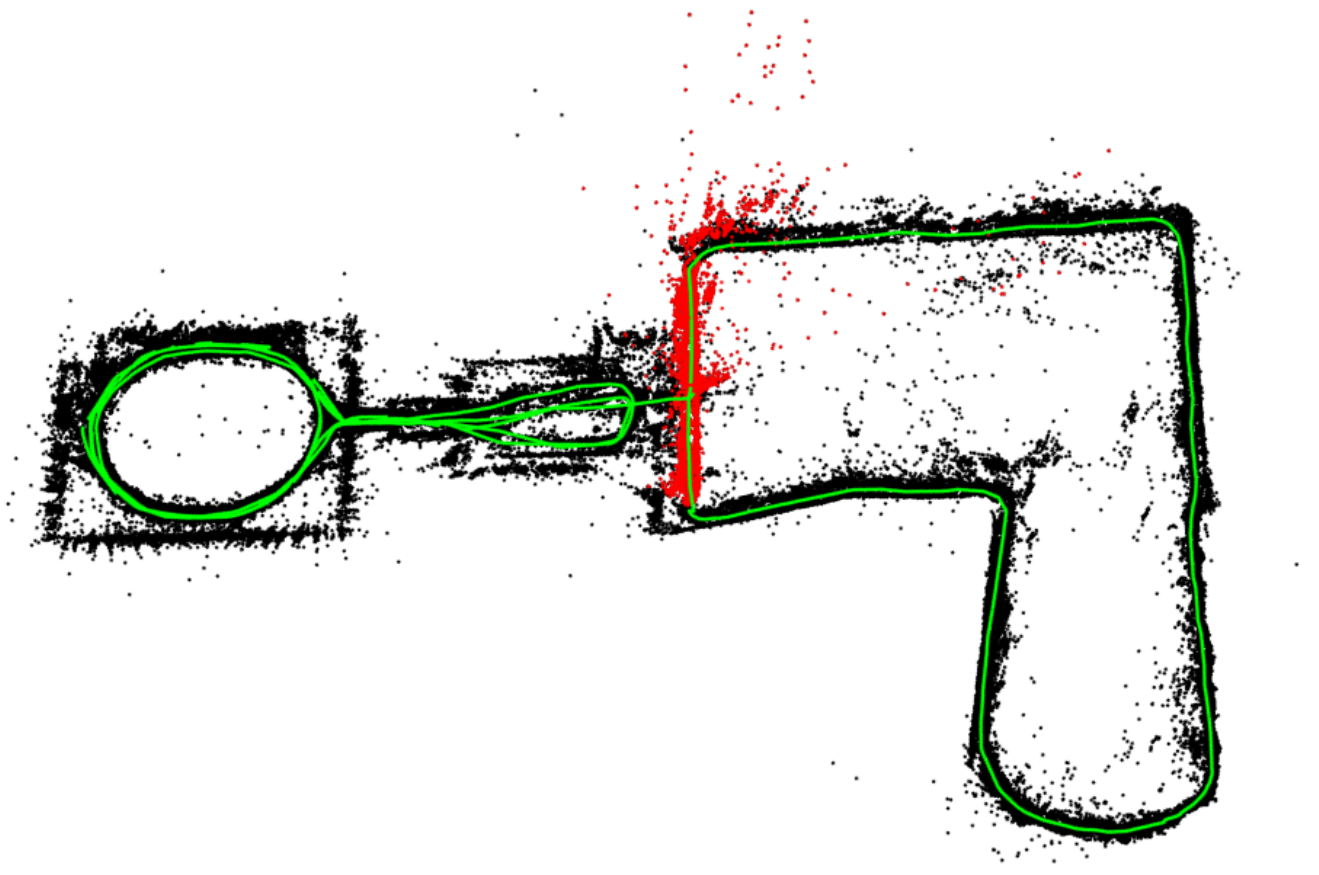}
        }     
        \caption{Map before and after a loop closure in the NewCollege sequence. The loop closure match is drawn in blue, the trajectory in green, and
        the local map for the tracking at that moment in red. The local map is extended along both sides of the loop after it is closed.}
        \label{fig:nc:loopmap}
\end{figure}

\begin{figure}[t]
    \includegraphics[width=0.47\textwidth]{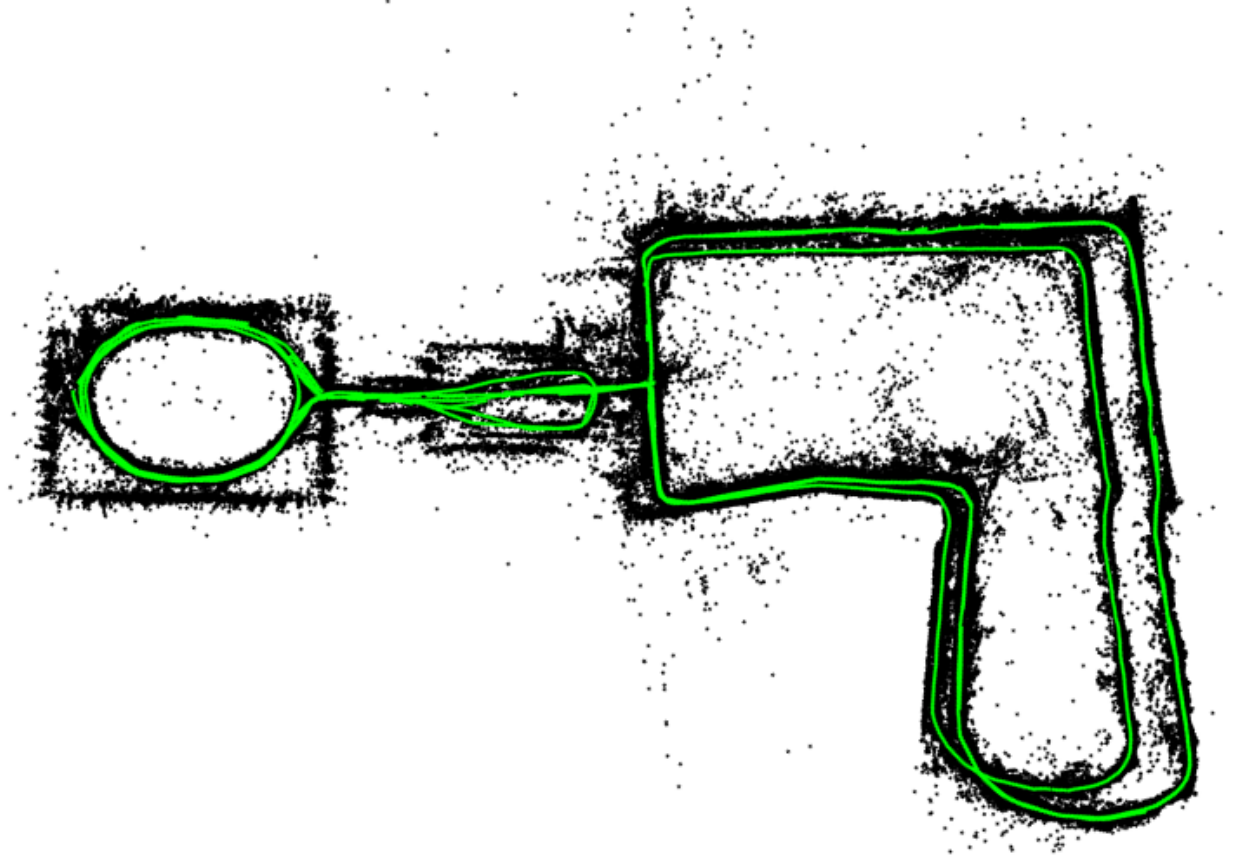}
    \caption{ORB-SLAM reconstruction of the full sequence of NewCollege. The bigger loop on the right is traversed in opposite directions and not visual loop closures were found, therefore they do not perfectly align.}
    \label{fig:nc:map}
 
\end{figure}

As an example of our loop closing procedure we show in Fig. \ref{fig:nc:loop} the detection of a loop with the inliers that support the similarity transformation.
Fig. \ref{fig:nc:loopmap} shows the reconstruction before and after the loop closure. 
In red it is shown the local map, which after the loop closure extends along both sides of the loop closure. 
The whole map after processing the full sequence at its real frame-rate is shown in Fig. \ref{fig:nc:map}. The big loop on the right
does not perfectly align because it was traversed in opposite directions and the place recognizer was not able to find loop closures.

We have extracted statistics of the times spent by each thread in this experiment. Table \ref{tb:nc:t1} shows the results for the tracking and the local mapping. 
Tracking works at frame-rates around 25-30Hz, being the most demanding task to track
the local map. If needed this time could be reduced limiting the number of keyframes that are included in the local map.
In the local mapping thread the most demanding task is local bundle adjustment. The \emph{local BA} time varies if the robot is exploring or in a well mapped area, because
during exploration bundle adjustment is interrupted if tracking inserts a new keyframe, as explained in section \ref{sec:track:kf}. In case
of not needing new keyframes local bundle adjustment performs a generous number of prefixed iterations.

Table \ref{tb:nc:t2} shows the results for each of the 6 loop closures found. It can be seen how the loop detection increases sublinearly with the number of keyframes. 
This is due
to the efficient querying of the database that only compare the subset of images with words in common, 
which demonstrates the potential of bag of words for place recognition. Our \emph{Essential Graph} includes edges
around 5 times the number of keyframes, which is a quite sparse graph.

\subsection{Localization Accuracy in the TUM RGB-D Benchmark} \label{sec:exp:tum}
The TUM RGB-D benchmark \cite{tumrgbd} is an excellent dataset to evaluate the accuracy of camera localization as it provides several sequences with accurate
ground truth obtained with an external motion capture system. We have discarded all those sequences that we consider that are not suitable for pure monocular
SLAM systems, as they contain strong rotations, no texture or no motion. 

For comparison we have also executed the novel, direct, semi-dense LSD-SLAM \cite{LSDSLAM} and PTAM \cite{ptam} in the benchmark. 
We compare also with the trajectories generated by RGBD-SLAM \cite{rgbdslam} which are provided for some of the sequences in the benchmark website. 
In order to compare ORB-SLAM, LSD-SLAM and PTAM with the ground truth, we align the keyframe trajectories using a similarity transformation, as scale is unknown, 
and measure the absolute trajectory error (ATE) \cite{tumrgbd}. In the case of RGBD-SLAM we align the trajectories with a rigid body transformation, but also a similarity
to check if the scale was well recovered. LSD-SLAM initializes from random depth values and takes time to converge, therefore we have discarded the first 10 keyframes when comparing with 
the ground truth. For PTAM we manually selected two frames from which we get a good initialization.
Table \ref{tb:tum} shows the median results over 5 executions in each of the 16 sequences selected. 

It can be seen that ORB-SLAM is able to process all the sequences, except for \emph{fr3\_nostructure\_texture\_far} (fr3\_nstr\_tex\_far). This is a planar scene that because of the camera
 trajectory with respect to the plane has two possible interpretations, i.e. the twofold ambiguity described in \cite{longuet}. Our initialization method detects the ambiguity 
 and for safety refuses to initialize. PTAM initializes selecting sometimes the true solution and others the corrupted one, in which case the error is unacceptable. 
 We have not noticed two different reconstructions from LSD-SLAM but the error in this sequence is very high. 
 In the rest of the sequences, PTAM and LSD-SLAM exhibit less robustness than our method, loosing track in eight and three sequences respectively.
 
In terms of accuracy ORB-SLAM and PTAM are similar in open trajectories, while ORB-SLAM achieves higher accuracy when detecting large loops as in the 
sequence \emph{fr3}\_\emph{nostructure}\_\emph{texture}\_\emph{near}\_\emph{withloop} (fr3\_nstr\_tex\_near). The most surprising results is that both PTAM and ORB-SLAM are clearly more accurate than LSD-SLAM and RGBD-SLAM.
One of the possible causes can be that they reduce
the map optimization to a pose-graph optimization were sensor measurements are discarded, while we perform bundle adjustment and 
jointly optimize cameras and map over sensor measurements, which is the gold standard
algorithm to solve structure from motion \cite{hartley}. We further discuss this result in Section \ref{sec:concl:vs}. Another interesting result 
is that LSD-SLAM seems to be less robust to dynamic objects than our system 
as seen in \emph{fr2\_desk\_with\_person}  and \emph{fr3\_walking\_xyz}. 

We have noticed that RGBD-SLAM has a bias in the scale in \emph{fr2} sequences,
as aligning
the trajectories with 7 DoF significantly reduces the error.
Finally it should be noted that Engel et al. \cite{LSDSLAM} reported that PTAM has less accuracy than 
LSD-SLAM in \emph{fr2\_xyz} with an RMSE of 24.28cm. However, the paper does not give enough details on how those results were obtained, 
and we have been unable to reproduce them.

\begin{table}[t] 
\caption{Keyframe Localization Error Comparison in the
TUM RGB-D Benchmark\cite{tumrgbd}}
\begin{center}
\begin{tabular}{|c|c|c|c|c|}
\cline{2-5}
\multicolumn{1}{c|}{}&\multicolumn{4}{c|}{} \\[-0.8em]
\multicolumn{1}{c|}{} & \multicolumn{4}{c|}{Absolute KeyFrame Trajectory RMSE (cm)}\\[0.2em]
\cline{2-5}
\multicolumn{1}{c|}{}&&& & \\[-0.8em]
\multicolumn{1}{c|}{} & ORB-SLAM & PTAM & LSD-SLAM & \parbox[c][2.5em]{1.0cm}{\centering RGBD- \\SLAM} \\[0.2em]
\hline
\parbox[c][2.5em]{1.0cm}{\centering fr1\_xyz} & \textbf{0.90} & 1.15 & 9.00 &  1.34 (1.34) \\
\hline
\parbox[c][2.5em]{1.0cm}{\centering fr2\_xyz} & 0.30 & \textbf{0.20} & 2.15 &  2.61 (1.42) \\
\hline
\parbox[c][2.5em]{1.0cm}{\centering fr1\_floor} & \textbf{2.99} & X & 38.07 &  3.51 (3.51) \\
\hline
\parbox[c][2.5em]{1.0cm}{\centering fr1\_desk} & \textbf{1.69} & X & 10.65 &  2.58 (2.52) \\
\hline
\parbox[c][2.5em]{1.0cm}{\centering fr2\_360 \\ \_kidnap} & 3.81 & \textbf{2.63} & X & 393.3 (100.5)\\
\hline
\parbox[c][2.5em]{1.0cm}{\centering fr2\_desk} & \textbf{0.88} & X & 4.57 & 9.50 (3.94) \\
\hline
\parbox[c][2.5em]{1.0cm}{\centering fr3\_long\\ \_office} & \textbf{3.45} & X & 38.53 & - \\
\hline
\parbox[c][2.5em]{1.3cm}{\centering fr3\_nstr\_\\ tex\_far} & \parbox[c][2.5em]{1.2cm}{\centering ambiguity \\ detected} & \parbox[c][2.5em]{1.0cm}{\centering 4.92 / \\ 34.74}  & 18.31 & - \\
\hline
\parbox[c][2.5em]{1.3cm}{\centering fr3\_nstr\_\\ tex\_near\\} & \textbf{1.39} & 2.74 & 7.54 & - \\
\hline
\parbox[c][2.5em]{1.0cm}{\centering fr3\_str\_\\ tex\_far} & \textbf{0.77} & 0.93 & 7.95 & - \\
\hline
\parbox[c][2.5em]{1.0cm}{\centering fr3\_str\_\\ tex\_near} & 1.58 & \textbf{1.04} & X & - \\
\hline
\parbox[c][2.5em]{1.0cm}{\centering fr2\_desk\\ \_person} & \textbf{0.63} & X & 31.73 & 6.97 (2.00) \\
\hline
\parbox[c][2.5em]{1.0cm}{\centering fr3\_sit\_\\ xyz} & \textbf{0.79} & 0.83 & 7.73 & - \\
\hline
\parbox[c][2.5em]{1.0cm}{\centering fr3\_sit\_\\ \_halfsph} & \textbf{1.34} & X & 5.87 & - \\
\hline
\parbox[c][2.5em]{1.0cm}{\centering fr3\_walk\\ \_xyz} & \textbf{1.24} & X & 12.44 & -\\
\hline
\parbox[c][2.5em]{1.0cm}{\centering fr3\_walk\\ \_halfsph} & \textbf{1.74} & X & X & - \\
\hline
\end{tabular}
\end{center}
\raggedright
Results for ORB-SLAM, PTAM and LSD-SLAM are the median over 5 executions in each sequence. The
trajectories have been aligned with 7DoF with the ground truth.
Trajectories for RGBD-SLAM are taken from the benchmark website, only available for fr1 and fr2 sequences, and
have been aligned with 6DoF and 7DoF (results between brackets). 
X means that the tracking is lost at some point and a significant portion of the sequence is not processed by the system.
\label{tb:tum}
 
\end{table}

\subsection{Relocalization in the TUM RGB-D Benchmark}

\begin{figure}[t]
    \includegraphics[width=0.47\textwidth]{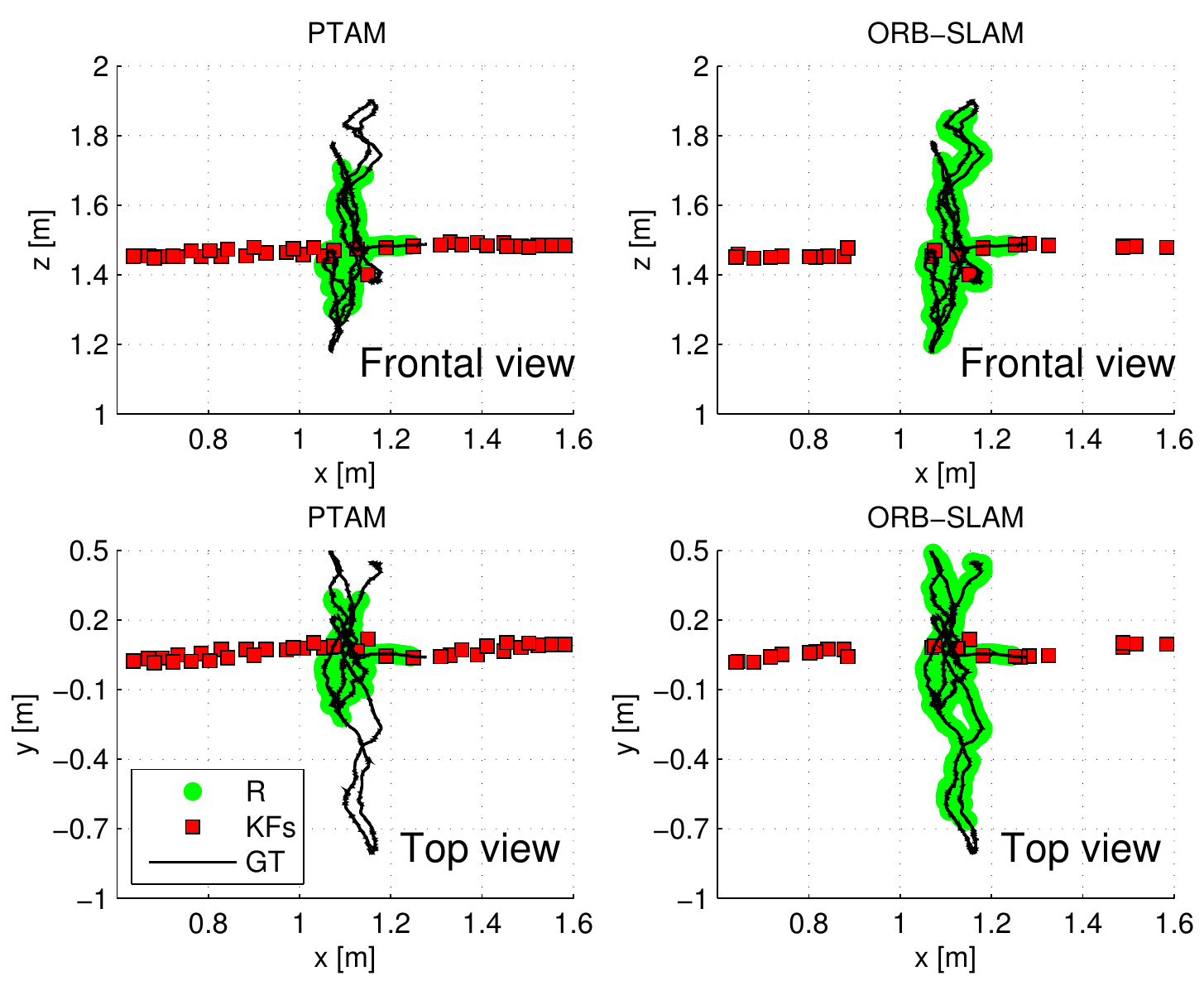}
    \caption{Relocalization experiment in \emph{fr2\_xyz}. Map is initially created during the first 30 seconds of the sequence (KFs). The goal is to relocalize subsequent frames.
     Successful relocalizations (R) of our system and PTAM are shown. The ground truth (GT) is only shown for the frames to relocalize.}
    \label{fig:reloc1}
\end{figure}

\begin{figure}[t]
\centering
        \subfigure
        {
    \includegraphics[width=0.47\textwidth]{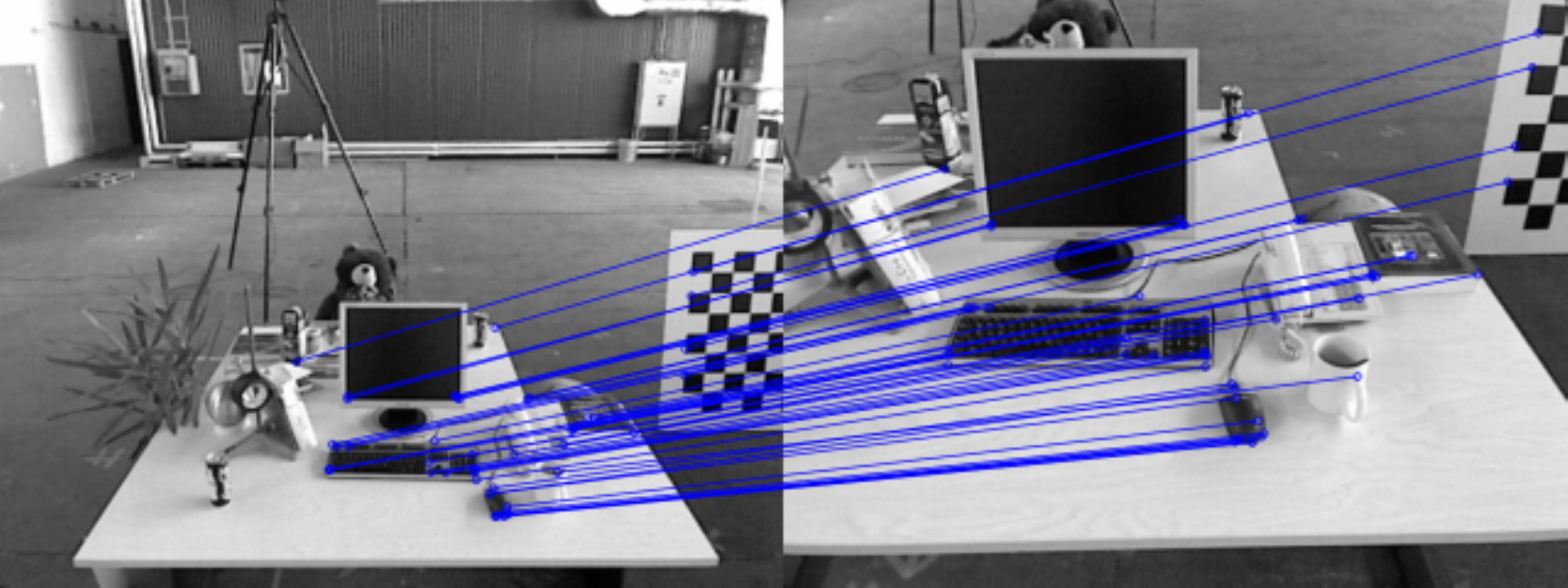}
    }
    \subfigure
        {
    \includegraphics[width=0.47\textwidth]{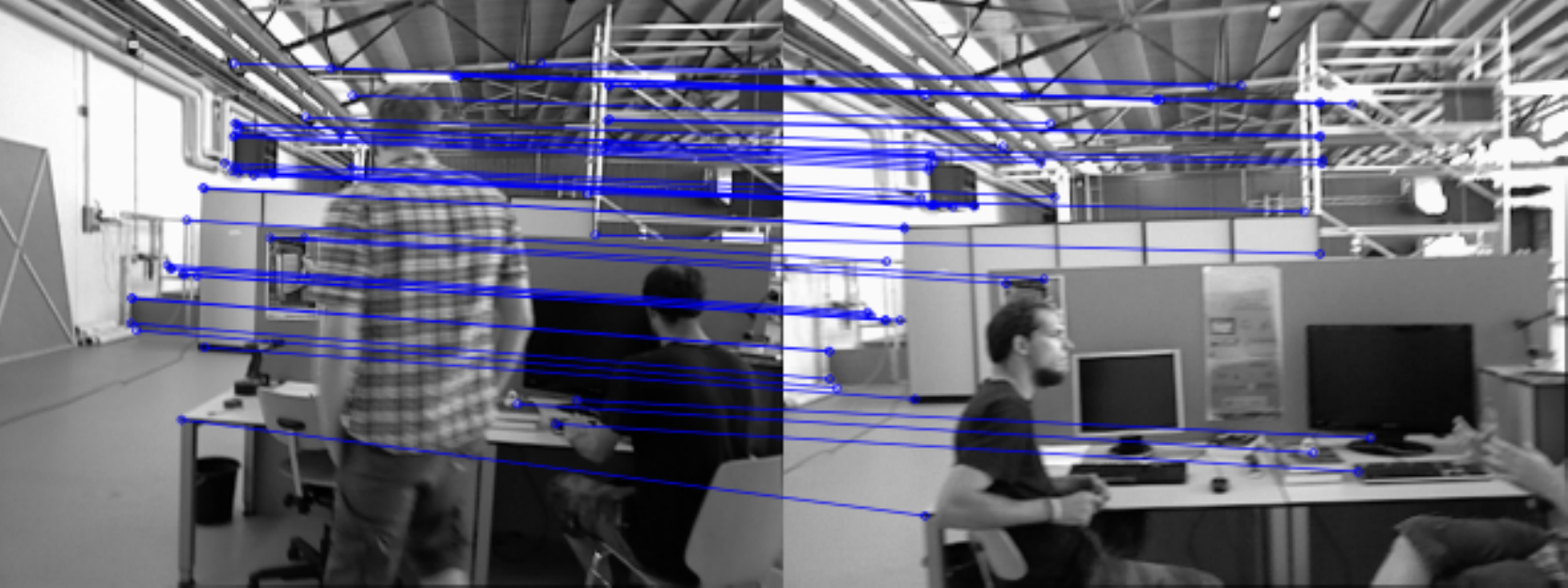}
    }
    
    \caption{Example of challenging relocalizations (severe scale change, dynamic objects) that our system successfully found in the relocalization experiments.}
    \label{fig:reloc2}
 
\end{figure}

We perform two relocalization experiments in the TUM RGB-D benchmark. In the first experiment we build a map with the first 30 seconds
of the sequence \emph{fr2\_xyz} and perform global relocalization with every successive frame and evaluate the accuracy of the recovered poses.
We perform the same experiment with PTAM for comparison. Fig. \ref{fig:reloc1} shows the keyframes used to create the initial map, the 
poses of the relocalized frames and the ground truth for those frames. It can be seen that PTAM is only able to relocalize frames which are near to the 
keyframes due to the little invariance of its relocalization method. Table \ref{tb:res:reloc1} shows the recall and the error with respect to the ground truth. 
ORB-SLAM accurately relocalizes more than the double of frames than PTAM.
In the second experiment we create an initial map with sequence \emph{fr3\_sitting\_xyz} and try to relocalize all frames from \emph{fr3\_walking\_xyz}. This is a challenging
experiment as there are big occlusions due to people moving in the scene. Here PTAM finds no relocalizations while our system relocalizes 78\% of the frames,
as can be seen in Table \ref{tb:res:reloc1}. Fig. \ref{fig:reloc2} shows some examples of challenging relocalizations performed by our system in these experiments.

\begin{table} [t]
 \caption{Results for the relocalization experiments}
 \label{tb:res:reloc1}
\begin{center}
 \begin{tabular}{|c|c|c|c|c|c|}
 \cline{2-6} 
   \multicolumn{1}{c}{} & \multicolumn{2}{|c|}{} & \multicolumn{3}{c|}{} \\[-0.8em]
 \multicolumn{1}{c}{} & \multicolumn{2}{|c|}{Initial Map} & \multicolumn{3}{c|}{Relocalization} \\[0.2em]
    \hline
    &&&&& \\[-0.8em]
  System & KFs &  \parbox[c][2em]{0.75cm}{\centering  RMSE\\(cm)} & \parbox[c][2em]{0.75cm}{\centering Recall \\ (\%)} & 
  \parbox[c][2em]{0.75cm}{\centering  RMSE\\(cm)} & \parbox[c][2em]{1.5cm}{\centering  Max. Error \\ (cm)} \\[0.8em]
  \hline
  \multicolumn{6}{|c|}{} \\[-0.8em]
 \multicolumn{6}{|c|}{\emph{fr2\_xyz}. 2769 frames to relocalize}\\[0.2em]
 \hline
 &&&&& \\[-0.8em]
 PTAM & 37 & 0.19 & 34.9 & 0.26 & 1.52  \\[0.2em]
 \hline
 &&&&& \\[-0.8em]
 ORB-SLAM & 24 & 0.19 &  \textbf{78.4} & 0.38 & 1.67  \\[0.2em]
 \hline

  \multicolumn{6}{|c|}{} \\[-0.8em]
 \multicolumn{6}{|c|}{\emph{fr3\_walking\_xyz}. 859 frames to relocalize}\\[0.2em]
 
 \hline 
 &&&&&\\[-0.8em]
    PTAM & 34 & 0.83 & 0.0 & - & - \\[0.2em]
 \hline 
  &&&&& \\[-0.8em]
 ORB-SLAM & 31 & 0.82 &  \textbf{77.9} & 1.32 & 4.95  \\[0.2em]
  \hline
  
 \end{tabular} 
\end{center}
\end{table}

\subsection{Lifelong Experiment in the TUM RGB-D Benchmark}

\begin{figure}[t]
    \includegraphics[width=0.47\textwidth]{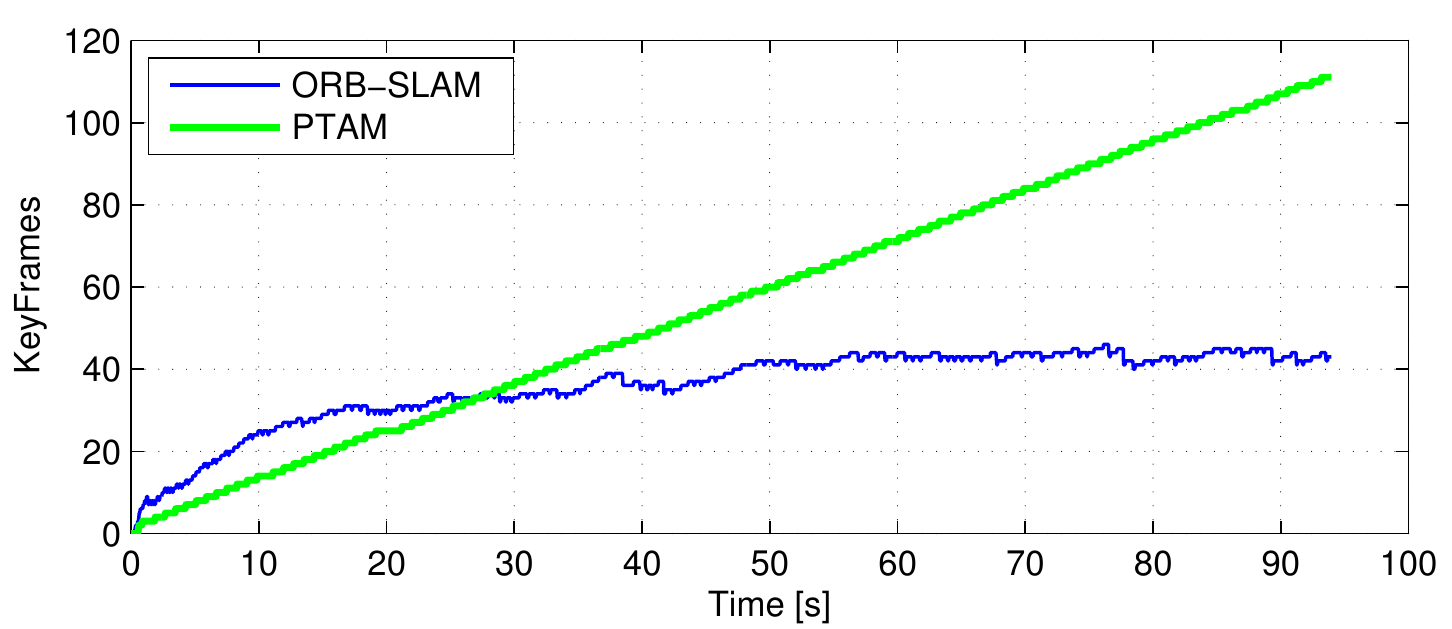}
    \caption{Lifelong experiment in a static environment where the camera is always looking at the same place from different viewpoints. 
    PTAM is always inserting keyframes, while ORB-SLAM is able to prune redundant keyframes and maintains a bounded-size map.}
    \label{fig:lifelong}
\end{figure}

\begin{figure}[t]
\centering
        \subfigure[Evolution of the number of keyframes in the map]
        {
    \includegraphics[width=0.47\textwidth]{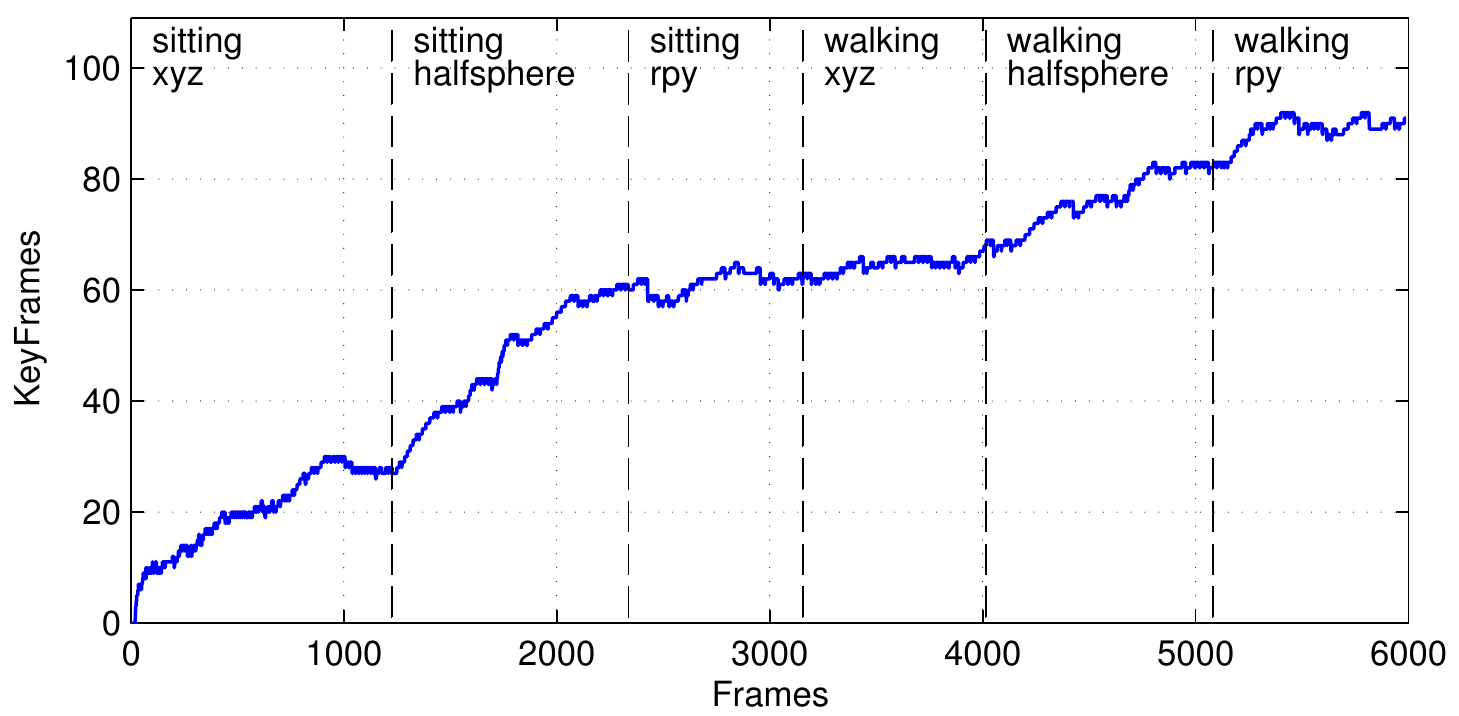}
    \label{fig:life:1}
    }

    \subfigure[Keyframe creation and destruction. Each horizontal line corresponds to a keyframe, from its creation frame until its destruction]
        {
    \includegraphics[width=0.47\textwidth]{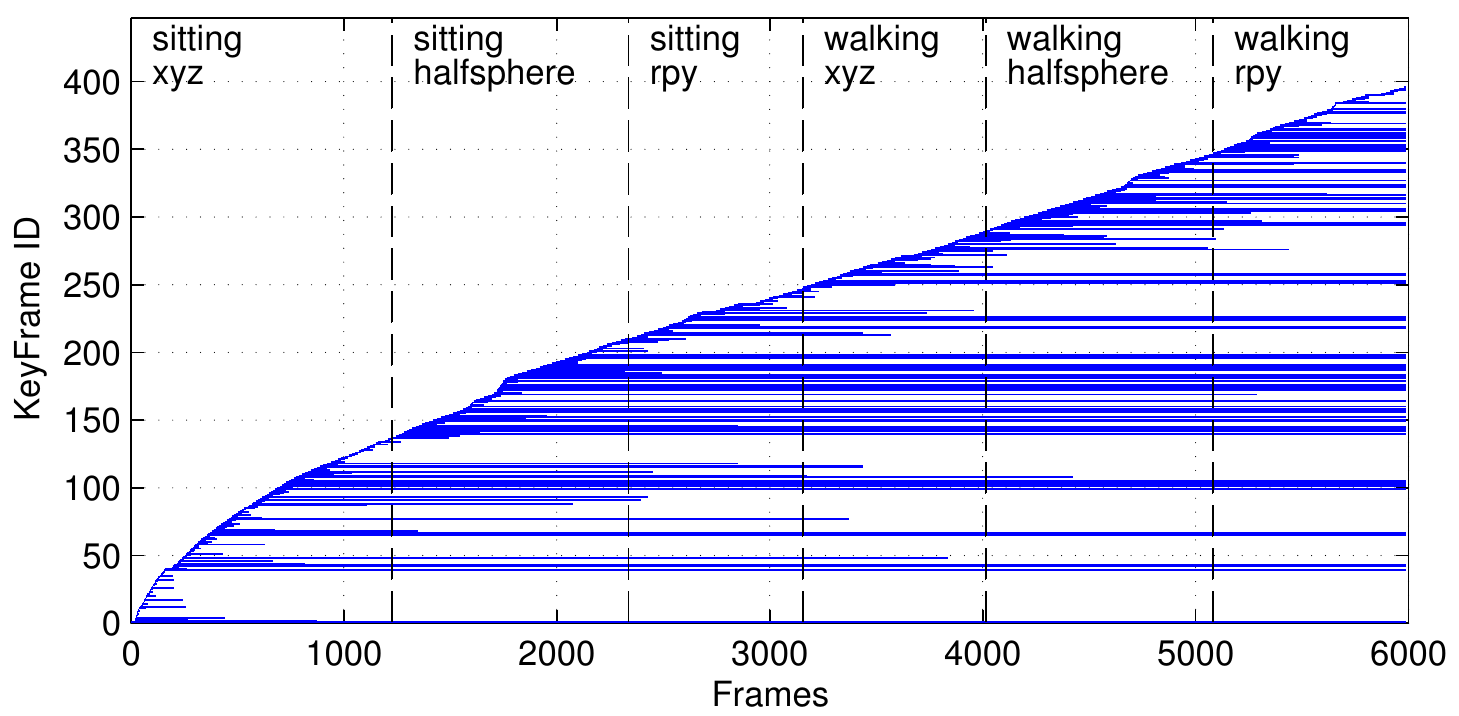}
    \label{fig:life:2}
    }
    \subfigure[Histogram of the survival time of all spawned keyframes with respect to the remaining time of the experiment]
        {
    \includegraphics[width=0.47\textwidth]{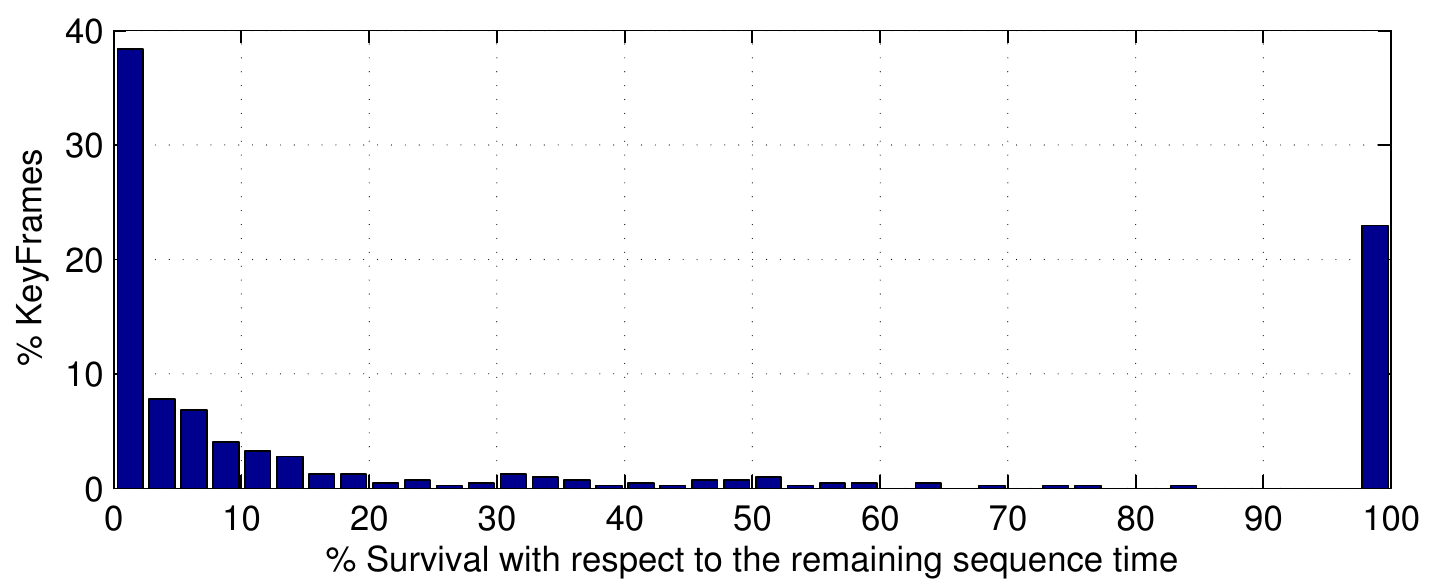}
    \label{fig:life:3}
    }
    \caption{Lifelong experiment in a dynamic environment from the TUM RGB-D Benchmark.}
    \label{fig:lifelong2}
 
\end{figure}

Previous relocalization experiments have shown that our system is able to localize in a map from very different viewpoints and robustly under moderate dynamic changes.
This property in conjunction with our keyframe culling procedure allows to operate lifelong in the same environment under different viewpoints and
some dynamic changes. 

In the case of a completely static scenario our system is able to maintain the number of keyframes bounded even if the camera is looking
at the scene from different viewpoints. We demonstrate it in a custom sequence were the camera is looking at the same desk during 93 seconds but performing
a trajectory so that the viewpoint is always changing. We compare the evolution of the number of keyframes in our map and those generated by PTAM in Fig. \ref{fig:lifelong}.
It can be seen how PTAM is always inserting keyframes, while our mechanism to prune redundant keyframes makes its number to saturate. 

While the lifelong operation in a static scenario should be a requirement of any SLAM system, more interesting is the case where dynamic
changes occur. We analyze the behavior of our system in such scenario by running consecutively the dynamic sequences from \emph{fr3}: \emph{sitting\_xyz}, \emph{sitting\_halfsphere}, \emph{sitting\_rpy}, 
\emph{walking\_xyz}, \emph{walking\_halfspehere} and \emph{walking\_rpy}. All the sequences focus the camera to the same desk but perform different
trajectories, while people are moving and change some objects like chairs. Fig. \ref{fig:life:1} shows the evolution of the total number of keyframes in the map, and
Fig. \ref{fig:life:2} shows for each keyframe its frame of creation and destruction, showing how long the keyframes have survived in the map.
It can be seen that during the first two sequences the map size grows as all the views of the scene are being seen for the first time. In Fig. \ref{fig:life:2} we can 
see that several keyframes created during these two first sequences are maintained in the map during the whole experiment. During the sequences \emph{sitting\_rpy} and 
\emph{walking\_xyz} the map does not grow, because the map created so far explains well the scene. In contrast, during the last two sequences, more keyframes are inserted
showing that there are some novelties in the scene that were not yet represented, due probably to dynamic changes. Finally Fig. \ref{fig:life:3} shows a histogram of the 
keyframes according to the time they have survived with respect to the remaining time of the sequence from its moment of creation. It can be seen that most of the keyframes
are destroyed by the culling procedure soon after creation, and only a small subset survive until the end of the experiment. On one hand, this shows that our system
has a generous keyframe spawning policy, which is very useful when performing abrupt motions in exploration. On the other hand the system is eventually able
to select a small representative subset of those keyframes.

In these lifelong experiments we have shown that our map grows with the content of the scene but not with the time, and that is able to store the dynamic changes
of the scene which could be useful to perform some scene understanding by accumulating experience in an environment.

\subsection{Large Scale and Large Loop Closing in the KITTI Dataset}\label{sec:exp:kitti}

\begin{figure*}[p]
\centering
    \subfigure
    {
    \includegraphics[width=0.31\textwidth]{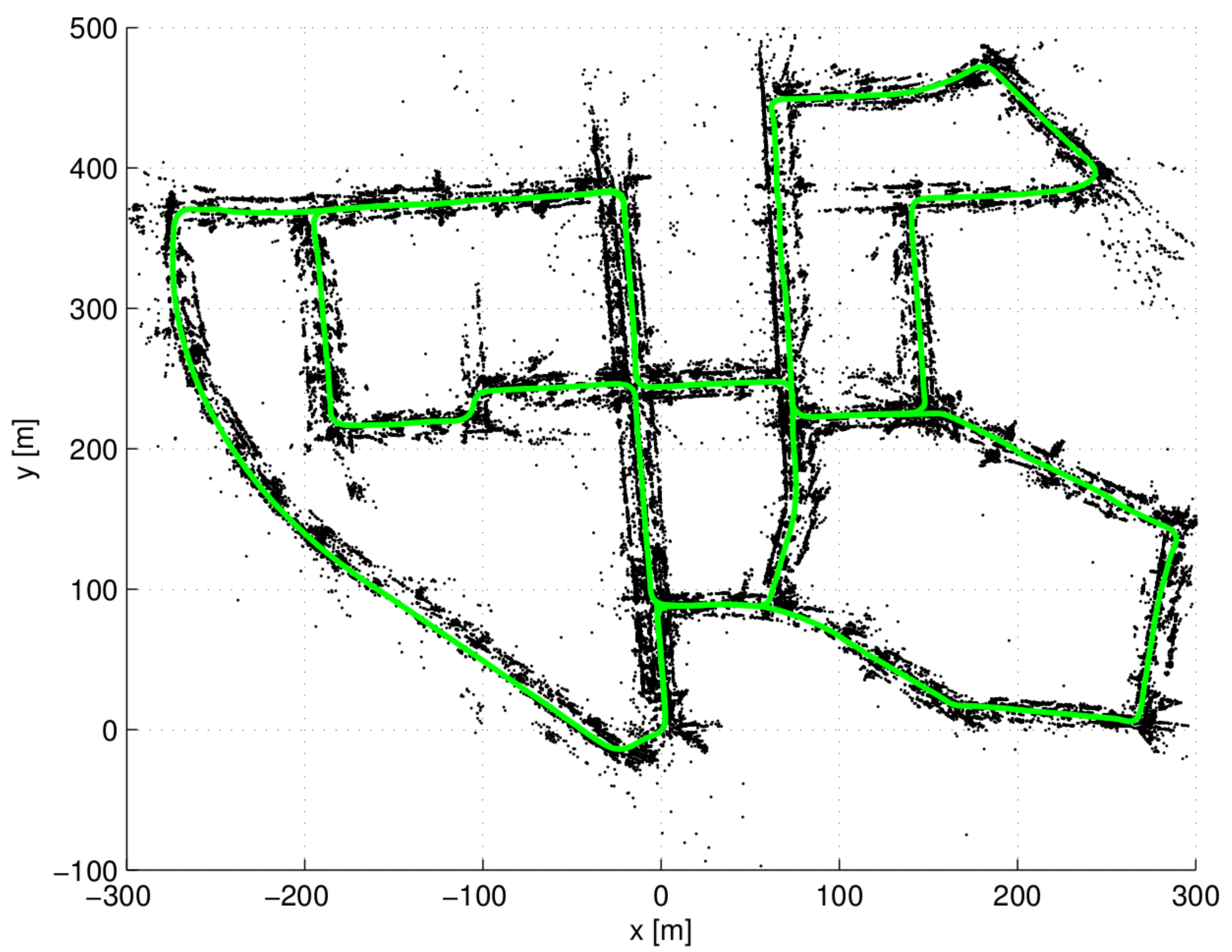}
    }
    \subfigure
    {
    \includegraphics[width=0.31\textwidth]{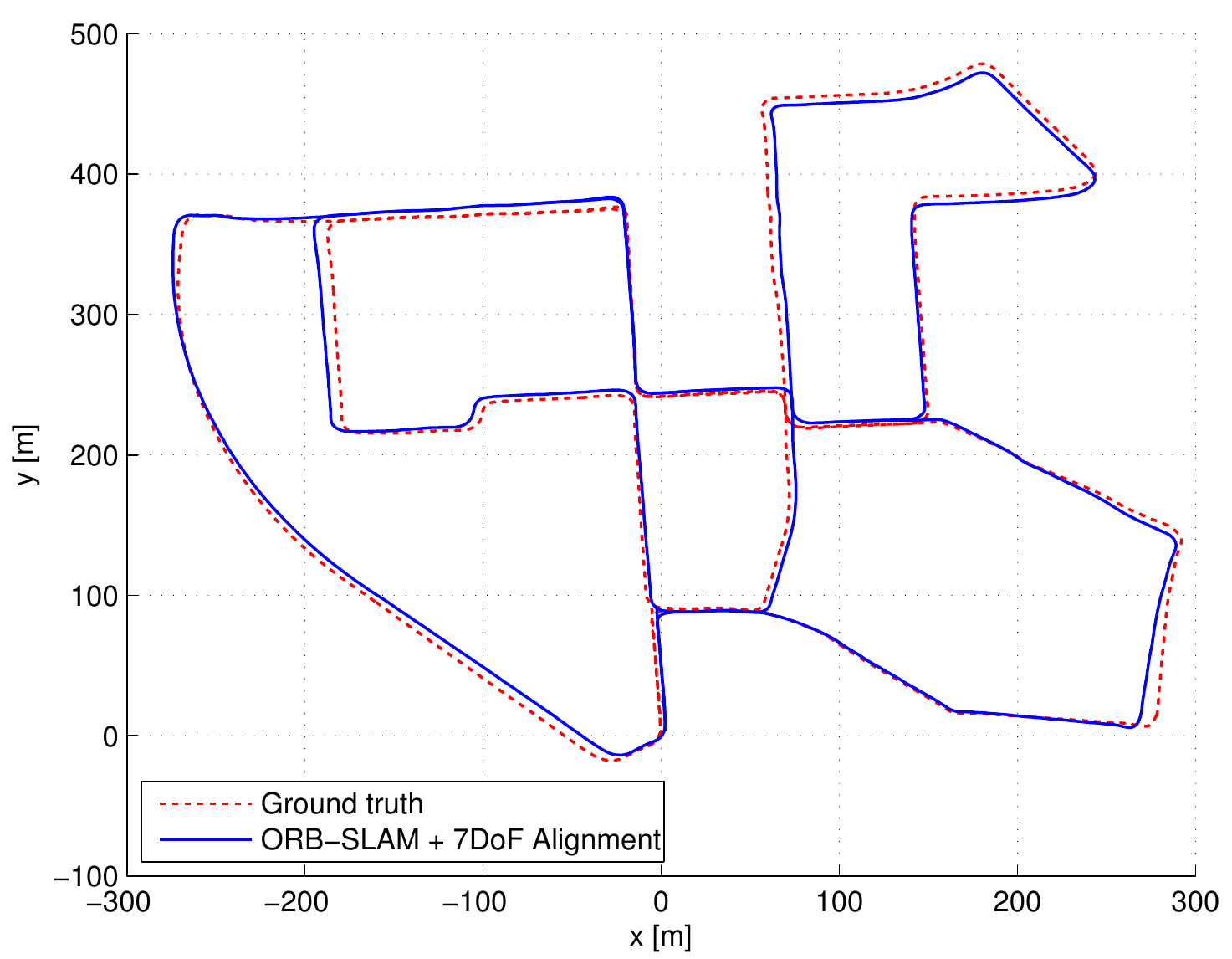}
    }    
    \subfigure
    {
    \includegraphics[width=0.31\textwidth]{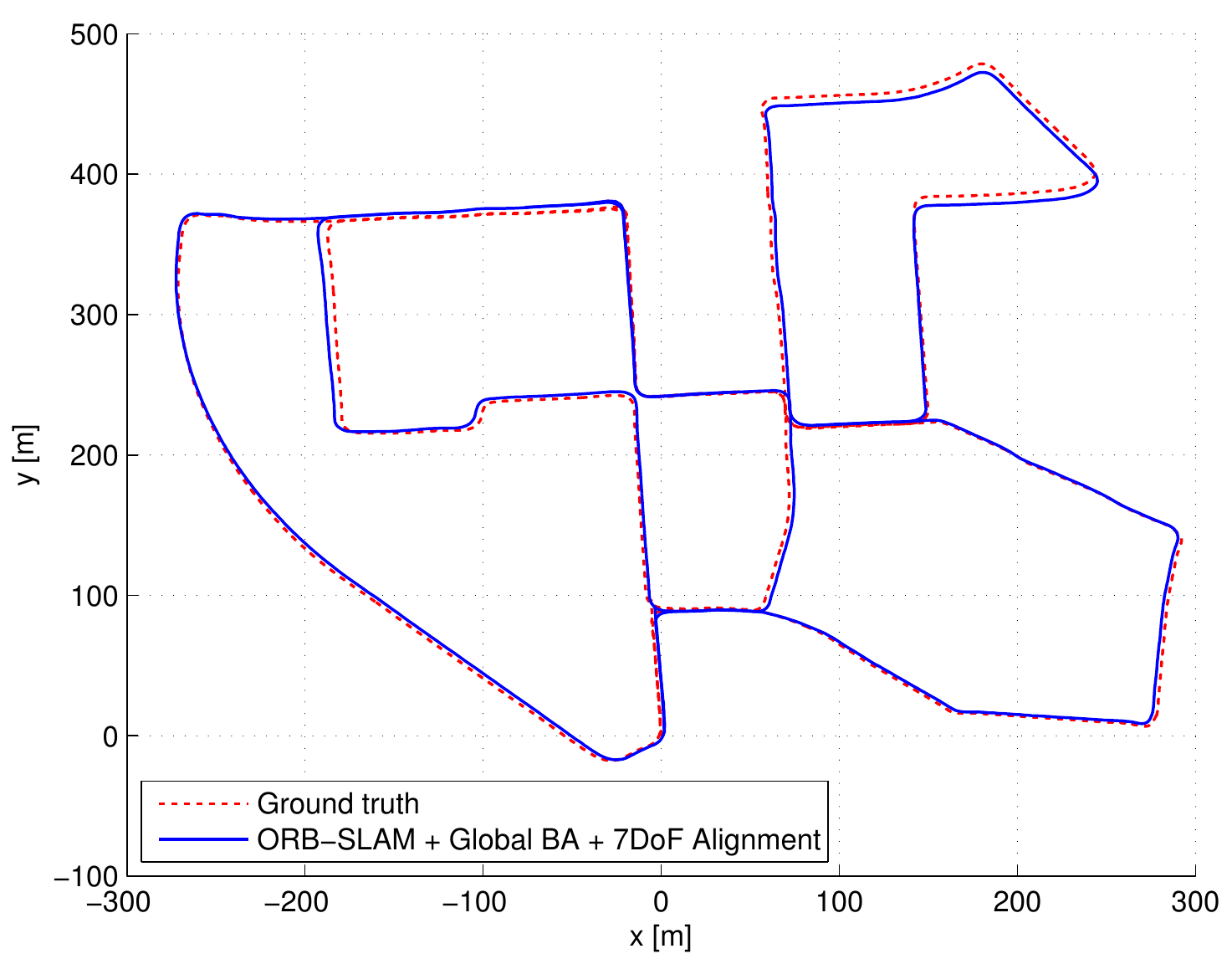}
    }
    
    \subfigure
    {
    \includegraphics[width=0.31\textwidth]{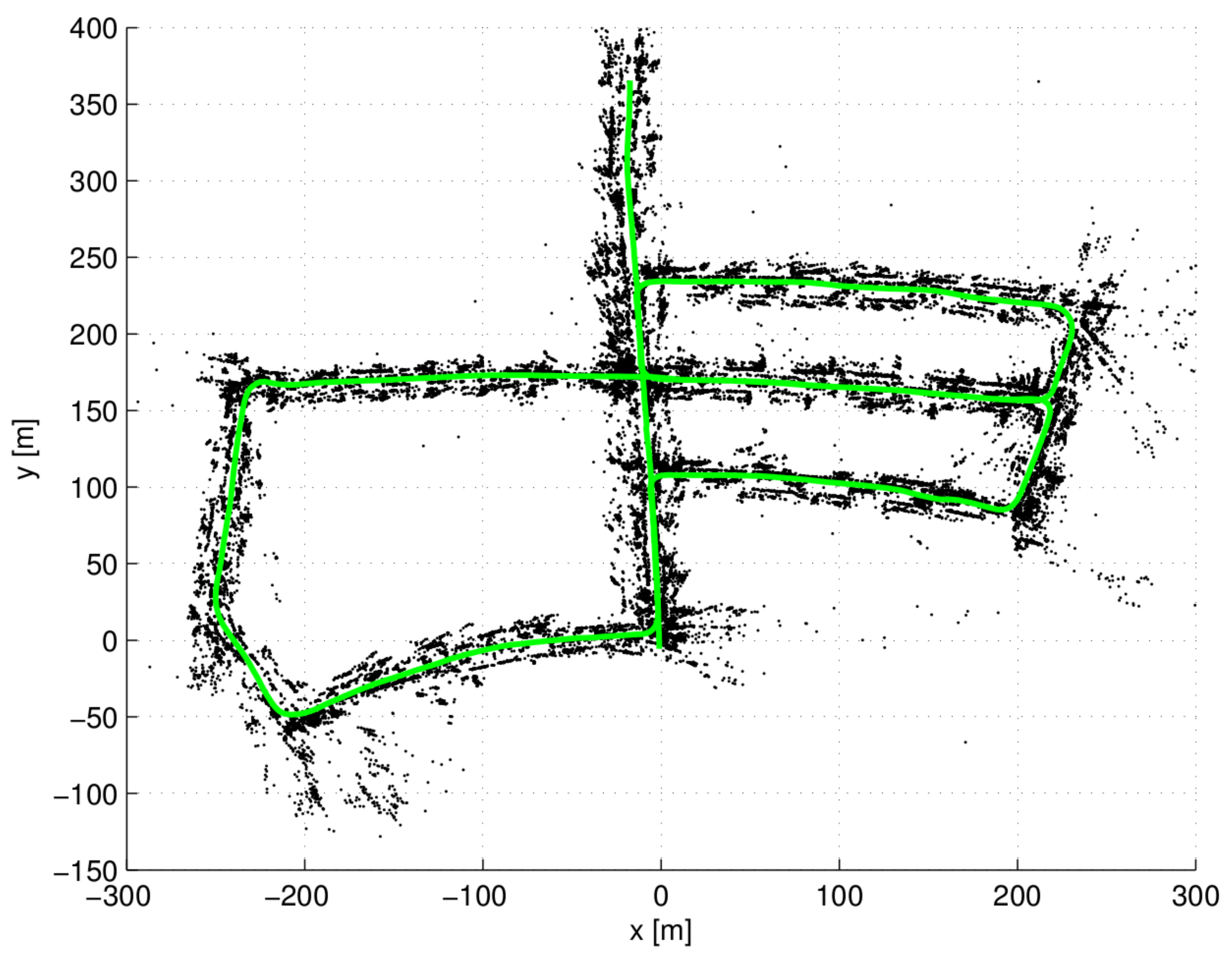}
    }
    \subfigure
    {
    \includegraphics[width=0.31\textwidth]{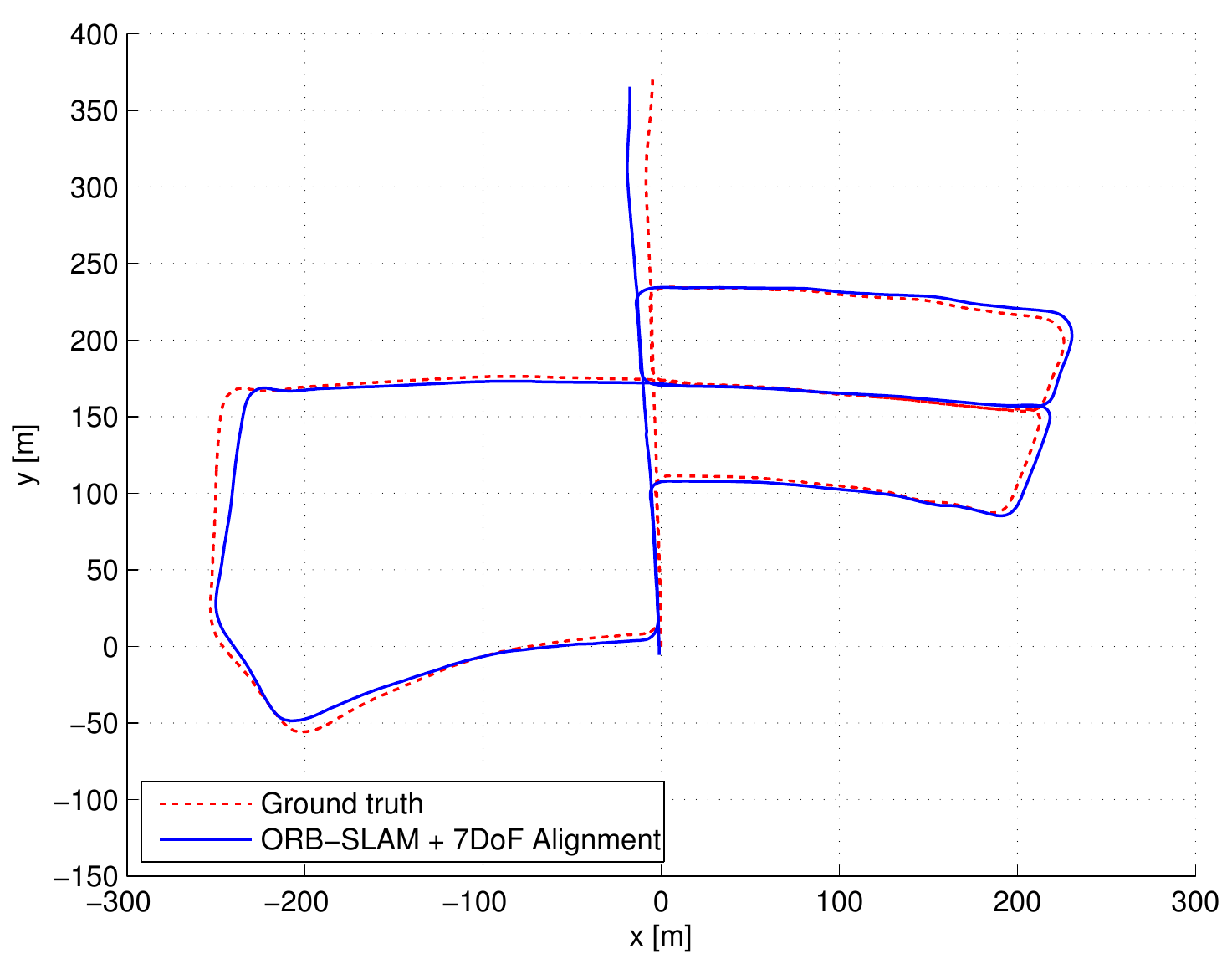}
    }    
    \subfigure
    {
    \includegraphics[width=0.31\textwidth]{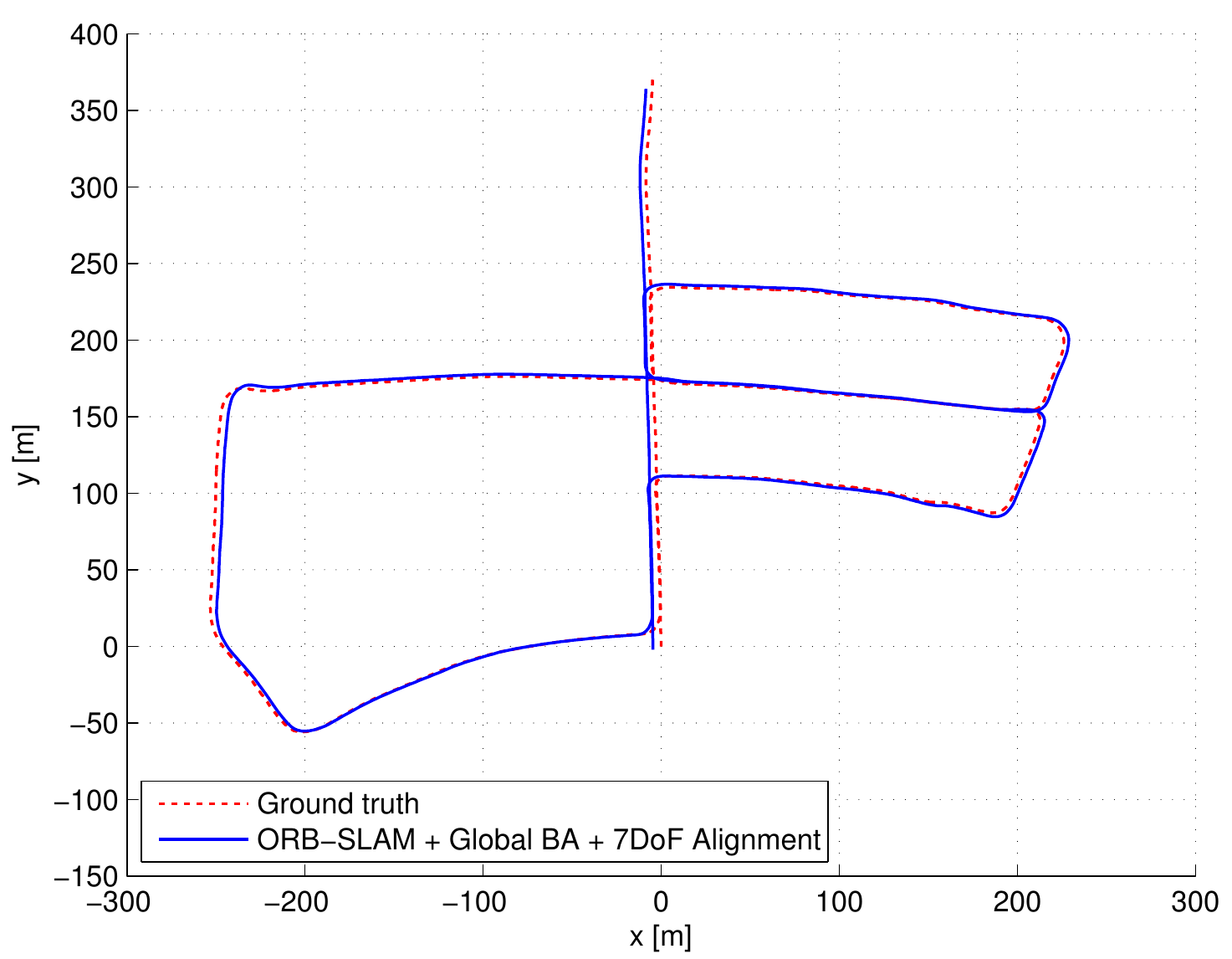}
    }
    
    \subfigure
    {
    \includegraphics[width=0.31\textwidth]{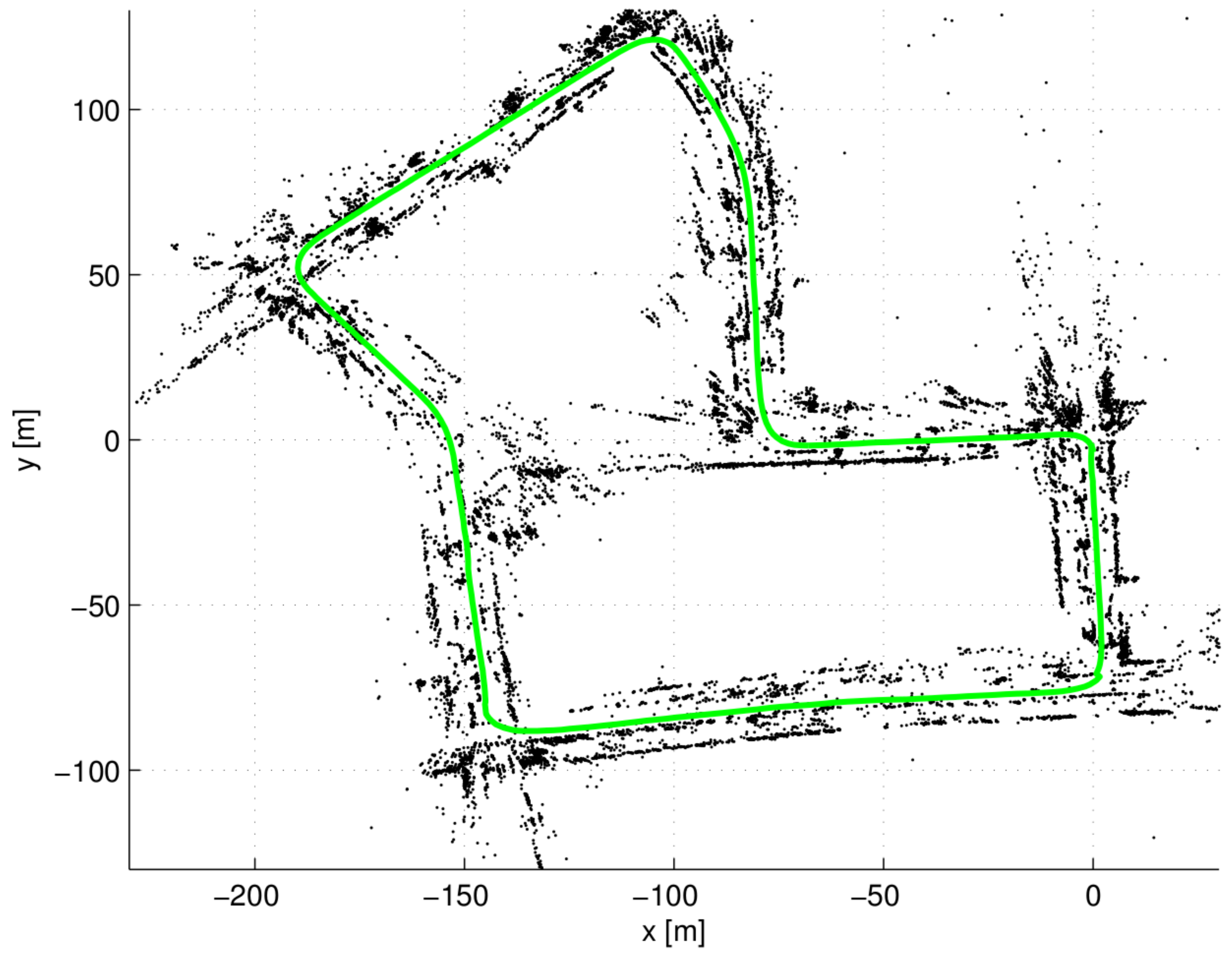}
    }
    \subfigure
    {
    \includegraphics[width=0.31\textwidth]{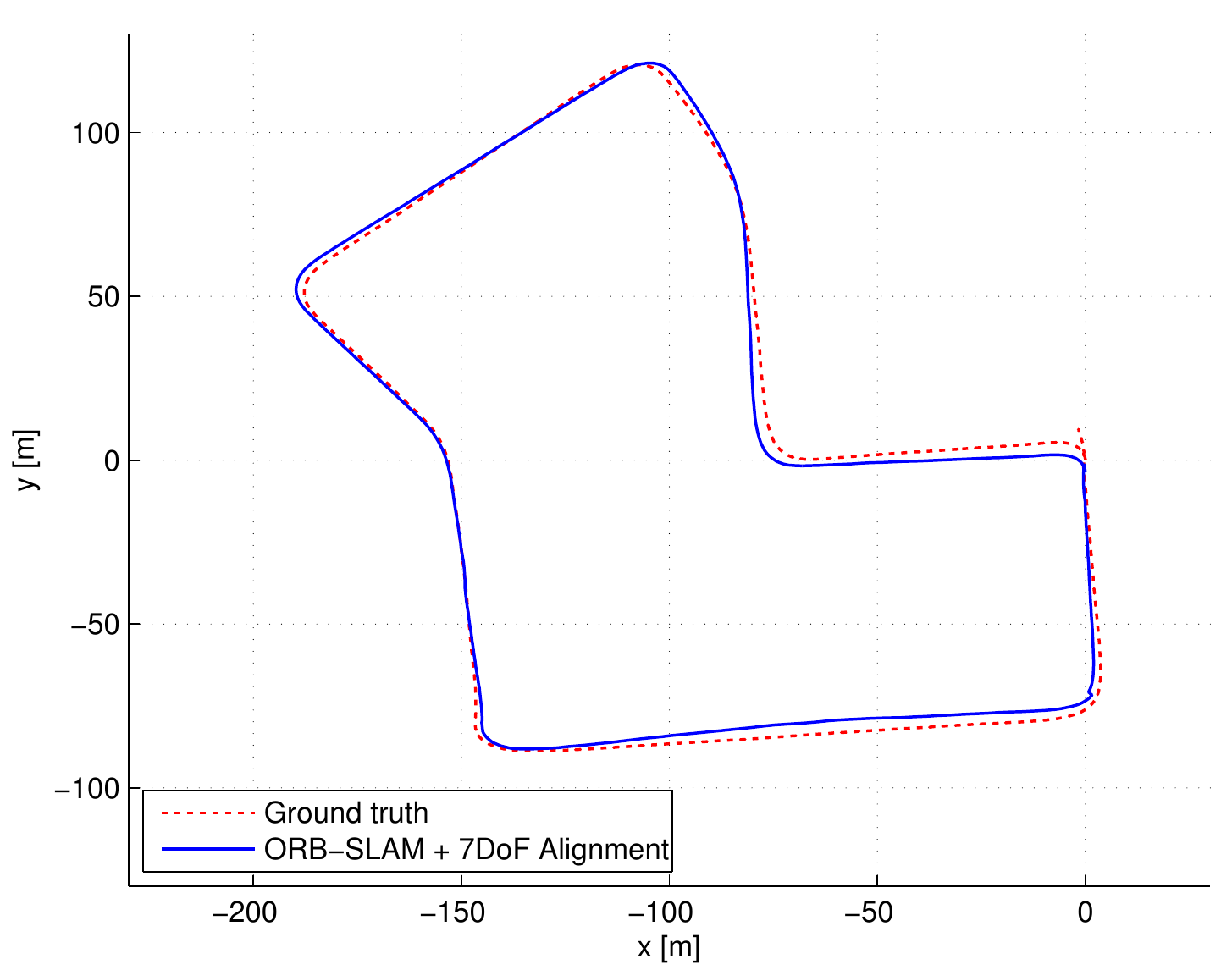}
    }    
    \subfigure
    {
    \includegraphics[width=0.31\textwidth]{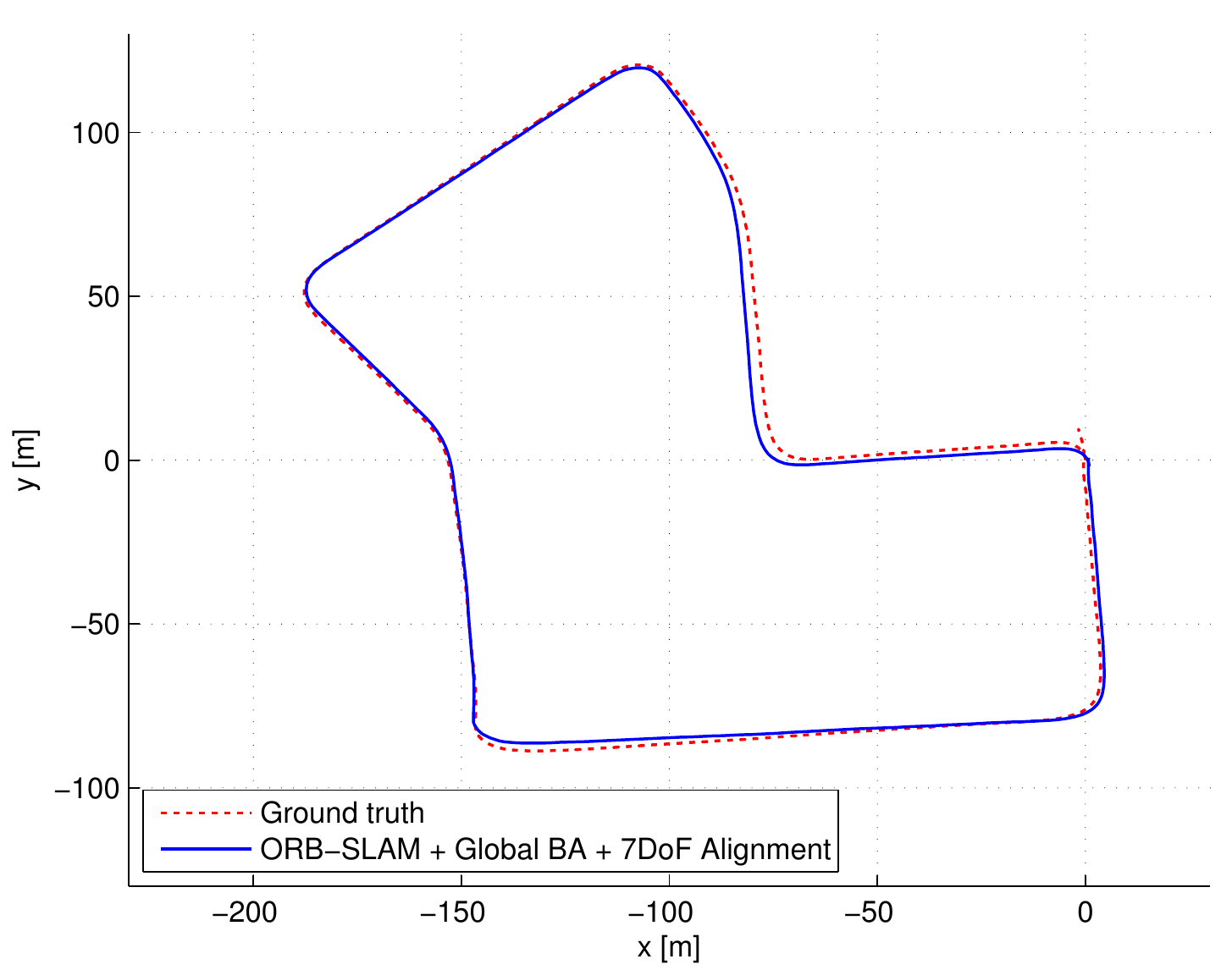}
    }
   
    \caption{Sequences 00, 05 and 07 from the odometry benchmark of the KITTI dataset.
    Left: points and keyframe trajectory. Center: trajectory and ground truth. Right: trajectory after 20 iterations of full BA. The output of our system is quite accurate,
    while it can be slightly improved with some iterations of BA.}
\label{fig:kitti1} 

\end{figure*}
\begin{figure*}[p]
\centering
    
    \subfigure[Sequence 02]
    {
    \includegraphics[width=0.22\textwidth]{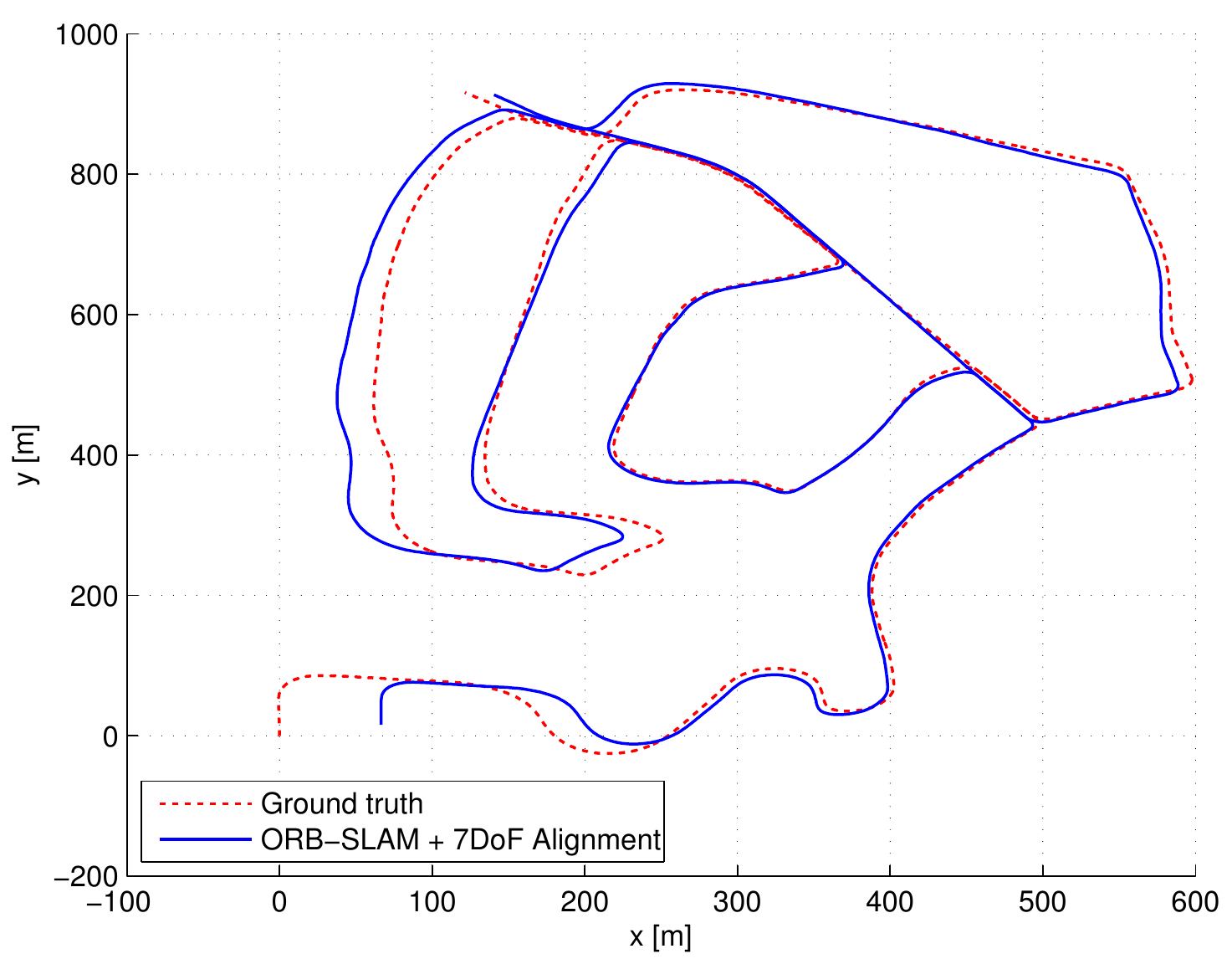}
    }    
    \subfigure[Sequence 03]
    {
    \includegraphics[width=0.22\textwidth]{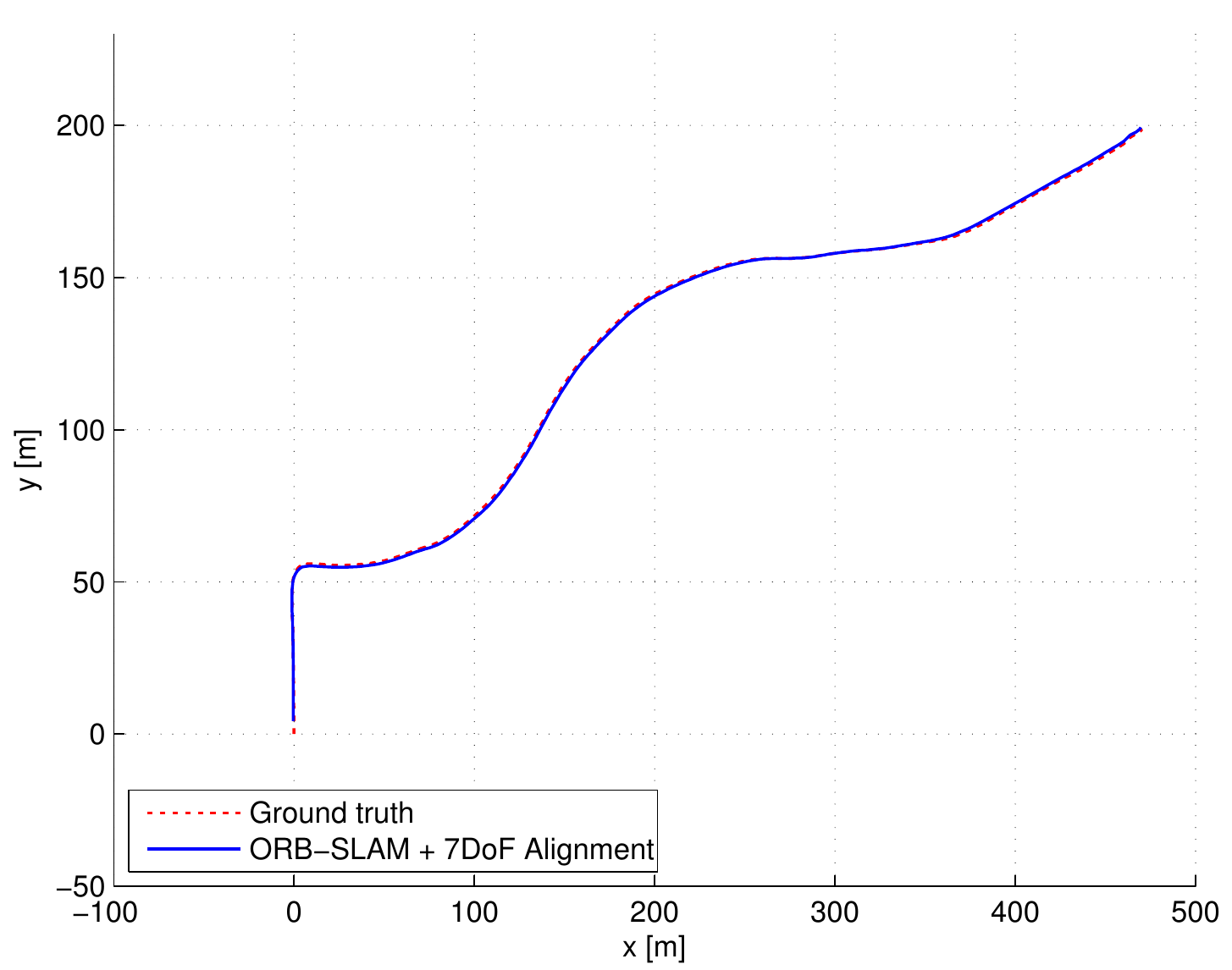}
    }
    \subfigure[Sequence 04]
    {
    \includegraphics[width=0.22\textwidth]{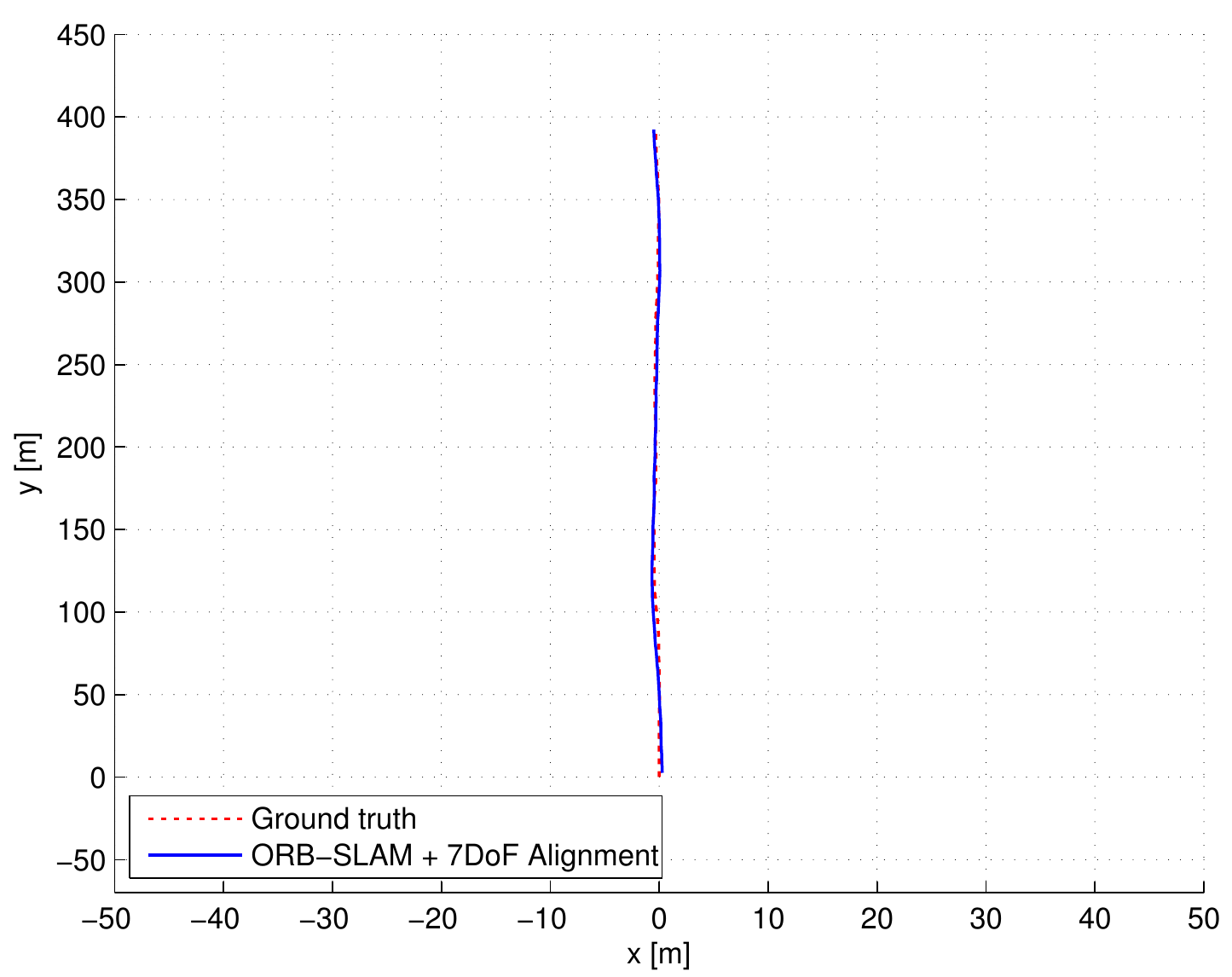}
    }

    \subfigure[Sequence 06]
    {
    \includegraphics[width=0.22\textwidth]{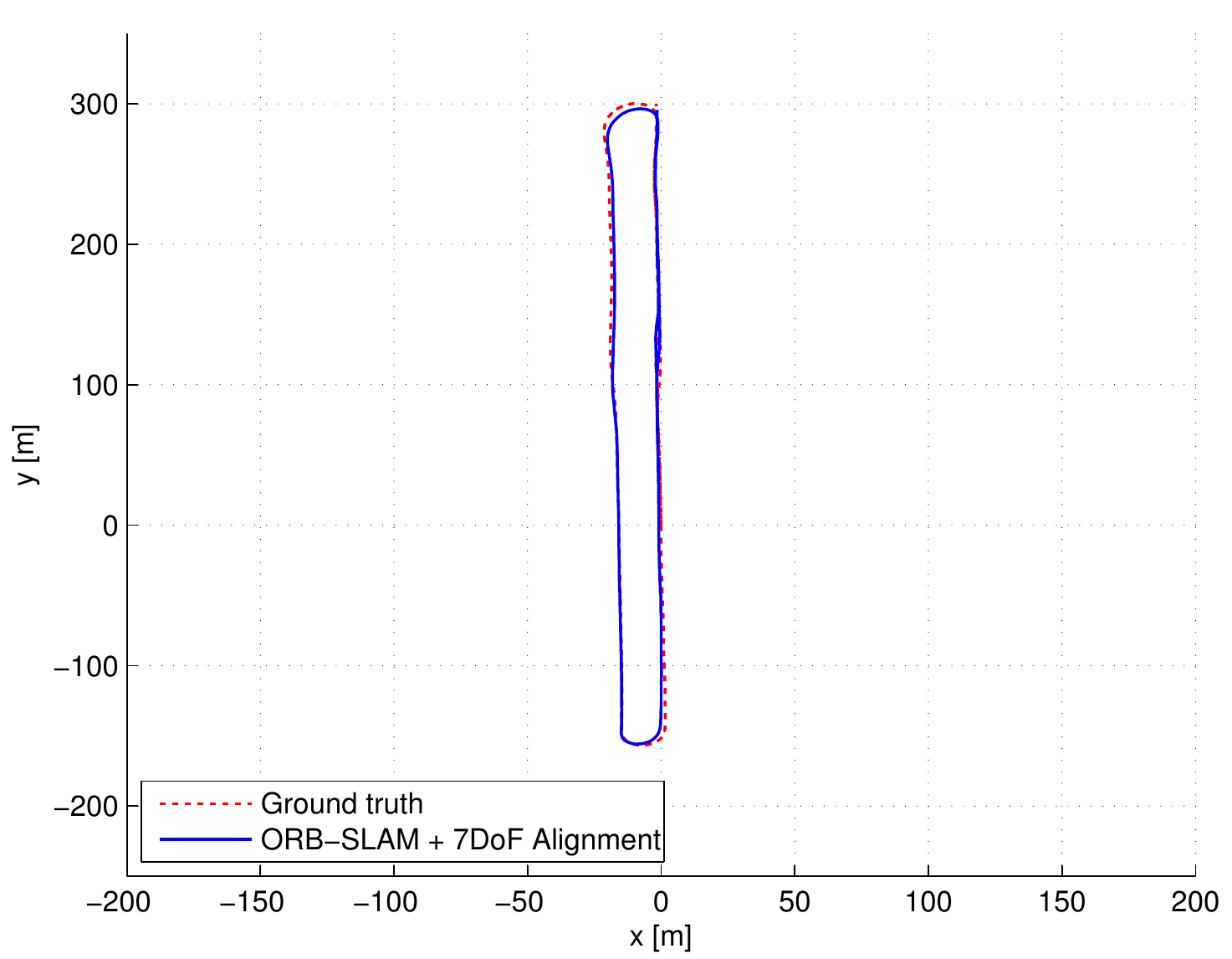}
    } 
    \subfigure[Sequence 08]
    {
    \includegraphics[width=0.22\textwidth]{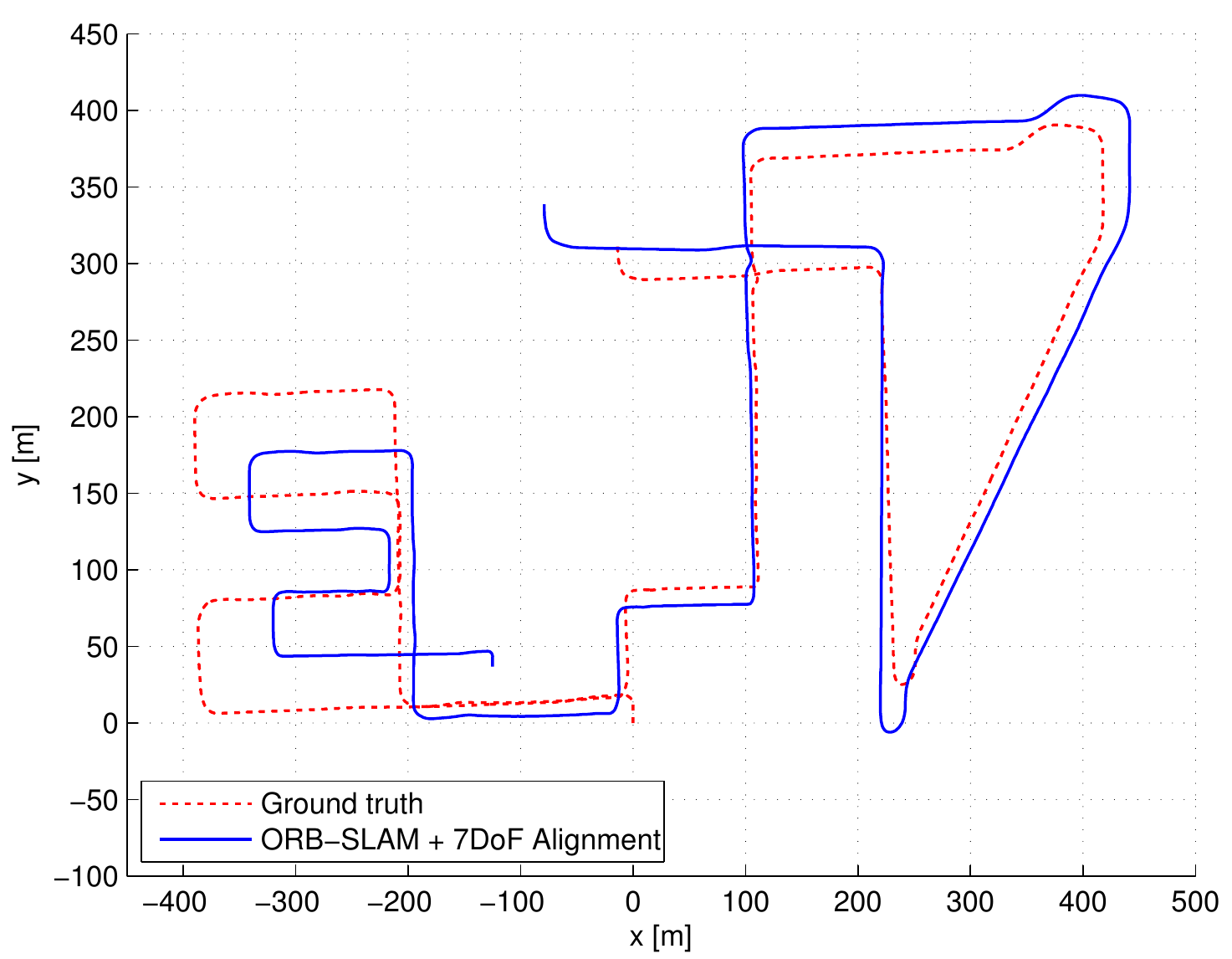}
    }
    \subfigure[Sequence 09]
    {
    \includegraphics[width=0.22\textwidth]{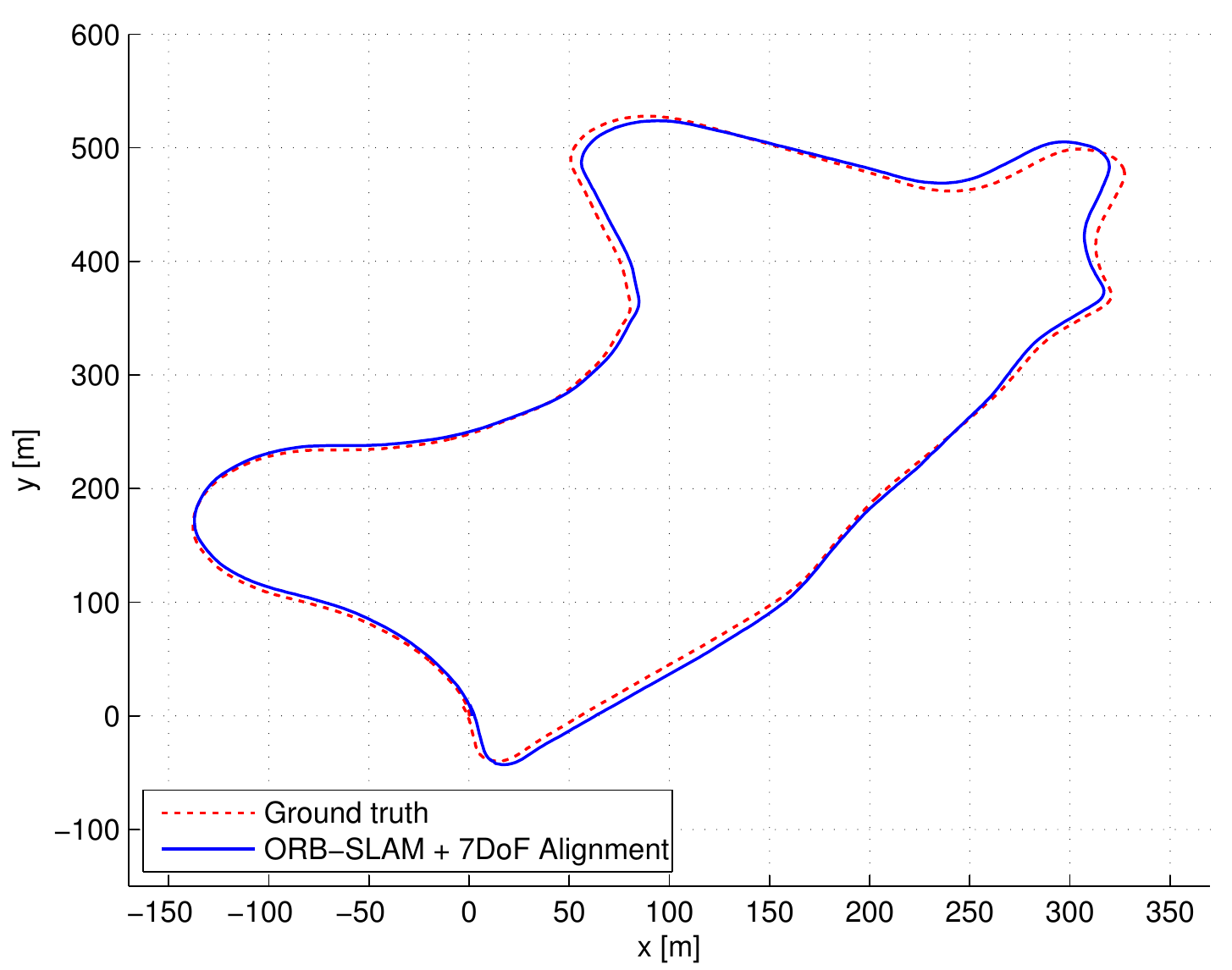}
    }
    \subfigure[Sequence 10]
    {
    \includegraphics[width=0.22\textwidth]{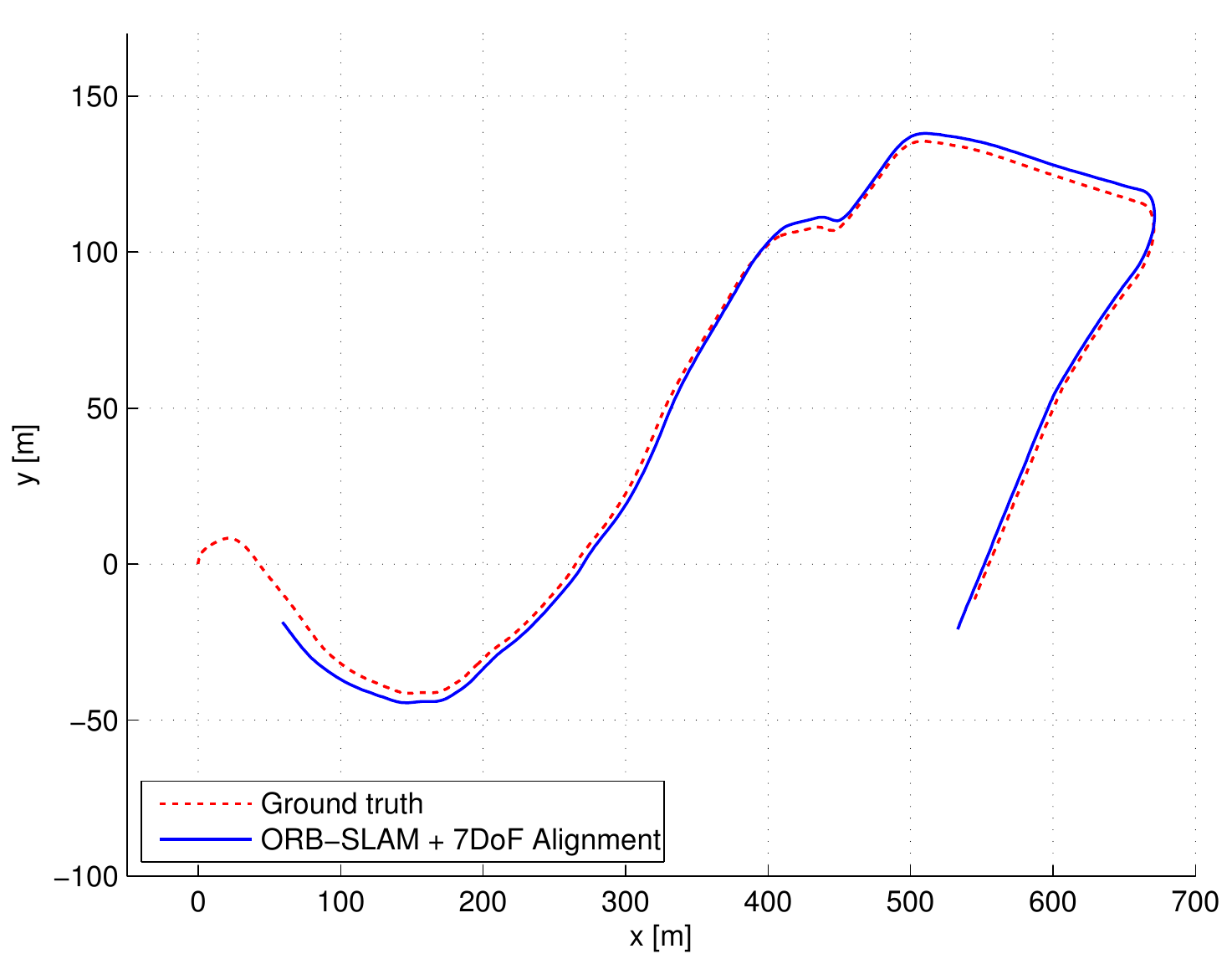}
    }
    
    \caption{ORB-SLAM keyframe trajectories in sequences 02, 03, 04 ,06, 08, 09 and 10 from the odometry benchmark of the KITTI dataset.
    Sequence 08 does not contains loops and drift (especially scale) is not corrected.} 
\label{fig:kitti2}
\end{figure*}

The odometry benchmark from the KITTI dataset \cite{kitti} contains 11 sequences from a car driven around a residential area with
accurate ground truth from GPS and a Velodyne laser scanner. This is a very challenging dataset for monocular vision due to fast rotations,
areas with lot of foliage, which make more difficult data association, and relatively high car speed, being the sequences recorded at 10 fps.
We play the sequences at the real frame-rate they were recorded and ORB-SLAM is able to process
all the sequences by the exception of sequence 01 which is a highway with few trackable close objects.
Sequences 00, 02, 05, 06, 07, 09 contain loops that were correctly detected and closed by our system. Sequence 09 contains a loop that can be detected only
in a few frames at the end of the sequence, and our system not always detects it (the results provided are for the executions in which it was detected).

Qualitative comparisons of our trajectories and the ground truth are shown in Fig. \ref{fig:kitti1} and Fig. \ref{fig:kitti2}.
As in the TUM RGB-D benchmark we have aligned the keyframe trajectories of our system and the ground truth with a similarity transformation.
We can compare qualitatively our results from Fig. \ref{fig:kitti1} and Fig. \ref{fig:kitti2}  
with the results provided for sequences 00, 05, 06, 07 and 08 by the recent monocular SLAM approach of Lim et. al \cite{icra14BRIEF} 
in their figure 10. ORB-SLAM produces clearly more accurate trajectories for all those sequences by the exception of sequence 08 in which they seem to suffer less 
drift.

Table \ref{tb:kitti}
shows the median RMSE error of the keyframe trajectory over five executions in each sequence. 
We also provide the dimensions of the maps to put in context the errors. 
The results demonstrate that our system is very accurate being the trajectory error typically around the 1\% of its dimensions, sometimes less as in sequence 03 
with an error of the 0.3\% or higher as in sequence 08 with the 5\%. In sequence 08 there are no loops and drift cannot be corrected, which makes clear the need
of loop closures to achieve accurate reconstructions.

\begin{table}[t]
\caption{Results of our system in the KITTI dataset.}

\begin{center}
\begin{tabular}{|c|c|c|c|c|c|}
 \multicolumn{6}{c}{} \\
\cline{3-6}
 \multicolumn{2}{c|}{} & \multicolumn{2}{c|}{} & \multicolumn{2}{c|}{}    \\[-0.8em]
 \multicolumn{2}{c|}{} & \multicolumn{2}{c|}{ORB-SLAM} & \multicolumn{2}{c|}{+ Global BA (20 its.)} \\[0.2em]
\hline
&&&&&\\[-0.8em]
Sequence & \parbox[c][2em]{1.3cm}{\centering Dimension \\ (m$\times$m)} & KFs & \parbox[c][2em]{1cm}{\centering RMSE\\(m)}  & \parbox[c][2em]{1cm}{ \centering RMSE\\(m)}  &
\parbox[c][2em]{1.1cm}{\centering Time BA \\ (s)}  \\[0.8em]
\hline
\hline
&&&&& \\[-0.8em]
KITTI 00 & $564 \times 496$ & 1391 & 6.68  & 5.33 &  24.83  \\[0.2em]
\hline
&&&&& \\[-0.8em]
KITTI 01 & $1157 \times 1827$ & X & X  & X &  X  \\[0.2em]
\hline
&&&&& \\[-0.8em]
KITTI 02 & $599 \times 946$ &1801 & 21.75  & 21.28 &  30.07   \\[0.2em]
\hline
&&&&& \\[-0.8em]
KITTI 03 & $471 \times 199$ & 250 & 1.59 & 1.51 & 4.88  \\[0.2em]
\hline
&&&&& \\[-0.8em]
KITTI 04 & $0.5 \times 394$ &  108 & 1.79 & 1.62 &  1.58   \\[0.2em]
\hline
&&&&& \\[-0.8em]
KITTI 05 & $479 \times 426$ & 820 & 8.23 & 4.85 &  15.20   \\[0.2em]
\hline 
&&&&& \\[-0.8em]
KITTI 06 & $23 \times 457$ & 373 & 14.68 & 12.34 &  7.78   \\[0.2em]
\hline 
&&&&& \\[-0.8em]
KITTI 07 & $191 \times 209$ & 351 & 3.36 & 2.26 &  6.28   \\[0.2em]
\hline 
&&&&& \\[-0.8em]
KITTI 08 & $808 \times 391$ & 1473 & 46.58 & 46.68 & 25.60    \\[0.2em]
\hline 
&&&&& \\[-0.8em]
KITTI 09 & $465 \times 568$ & 653 & 7.62 & 6.62  &  11.33   \\[0.2em]
\hline 
&&&&& \\[-0.8em]
KITTI 10 & $671 \times 177$ & 411 & 8.68 & 8.80 &  7.64   \\[0.2em]
\hline 
\end{tabular}
\end{center}
\label{tb:kitti}
\end{table}

In this experiment we have also checked how much the reconstruction can be improved by performing
20 iterations of \emph{full BA}, see the Appendix for details, at the end of each sequence. 
We have noticed that some iterations of \emph{full BA} slightly improves the accuracy
in the trajectories with loops but it has negligible effect in open trajectories, which means that the output of our system is already
very accurate. In any case if the most accurate results are needed our algorithm provides
a set of matches, which define a strong camera network, and an initial guess, so that \emph{full BA} converge in few iterations.

Finally we wanted to show the efficacy of our loop closing approach and the influence of the $\theta_\mathrm{min}$ used 
to include edges in the essential graph. We have selected the sequence 09 (a very long sequence with a loop closure at the end), and in the same execution 
we have evaluated different loop closing strategies. In table \ref{tb:k9} we show the keyframe trajectory RMSE and the time spent in the optimization in different cases: 
without loop closing, if we directly apply a \emph{full BA} (20 or 100 iterations), if we apply only pose graph optimization (10 iterations with different number of edges) 
and if we apply pose graph optimization and \emph{full BA} afterwards. The results clearly show that before loop closure, the solution is so far from the optimal, that BA has convergence problems. Even after 
100 iterations still the error is very high. On the other hand essential graph optimization shows fast convergence and more accurate results. 
It can be seen that the choice of $\theta_\mathrm{min}$ has not significant effect in accuracy but decreasing the number of edges the time can be significantly reduced. 
Performing an additional BA after the pose graph optimization slightly improves the accuracy while increasing substantially the time.

\begin{table}[t]
\caption{Comparison of loop closing strategies in KITTI 09}

\begin{center}
\begin{tabular}{|c|c|c|c|}
 \multicolumn{4}{c}{} \\
\cline{1-4}
 & & &  \\[-0.8em]
Method & Time (s) & Pose Graph Edges & RMSE (m) \\[0.2em]
\hline
\hline
&&&\\[-0.8em]
- & - & - & 48.77  \\[0.2em]
\hline
&&&\\[-0.8em]
BA (20) & 14.64 & - &  49.90\\[0.2em]
\hline
&&&\\[-0.8em]
BA (100) & 72.16 & - & 18.82  \\[0.2em]
\hline
&&&\\[-0.8em]
EG (200) & 0.38 & 890 & 8.84  \\[0.2em]
\hline
&&&\\[-0.8em]
EG (100) & 0.48 & 1979 & 8.36  \\[0.2em]
\hline
&&&\\[-0.8em]
EG (50) & 0.59 & 3583 & 8.95  \\[0.2em]
\hline
&&&\\[-0.8em]
EG (15) & 0.94 & 6663 & 8.88  \\[0.2em]
\hline
&&&\\[-0.8em]
EG (100) + BA (20) & 13.40 & 1979 & 7.22  \\[0.2em]
\hline
\end{tabular}
\end{center}
\raggedright
First row shows results without loop closing. Number between brackets for BA (Bundle Adjustment) means number of Levenberg-Marquardt (LM) iterations, while for EG (Essential Graph) is the $\theta_\mathrm{min}$ to build the Essential Graph.
All EG optimizations perform 10 LM iterations.
\label{tb:k9}
\end{table}

\begin{figure}[t]
\centering
        \subfigure[Without Loop Closing]
        {
    \includegraphics[width=0.22\textwidth]{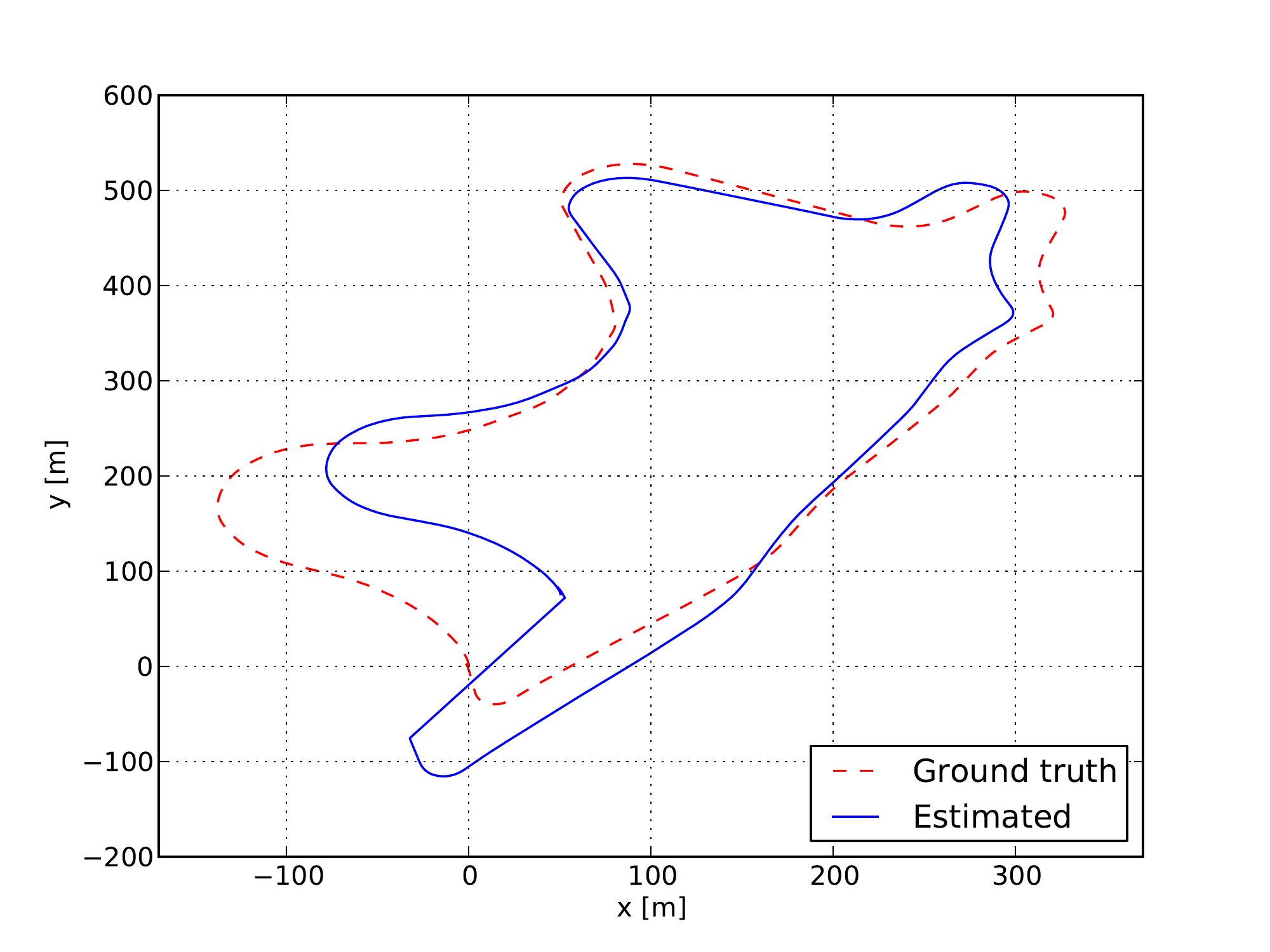}
    \label{fig:09:open}
    }
    \subfigure[BA (20)]
        {
    \includegraphics[width=0.22\textwidth]{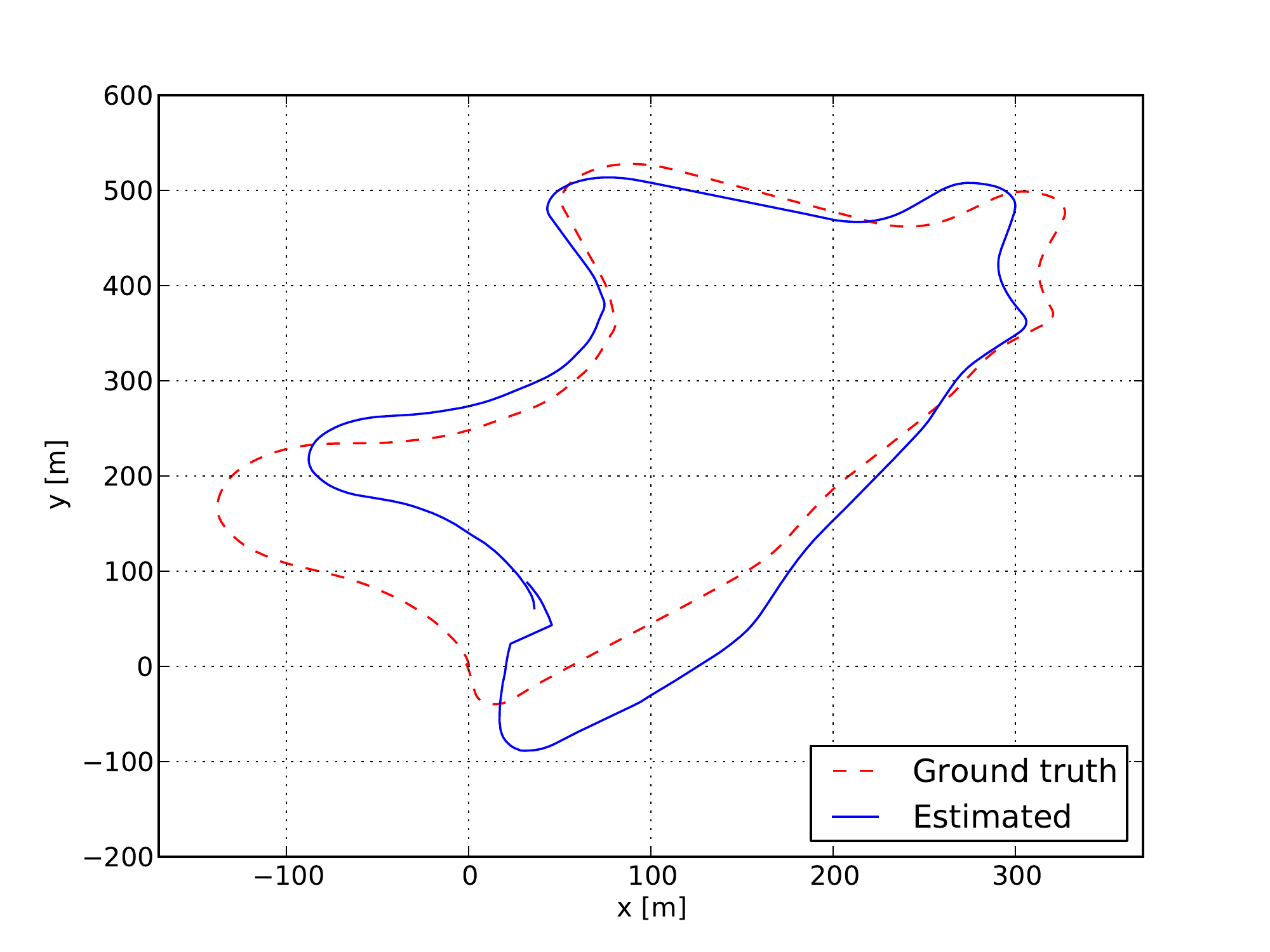}
        \label{fig:09:BA20}
    }
    
    \subfigure[EG (100)]
        {
    \includegraphics[width=0.22\textwidth]{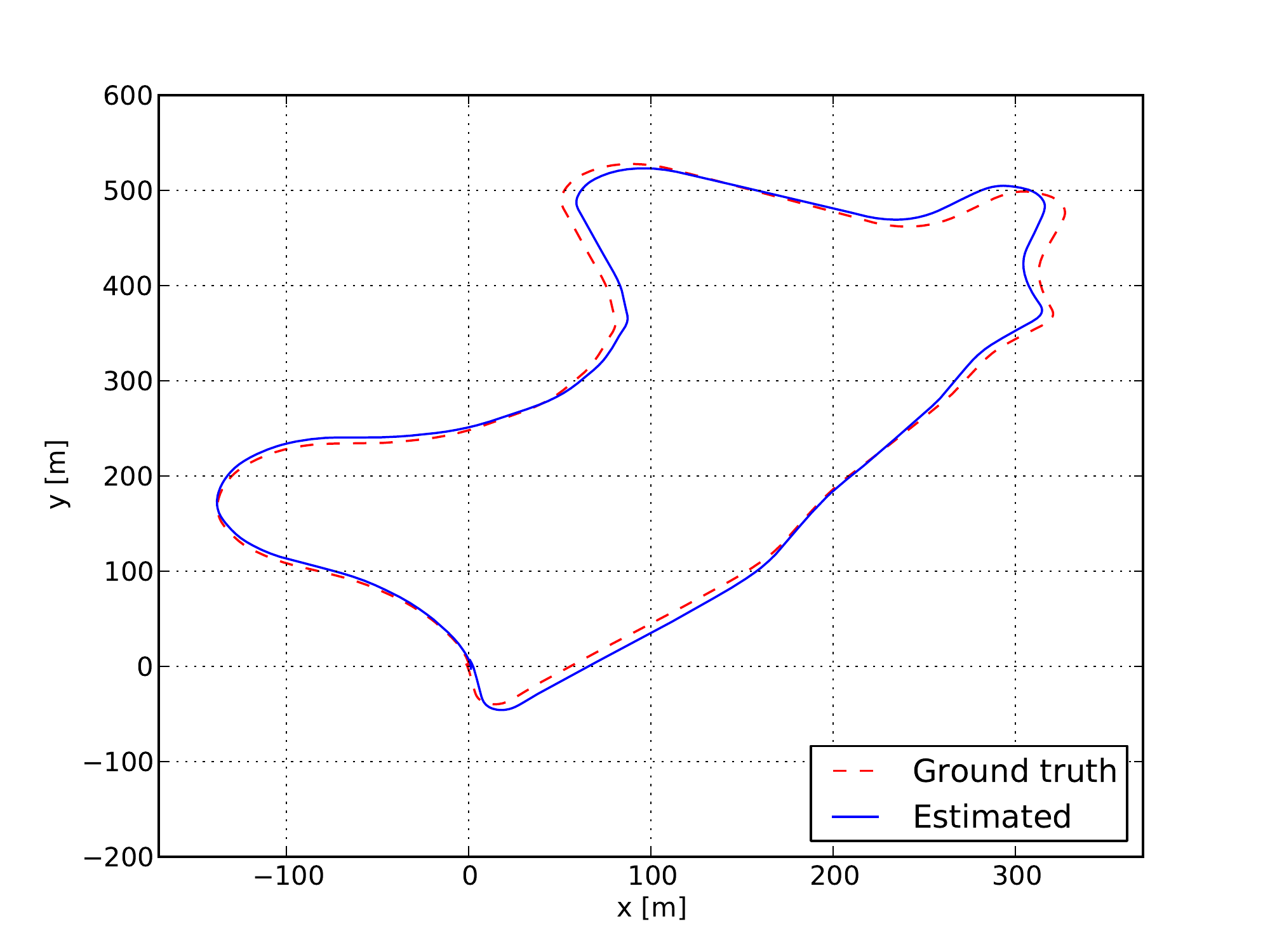}
        \label{fig:09:EG100}
    }
    \subfigure[EG (100) + BA (20)]
        {
    \includegraphics[width=0.22\textwidth]{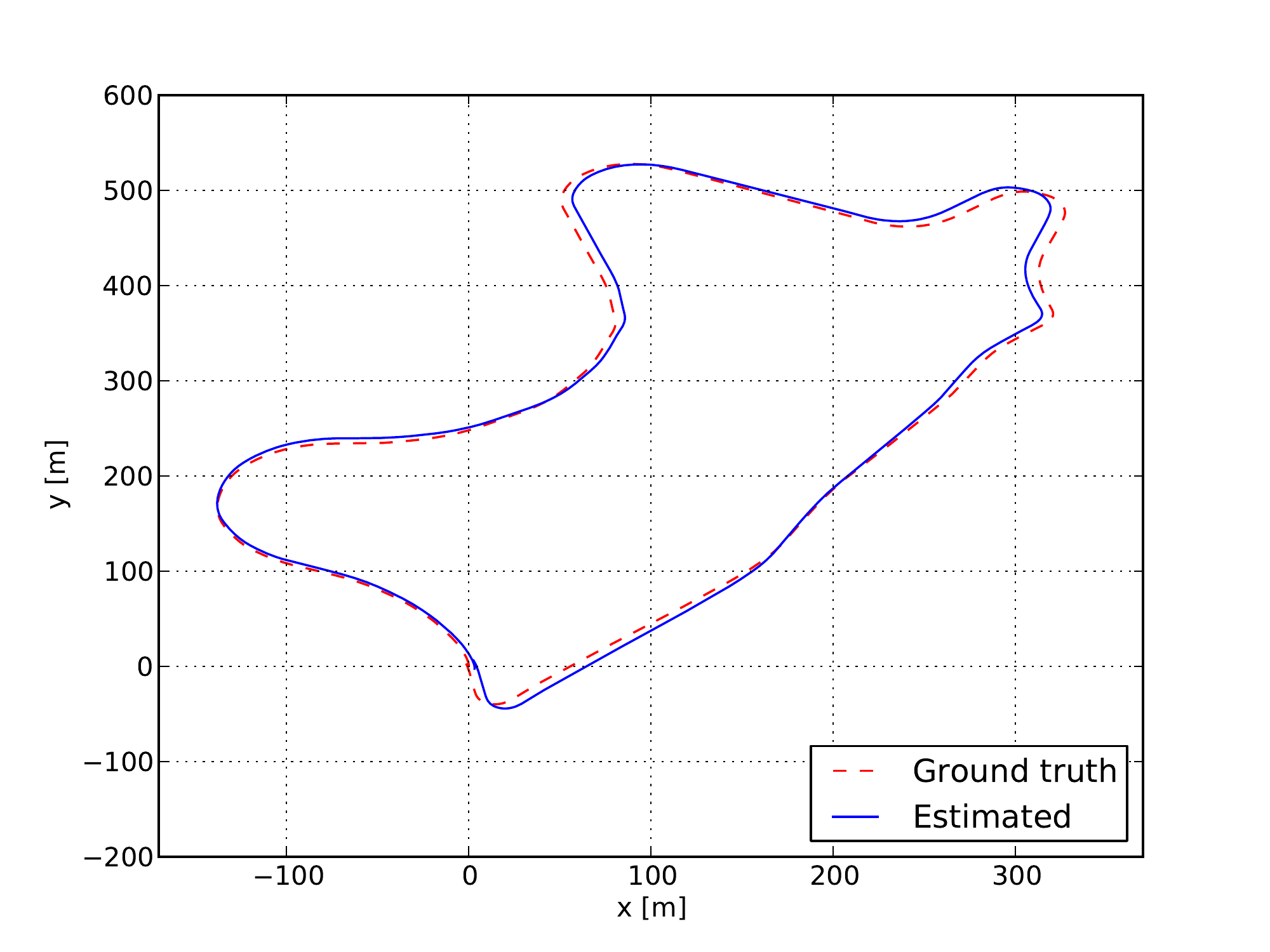}
        \label{fig:09:EG100BA20}
    }
    
    \caption{Comparison of different loop closing strategies in KITTI 09.}
    \label{fig:strategies}
 
\end{figure}

\section{Conclusions and Discussion}\label{sec:concl}

\subsection{Conclusions}
In this work we have presented a new monocular SLAM system with a detailed description of its building blocks and an
exhaustive evaluation in public datasets. Our system has demonstrated that it can process sequences from indoor and outdoor
scenes and from car, robot and hand-held motions. The accuracy of the system is typically below 1 cm in small indoor scenarios and of a few meters
in large outdoor scenarios (once we have aligned the scale with the ground truth). 

Currently PTAM by Klein and Murray \cite{ptam} is considered the most accurate SLAM method from monocular video in real time. 
It is not coincidence that the backend of PTAM is bundle adjustment, which is well known to be
the gold standard method for the offline Structure From Motion problem \cite{hartley}. One of the main successes of PTAM, and the earlier work of Mouragnon \cite{mouragnon}, was 
to bring that knowledge into the robotics SLAM community and demonstrate its real time performance. The main contribution
 of our work is to expand the versatility of PTAM to environments that are intractable for that system. To achieve this, we have designed from scratch a new monocular SLAM system 
 with some new ideas and algorithms, but also incorporating excellent works developed in the past few years, such as 
 the loop detection of G\'alvez-L\'opez and Tard\'os \cite{dorian}, the loop closing procedure and covisibility graph of Strasdat et.al \cite{DWO,haukeScale}, 
 the optimization framework g2o by Kuemmerle et. al \cite{g2o} and ORB features by Rubble et. al \cite{orb}. To the best of our knowledge, 
 no other system has demonstrated to work in as many different scenarios and with such accuracy. Therefore our system is currently the most reliable and complete
 solution for monocular SLAM. Our novel policy to spawn and cull keyframes, permits to create keyframes every few frames, which are eventually removed when considered redundant.  
 This flexible map expansion is really useful in poorly conditioned exploration 
 trajectories, i.e. close to pure rotations or fast movements. When operating repeatedly in the same environment, the map only grows if the visual content of the scene changes, storing a history of its different visual appearances. 
 Interesting results for long-term mapping could be extracted analyzing this history.
 
 Finally we have also demonstrated that ORB features have enough recognition power to enable place recognition from severe viewpoint change. Moreover they are so fast to extract
 and match (without the need of multi-threading or GPU acceleration) that enable real time accurate tracking and mapping.

\subsection{Sparse/Feature-based vs. Dense/Direct Methods} \label{sec:concl:vs}
Recent real-time monocular SLAM algorithms such as DTAM \cite{DTAM} and LSD-SLAM \cite{LSDSLAM} are able to perform dense or semi dense 
reconstructions of the environment, while the camera is localized by optimizing directly over image pixel intensities. These direct approaches do not need
 feature extraction and thus avoid the corresponding artifacts.  They are also more robust to blur, low-texture environments and high-frequency texture like asphalt \cite{rearCamera}.  Their denser reconstructions, as compared to the sparse point map of our system or PTAM, could be more useful for other tasks than just camera localization.

However, direct methods have their own limitations. Firstly, these methods assume a surface reflectance model that in real scenes
produces its own artifacts. The photometric consistency limits the baseline of the matches, typically narrower than those that features allow. 
This has a great impact in reconstruction accuracy, which requires wide baseline observations to reduce depth uncertainty.
Direct methods, if not correctly modeled, are quite affected by rolling-shutter, auto-gain and auto-exposure artifacts (as in the TUM RGB-D Benchmark).
Finally, because direct methods are in general very computationally demanding, the map is just incrementally expanded as in DTAM, or map optimization is reduced to 
a pose graph, discarding all sensor measurements as in LSD-SLAM. 

In contrast, feature-based methods are able to match features with a wide baseline, thanks to their good invariance to viewpoint and illumination changes.  Bundle adjustment jointly optimizes
camera poses and points over sensor measurements. In the context of structure and motion estimation, Torr and Zisserman \cite{torrFeature} already pointed the benefits 
of feature-based against direct methods. In this work we provide experimental evidence (see Section \ref{sec:exp:tum}) of the superior accuracy of feature-based methods
in real-time SLAM. We consider that the future of monocular SLAM should incorporate the best of both approaches.

  \subsection{Future Work}  
  The accuracy of our system can still be improved incorporating points at infinity in the tracking. These points, which are not seen with sufficient
  parallax and our system does not include in the map, are very informative of the rotation of the camera \cite{inverseDepth}.
  
  Another open way is to upgrade the sparse map of our system to a denser and more useful reconstruction. Thanks to our keyframe selection,
  keyframes comprise a compact summary of the environment with a very high pose accuracy and rich information of covisibility. 
  Therefore the ORB-SLAM sparse map can be an excellent initial guess and skeleton, on top of which a dense and accurate map of the scene can be built.  
  A first effort in this line is presented in \cite{rss15}.


%

\appendix[Non-Linear Optimizations]

 \begin{itemize}\itemsep0.5em
  \item Bundle Adjustment (BA) \cite{triggs}: 
  
  Map point 3D locations $\mathbf{X}_{w,j} \in \mathbb{R}^3$ and keyframe poses $\mathbf{T}_{iw}  \in \mathrm{SE}(3)$, where $w$ stands for the world reference, are optimized minimizing the reprojection error with respect to the matched 
  keypoints $\mathbf{x}_{i,j} \in \mathbb{R}^2$. The error term for the observation of a map point $j$ in a keyframe $i$ is:
  \begin{equation}
   \mathbf{e}_{i,j} =  \mathbf{x}_{i,j}-\pi_i(\mathbf{T}_{iw},\mathbf{X}_{w,j}) 
  \end{equation}
   where $\pi_i$ is the projection function:
  \begin{equation}
  \begin{gathered}
 \pi_i(\mathbf{T}_{iw},\mathbf{X}_{w,j})  = 
  \begin{bmatrix}
 f_{i,u} \frac{x_{i,j}}{z_{i,j}}+c_{i,u} \\  f_{i,v} \frac{y_{i,j}}{z_{i,j}}+c_{i,v}   
  \end{bmatrix}
  \\
  \begin{bmatrix} x_{i,j} & y_{i,j} & z_{i,j} \end{bmatrix} ^T = \mathbf{R}_{iw} \mathbf{X}_{w,j}+\mathbf{t}_{iw}
  \end{gathered}
  \end{equation}
   where $\mathbf{R}_{iw} \in \mathrm{SO}(3)$ and $\mathbf{t}_{iw} \in\mathbb{R}^3$ are respectively the rotation and translation parts of $\mathbf{T}_{iw}$, and $(f_{i,u}, f_{i,v})$ and $(c_{i,u}, c_{i,v})$ are the focal
  length and principle point associated to camera $i$.  The cost function to 
  be minimized is:
  \begin{equation}
   C = \sum_{i,j} \rho_{h}(\mathbf{e}_{i,j}^T\mathbf{\Omega}^{-1}_{i,j} \mathbf{e}_{i,j})
  \end{equation}
  where $\rho_h$ is the Huber robust cost function and $\mathbf{\Omega}_{i,j}=\sigma_{i,j}^2\mathbf{I}_{2\times2}$ is the covariance matrix 
  associated to the scale at which the keypoint was detected. 
  In case of \emph{full BA} (used in the map initialization explained in Section \ref{sec:ini} and in the experiments in Section \ref{sec:exp:kitti}) we optimize all points and keyframes, by the exception of the first keyframe which remain fixed as the origin.
  In \emph{local BA} (see section \ref{sec:sub:localba}) all points included in the local area are optimized, while a 
  subset of keyframes is fixed. In pose optimization, or \emph{motion-only BA}, (see section \ref{sec:track}) all points are fixed and only the camera pose is optimized.

  \item Pose Graph Optimization over Sim(3) Constraints \cite{haukeScale}:
  
   Given a pose graph of binary edges (see Section \ref{sec:sub:pg}) we define the error in an edge as:
  \begin{equation}
   \mathbf{e}_{i,j} = \log_{\mathrm{Sim}(3)}(\mathbf{S}_{ij}\,\mathbf{S}_{jw}\,\mathbf{S}^{-1}_{iw})
  \end{equation}
  where $\mathbf{S}_{ij}$ is the relative Sim(3) transformation between both keyframes computed from the SE(3) poses just before the pose graph optimization and setting the scale factor to 1.
  In the case of the loop closure edge this relative transformation is computed with the method of Horn \cite{horn}. The $\log_{\text{Sim3}}$ \cite{haukeThesis} transforms to the tangent 
  space, so that the error is a vector in $\mathbb{R}^7$.
  The goal is to optimize the Sim(3) keyframe poses minimizing
  the cost function:
   \begin{equation}
   C = \sum_{i,j} (\mathbf{e}_{i,j}^T\mathbf{\Lambda}_{i,j} \mathbf{e}_{i,j})
  \end{equation}
  where $\mathbf{\Lambda}_{i,j}$ is the information matrix of the edge, which, as in \cite{haukeThesis}, we set to the identity. We fix the loop closure keyframe to fix the 7 degrees of gauge freedom.
  Although this method is a rough approximation of a \emph{full BA}, we demonstrate experimentally in Section \ref{sec:exp:kitti} that it has significantly faster and 
  better convergence than BA. 
  
  \item Relative Sim(3) Optimization:
  
  Given a set of $n$ matches $i\Rightarrow j$ (keypoints and their associated 3D map points) between keyframe $1$ and keyframe $2$, we want to optimize the 
  relative Sim(3) transformation $\mathbf{S}_{12}$ (see Section \ref{sec:sub:sim3}) that minimizes the reprojection error in both images:
   \begin{equation}
    \begin{gathered}
   \mathbf{e_{1}} =  \mathbf{x}_{1,i}-\pi_1(\mathbf{S}_{12},\mathbf{X}_{2,j}) \\
   \mathbf{e_{2}} =  \mathbf{x}_{2,j}-\pi_2(\mathbf{S}^{-1}_{12},\mathbf{X}_{1,i})   
  \end{gathered}
  \end{equation}  
  and the cost function to minimize is:
   \begin{equation}
   C = \sum_n \big(\rho_{h}(\mathbf{e}^T_{1}\mathbf{\Omega}^{-1}_{1,i} \mathbf{e}_{1}) + \rho_{h}(\mathbf{e}^T_{2}\mathbf{\Omega}^{-1}_{2,j} \mathbf{e}_{2}) \big)
  \end{equation}
  where  $\mathbf{\Omega}_{1,i}$ and $\mathbf{\Omega}_{2,i}$ are the covariance matrices associated to the scale in which keypoints in image 1 and image 2 were detected.
  In this optimization the points are fixed.

  \end{itemize}

\ifCLASSOPTIONcaptionsoff
  \newpage
\fi




%
%
%

%

\begin{IEEEbiography}[{\includegraphics[width=1in,height=1.25in,clip,keepaspectratio]{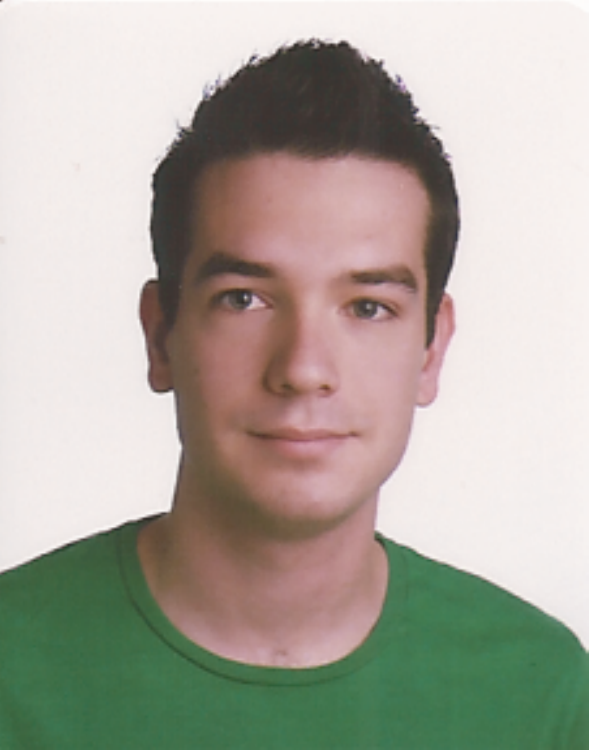}}]{Ra\'ul Mur Artal}
was born in Zaragoza, Spain in 1989. He received the Industrial Engineering degree (mention in Industrial Automation and Robotics) in 2012 and the M.S. degree in Systems and Computer Engineering
 in 2013 from the University of Zaragoza, where he is currently working towards the PhD. degree with the I3A Robotics, Perception and Real-Time Group.
 
 His research interests include Visual Localization and Long-Term Mapping.
\end{IEEEbiography}

\begin{IEEEbiography}[{\includegraphics[width=1in,height=1.25in,clip,keepaspectratio]{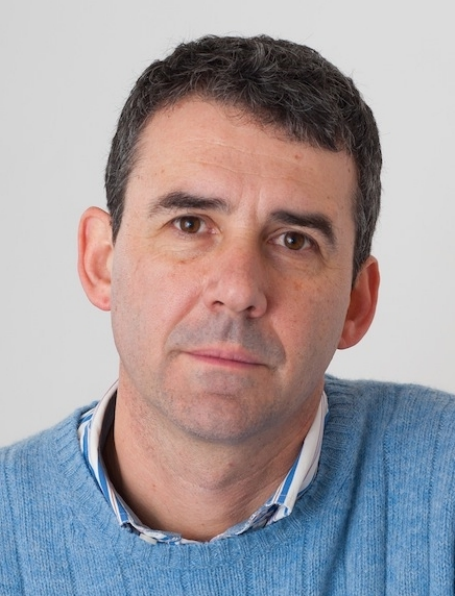}}]{J. M . M. Montiel}
was born in Arnedo, Spain, in 1967. He received the M.S. and Ph.D. degrees in
electrical engineering from the Universidad de Zaragoza,
Spain, in 1992 and 1996, respectively. He is currently a Full
Professor with the Departamento de
Inform\'atica e Ingenier\'ia de Sistemas, Universidad
de Zaragoza, where he is in charge of Perception and
Computer Vision research grants and courses. His current
interests include, real-time vision localization and semantic
mapping for rigid and non rigid environments, and the
transference of this technology to robotic and nonrobotic
application domains. Prof. Mart\'inez Montiel is a member
of the I3A Robotics, Perception, and Real-Time Group. He
has been awarded several Spanish MEC grants to fund research at the
University of Oxford, UK, and at Imperial College London, UK.
\end{IEEEbiography}

\begin{IEEEbiography}[{\includegraphics[width=1in,height=1.25in,clip,keepaspectratio]{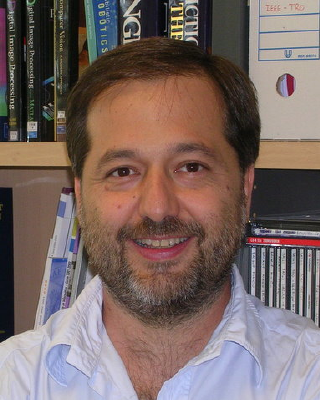}}]{Juan D. Tard\'os}
was born in Huesca, Spain, in
1961. He earned the M.S. and Ph.D. degrees in electrical
engineering from the University of Zaragoza, Spain, in 1985 and
1991, respectively. He is Full Professor with the Departamento de
Inform\'atica e Ingenier\'ia de Sistemas, University of Zaragoza,
where he is in charge of courses in robotics, computer vision, and
artificial intelligence. His research interests include
SLAM, perception and mobile robotics. Prof. Tard\'os is a member
of the I3A Robotics, Perception, and Real-Time Group
\end{IEEEbiography}







\end{document}